\documentclass[twoside]{article}
\usepackage[accepted]{aistats2024}
\usepackage{times}  
\usepackage{helvet}  
\usepackage{courier}  
\usepackage[hyphens]{url}  
\usepackage{graphicx} 
\usepackage[round]{natbib}

\usepackage{caption} 
\usepackage{colortbl}
\usepackage[title]{appendix}
\usepackage{afterpage}

\newtheorem{theorem}{Theorem}

\newtheorem{definition}[theorem]{Definition}

\usepackage{cite}
\usepackage{subcaption}
\usepackage{makecell}
\usepackage{siunitx}
\usepackage{amssymb}
\usepackage{algorithm}
\usepackage{algorithmic}
 \newcommand{\reference}{population attribute reference}
\newcommand{\volopt}{Vol-Opt}
\newcommand{\allopt}[1]{All-Opt$^{#1}$}
\newcommand{\sev}[1]{SEV$^{#1}$}
\newcommand{\bx}{\boldsymbol{x}}
\newcommand{\bb}{\boldsymbol{b}}
\newcommand{\be}{\boldsymbol{e}}
\newcommand{\q}{\textrm{query}}

\usepackage[utf8]{inputenc} 
\usepackage[T1]{fontenc}    
\usepackage{url}            
\usepackage{booktabs}       
\usepackage{amsmath}        
\usepackage{nicefrac}       
\usepackage{microtype}      
\usepackage{xcolor}         
\usepackage{multirow}
\usepackage{multicol}

\usepackage{newfloat}
\usepackage{listings}
\DeclareCaptionStyle{ruled}{labelfont=normalfont,labelsep=colon,strut=off} 
\floatstyle{ruled}
\newfloat{listing}{tb}{lst}{}
\floatname{listing}{Listing}

\newif\ifshow
\showtrue

\newif\ifappendix
\appendixtrue

\newif\ifmainpaper
\mainpapertrue

\begin{document}

\bibliographystyle{unsrtnat}

\twocolumn[

\aistatstitle{Sparse and Faithful Explanations Without Sparse Models}

\aistatsauthor{Yiyang Sun$^*$ \And Zhi Chen$^*$ \And  Vittorio Orlandi \And Tong Wang \And Cynthia Rudin}

\aistatsaddress{Duke University \And  Duke University \And Duke University \And Yale University \And Duke University} ]

\ifmainpaper{


    \def\thefootnote{*}\footnotetext{These authors contributed equally to this work}\def\thefootnote{\arabic{footnote}}

\begin{abstract}
Even if a model is not globally sparse, it is possible for decisions made from that model to be accurately and faithfully described by a small number of features. For instance, an application for a large loan might be denied to someone because they have no credit history, which overwhelms any evidence towards their creditworthiness. In this work, we introduce the \textit{Sparse Explanation Value (SEV)}, a new way of measuring sparsity in machine learning models. In the loan denial example above, the SEV is 1 because only one factor is needed to explain why the loan was denied. SEV is a measure of \textit{decision sparsity} rather than overall model sparsity, and we are able to show that many machine learning models -- even if they are not sparse -- actually have low decision sparsity, as measured by SEV. SEV is defined using movements over a hypercube, allowing \sev{} to be defined consistently over various model classes, with movement restrictions reflecting real-world constraints. We propose algorithms that reduce SEV without sacrificing accuracy, providing sparse and completely faithful explanations, even without globally sparse models. 
\end{abstract}

\section{Introduction}

As machine learning is increasingly leveraged in business and societal contexts to make important decisions, there is an increasing need for insight into why these decisions were made. One classical important measure of interpretability is sparsity; famously, humans can handle only $7 \pm 2$ cognitive entities at once \citep{miller1956magical}. Traditionally, sparsity is a property of the entire model, which we call global sparsity, for instance, the total number of terms in linear models, the total number of parameters in neural networks, or the number of leaves in decision trees \citep{murdoch2019definitions}. In this work, we argue that this notion of sparsity, while undeniably useful, may be overly restrictive. This is because \textit{we do not need the model to be globally sparse in order to generate a sparse explanation for use in decision-making}. A far more relevant desideratum for users is that the explanation for each \emph{individual prediction} is sparse. By this, we mean that the local explanation for a specific prediction depends only on a small number of features, which may differ across units. Since individuals care only about decisions made about themselves, we should consider aiming for sparse \emph{explanations} -- and not necessarily sparse \emph{models}. That is, having an explanation for each prediction that does not necessarily contain all factors used by the model. Having sparse explanations, where we might faithfully explain a given prediction using only 1-3 features, allows for increased transparency in model-assisted decision-making processes, even if the model is not globally sparse.

To this end, we introduce a new metric for measuring the decision sparsity of classification models: the \emph{Sparse Explanation Value (SEV)}. The SEV is defined by moving features from their values to the \textit{population commons} (we call this our \textit{reference}) or vice versa; the number of features that must be \textit{aligned} (i.e., equate) to (or from) the reference in order to change the sign of the prediction is the SEV of the observation. SEV explains how many features change for predictions to switch from one class to another. \sev{} is easily explained and understood, and as we will see, the decision sparsity from \sev{} can be much sparser for each instance than the number of terms in a globally sparse model. 

For example, consider a scenario where a loan applicant is denied a large loan due to their lack of credit history. This denial might occur even though the applicant has several positive factors, such as having a bank account, being employed, and not having a criminal record. In this case, the \sev{} is 1, which means that simply changing from having no credit history to having a credit history can flip the prediction, despite the other factors, and regardless of the total number of features in the model. If most of the feature values are 0 for a given individual (perhaps most people do not have complicated financial histories), or if most predictions depend only on a small set of factors (where important factors can differ across predictions), the global sparsity of the model is essentially irrelevant.
Hence, SEV is an alternative way of measuring sparsity for these cases.


 This work introduces the SEV and discusses its properties. SEV (as well as global sparsity) are most relevant for tabular datasets where each feature is meaningful. We show that decision sparsity, as measured by SEV, is naturally low in most common types of machine learning models already, without modification. \textbf{This means that despite the complexity of these models, their decisions are often based only on a few factors.} While SEV is useful as a post-hoc explanation method, it begs the question of whether models with better prediction-level sparsity can be created. In other words, if we care about sparse explanations, we should target them directly instead of obtaining globally sparse models and hoping they give the sparsest explanations. \textbf{Our second contribution is to create models that are optimized to yield sparse (i.e., low-SEV) explanations.} We do this by introducing SEV loss terms that can be used with a variety of models (linear, boosted trees, multi-layer perceptrons, etc.). 
 Using real-world datasets, we show that SEV optimization is effective in improving decision sparsity.

\section{Related Works}
\label{sec:related}
The concept of SEV revolves around finding models that are simple, in that the explanations for their predictions are sparse, while recognizing that different predictions can be simple in different ways (i.e., involving different features). In this way, it relates to (i) globally sparse models, (ii) local classification methods, which predict the outcomes of different units using local models, and (iii) black box explanation methods, which seek to explain predictions of complex models. We further comment on these below. 

\textbf{Global Model Sparsity.}  Loosely, globally sparse models \citep[e.g.,][]{RudinEtAlSurvey2022, sparsity_review, lasso,bach2012optimization, zhang2015survey} do not have many ``components,'' such as non-zero coefficients for additive models or leaves of a decision tree. Our focus in this paper is not overall model complexity, but rather the complexity of the individual predictions. These two notions of sparsity — global and local — are undoubtedly related. Even if every variable in a sparse model is used in every prediction, each explanation is necessarily sparse. 
However, non-sparse models can yield lower SEVs, as shown in Section \ref{subsec:low_sev}.

\noindent \textbf{Local Regression and Classification Methods.} Such methods \citep[e.g.,][]{local_regression, local_class, datum-wise_class} fit separate models for each data point, giving more weight to nearby points, and then possibly combining these models for final predictions. It is problematic to use numerous models instead of a single global model for applications like loan decisions since we do not want each person to be assigned their own credit scoring model; i.e., it is problematic that these models are not globally consistent.

\noindent \textbf{Instance-wise Explainability and Interpretability}

There has been a large push to develop methods to explain the predictions of black boxes \citep[e.g.,][]{explain_black_box, lime, anchors, shapley, Baehrens2010}. Those methods evaluate the contribution of features to the final outcome of the model. Some works are limited to particular domains and data types, such as images \citep[e.g.,][]{explain_image1, explain_image2, boreiko2022sparse}, or text \citep[e.g.,][]{ lei2016rationalizing, li2016understanding, explain_nlp, bastings2019interpretable, yu2019rethinking, yu2021understanding}. \citet{chen2018learning} provides a subset features selection algorithm called \textit{Learning to Explain (L2X)}, but with an entirely different premise -- that sparsity values are fixed by the user and not adaptive. A more detailed comparison can be found in Appendix \ref{subsec:chen2018-sev-compare}. 

\emph{Counterfactual explanations} and \emph{algorithmic recourse} study what could have happened if input to a model had been changed.  Some methods are limited to text \citep{martens2014explaining}, or specific model classes, like linear models \citep{ustun2019actionable}, tree ensembles \citep{cui2015optimal,tolomei2017interpretable,lucic2022focus}, support vector machines \citep{gupta2019equalizing}, or neural networks \citep{zhang2018interpreting}. 
Most research in counterfactual explanations or algorithmic recourse focuses on minimizing $\ell_2$ distance \citep{ross2021learning}, $\ell_1$ distance \citep{wachter2017counterfactual,zhang2018interpreting,grath2018interpretable,russell2019efficient,joshi2019towards}, cosine distance \citep{ramon2020comparison}, the overall cost of changing features by various amounts, a combination of distances \citep{laugel2018comparison,dhurandhar2018explanations,van2021interpretable,cheng2020dece} or model function outputs \citep{lash2017generalized,lash2017budget} between the instance's $\textbf{x}$ value and the closest point across the decision boundary. An important distinction is that we use $\ell_0$ distance between explanations to measure sparsity. The only paper we know of that uses $\ell_0$ distance is that of  \citet{fernandez2022explaining}, which is a special case of our \sev{-}; \citet{fernandez2022explaining} does not discuss sparsity as an evaluation metric for explanations. As discussed, prediction sparsity is important: a classic counterfactual explanation might involve many changes to someone's credit and lifestyle since the shortest path to crossing the decision boundary generally involves changes to many variables. Our explanations are sparse (e.g., ``you have no credit history''), so they are easier to understand.



Additionally, there is ample evidence from human studies that the counterfactual explanations generated by most methods are not interpretable to humans because they aim only to cross the decision boundary, where the decision made by the model is not obvious to a human. Extensive experiments of \citet{counterfac_img} indicate that minimal changes to an observation that alter the predicted class are not natural for humans because the decision boundary is crossed in a low-density part of the space where the observation is not clearly one class or the other. 
 Our best competitor (counterfactual explanation method DiCE) has precisely this problem, shown in Section \ref{subsec:sev_xai}.
By framing SEV with respect to a \reference{} (i.e., typical values of features), it provides explanations that humans are more likely to understand. Importantly, \citet{counterfac_img} show that counterfactual explanations created by humans are not close to the decision boundary; people create counterfactuals that appear to be class-specific prototypes.

\section{Sparse Explanation Values}
\label{sec:sev}

Let us define SEV. 
We are in the setting of classification, where we have observations (units) with features 
$\bx\in \mathcal{X} \subseteq \mathbb{R}^p$ and an outcome of interest. Here, we consider binary outcomes $\{0, 1\}$, though the extension to multi-class outcomes is straightforward. We predict the outcome using a classifier $f: \mathcal{X}\rightarrow \{0,1\}$ that maps the input $\bx$ to a binary prediction $\hat{y}$. 
We consider explanations for only one class; to consider explanations for the other class, one would use a completely symmetric procedure (defining a reference for the other class).
Without loss of generality, we assume class 1 is the class of interest (positive) and we focus on explaining the predictions of a unit (which we call a \textit{query}, denoted $\bx^{\q}$) that is classified as positive ($f(\bx^{\q})=1$). As shown by human studies \citep{counterfac_img}, it is natural to compare a query with a \textit{reference}; for example, a unit representative of the population as a whole or of negatively predicted (predicted to be class 0) units. 
Therefore, we define explanation sparsity with respect to the reference $\boldsymbol{\tilde{x}} \in \mathbb{R}^p$, where $\tilde{x}_j$ is the reference value of feature $j$, e.g., the population mean or median of that feature in the dataset. We will consider aligning the observed values of the query $\bx_{j}^{\q}$ with the corresponding reference feature values $\tilde{\bx}_j$ to see whether predictions change. Thinking of these adjustments as binary moves, it is convenient to represent all $2^p$ possible different feature sets as vertices of on a Boolean hypercube. The hypercube is defined below:


\begin{definition}[\sev{} hypercube]
    An \sev{} hypercube $\mathcal{L}_{f,\q}$ for a model $f$, reference $\tilde{\bx}$, and query $\bx^{\q}$, such that $f(\bx^{\q}) = 1$, is a graph with $2^{p}$ vertices, where each vertex $v$ is represented by a $p$-dimensional Boolean vector $\bb_v\in\{0,1\}^p$. Vertices $u$ and $v$ are adjacent when their Boolean vectors differ by one bit: $\|\bb_u-\bb_v\|_0=1$.  0's in $\bb_v$ indicate the reference value $\tilde{\bx}$ of the corresponding feature, while 1's indicate the true feature value of the query. Thus, the actual feature vector represented by the vertex $v$ is $\bx_v:=\bb_v \odot \bx^{\q} + (\mathbf{1}-\bb_v)\odot\tilde{\bx}$, where $\odot$ is the Hadamard product. The score of vertex $v$ is $f(\bx_v)$, also denoted as $\mathcal{L}_{f, \q}(\bb_v).$
\end{definition}

The \sev{} hypercube definition can also be extended from a hypercube to a Boolean lattice as they have the same geometric structure. Based on this definition of \sev{} hypercubes, we define two variants of the Sparse Explanation Value: one is the smallest number of features that, starting from the reference (corresponding to the origin of the hypercube $\bb=\mathbf{0}$), need to be aligned with the query's feature values to make the prediction flip from negative to positive (called \sev{+}). The other (called \sev{-}) is the counterfactual case: it is the minimal number of features, starting from the query (corresponding to $\bb=\mathbf{1}$) that need to be aligned with the reference's feature values to make the prediction flip from positive to negative. Formally:

\begin{definition}[SEV Plus, denoted \sev{+}]
    For a query predicted as positive ($f(\bx^{\q})=1$), the Sparse Explanation Value Plus (\sev{+}) is the length of the shortest path from the reference vertex to any positively predicted vertex, i.e.,
    \begin{equation}
    \begin{aligned}
        \min_{\bb \in \{0,1\}^p} & \quad \|\bb\|_0 \quad
        \textrm{s.t.} & \quad \mathcal{L}_{f, \q}(\bb)=1.
    \end{aligned}
    \end{equation}
\end{definition}

\begin{definition}[SEV Minus, denoted \sev{-}]
    For a query predicted as positive ($f(\bx^{\q})=1$), the Sparse Explanation Value Minus (\sev{-}) is the length of the shortest path from the query to any negatively predicted vertex, 
    \begin{equation}
    \begin{aligned}
        \min_{\bb \in \{0,1\}^p} & \quad \|\mathbf{1}-\bb\|_0 \quad
        \textrm{s.t.} & \quad \mathcal{L}_{f, \q}(\bb)=0.
    \end{aligned}
    \end{equation}
\end{definition}

\sev{-} is similar to a counterfactual explanation in which we use the length of the path on the \sev{} hypercube to measure the counterfactual change (see also Section \ref{subsec:sev_xai}). Figure \ref{fig:boolean_plus}, \ref{fig:boolean_minus} and Table \ref{tab:sev_plus example}, \ref{tab:sev_minus example} show examples of how \sev{+}$=1$ and \sev{-}$=2$ are calculated in a loan credit evaluation decision involving three features: housing = renting, loan amount $\geq$ 10k, and education level = high school level, where we need an \sev{} hypercube with $2^3=8$ vertices. The red vertex in Figure \ref{fig:boolean_plus} is encoded as $(1,1,1)$, which represents the query; the dark blue vertex is encoded as $(0,0,0)$, representing the negatively-predicted reference value; the orange vertices are positively-predicted vertices; and the light blue ones are negatively predicted vertices. To compute \sev{+}, we start from the reference $(0,0,0)$ and find the shortest path to any positive (red/orange) nodes. Here, the two shortest paths end at $(0,1,0)$, which means that if we change the loan amount from 5-10k to more than 10k, then the model will evaluate it as high risk. Since we are changing only one feature (loan amount), \sev{+} equals 1, and the feature vector that $(0,1,0)$ represents is called an \textit{\sev{+} explanation}. For \sev{-}, we start from the query $(1,1,1)$, and find the shortest path to any blue node. Here the shortest paths have length 2 (e.g., $(1,1,1)\to(0,1,1)\to(0,0,1)$), which means that if we changed the housing status of the renter from renting to managing and the loan amount from greater than 10k to 5-10k, then the credit risk will be evaluated as low instead of high. Moreover, there also exists other shortest paths like $(1,1,1)\to(1,1,0)\to(1,0,0)$ which means we can also align the loan amount and the education level with their reference values to obtain a low-risk prediction. This alternative path to a negative vertex shows that we are able to provide multiple sparse explanations through the SEV hypercube. The feature vector that $(1,0,0)$ represents is called an \textit{\sev{-} explanation}.

Figures \ref{fig:sev_plus} and \ref{fig:sev_minus} show how \sev{} is distributed in 2D cases. The blue line is the decision boundary. The orange horizontal and vertical lines are where the decision boundary intersects one of the reference's two feature values. If the query's feature values are in the yellow area of the \sev{+} figure (Figure \ref{fig:sev_plus}), it means that for both features, the reference feature value must be changed to the query's feature values in order to be predicted positive. In the yellow region of Figure \ref{fig:sev_minus} (\sev{-} case), for both features, the query's feature values must be aligned with the reference values in order to be predicted as negative.
 
\begin{figure}[ht]
     \centering
     \begin{subfigure}[b]{0.23\textwidth}
         \centering
         \includegraphics[width=\textwidth]{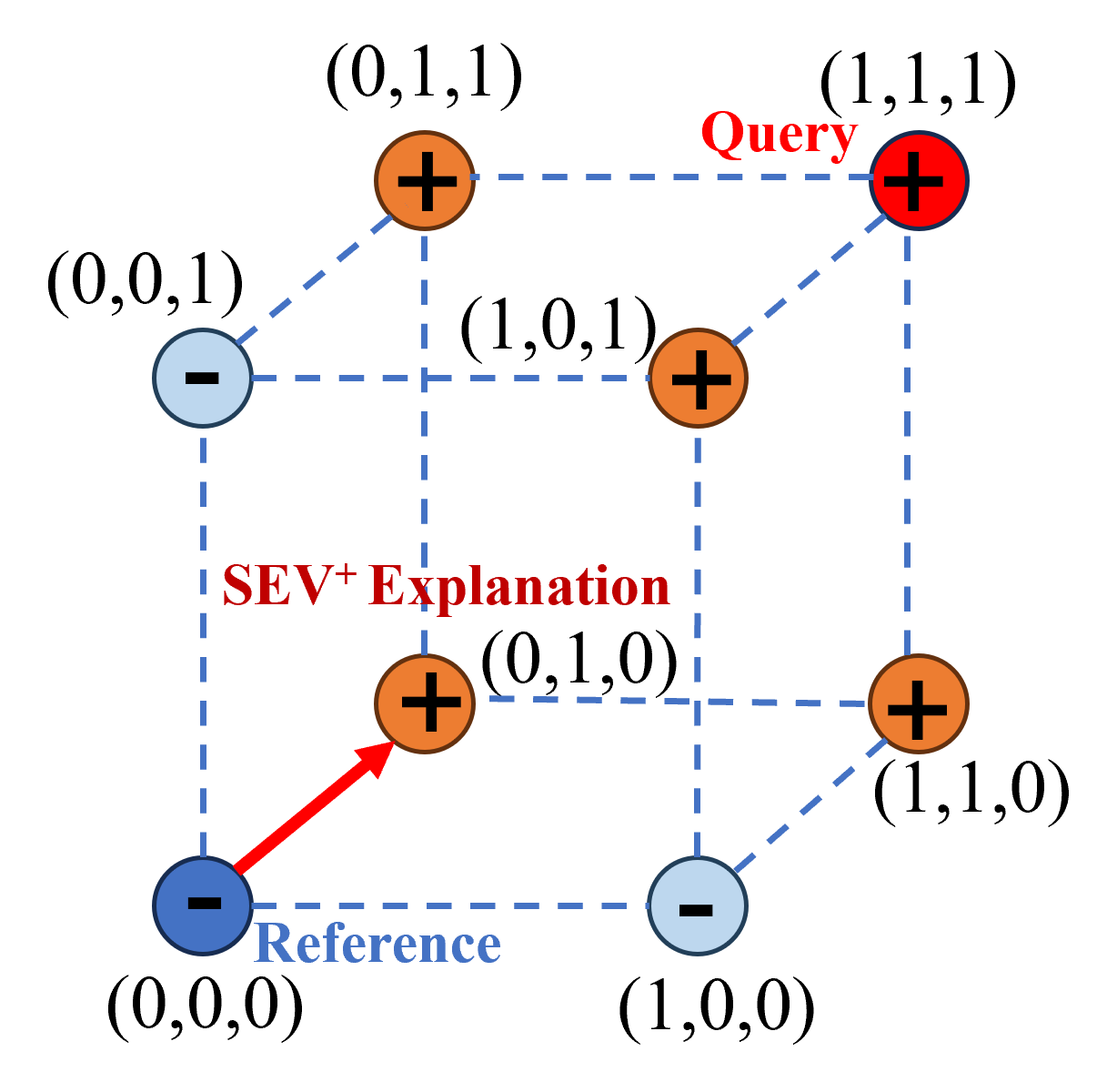}
         \caption{\sev{+}=1 calculation in \sev{} Hypercube}
         \label{fig:boolean_plus}
     \end{subfigure}
     \hfill
     \begin{subfigure}[b]{0.23\textwidth}
         \centering
         \includegraphics[width=\textwidth]{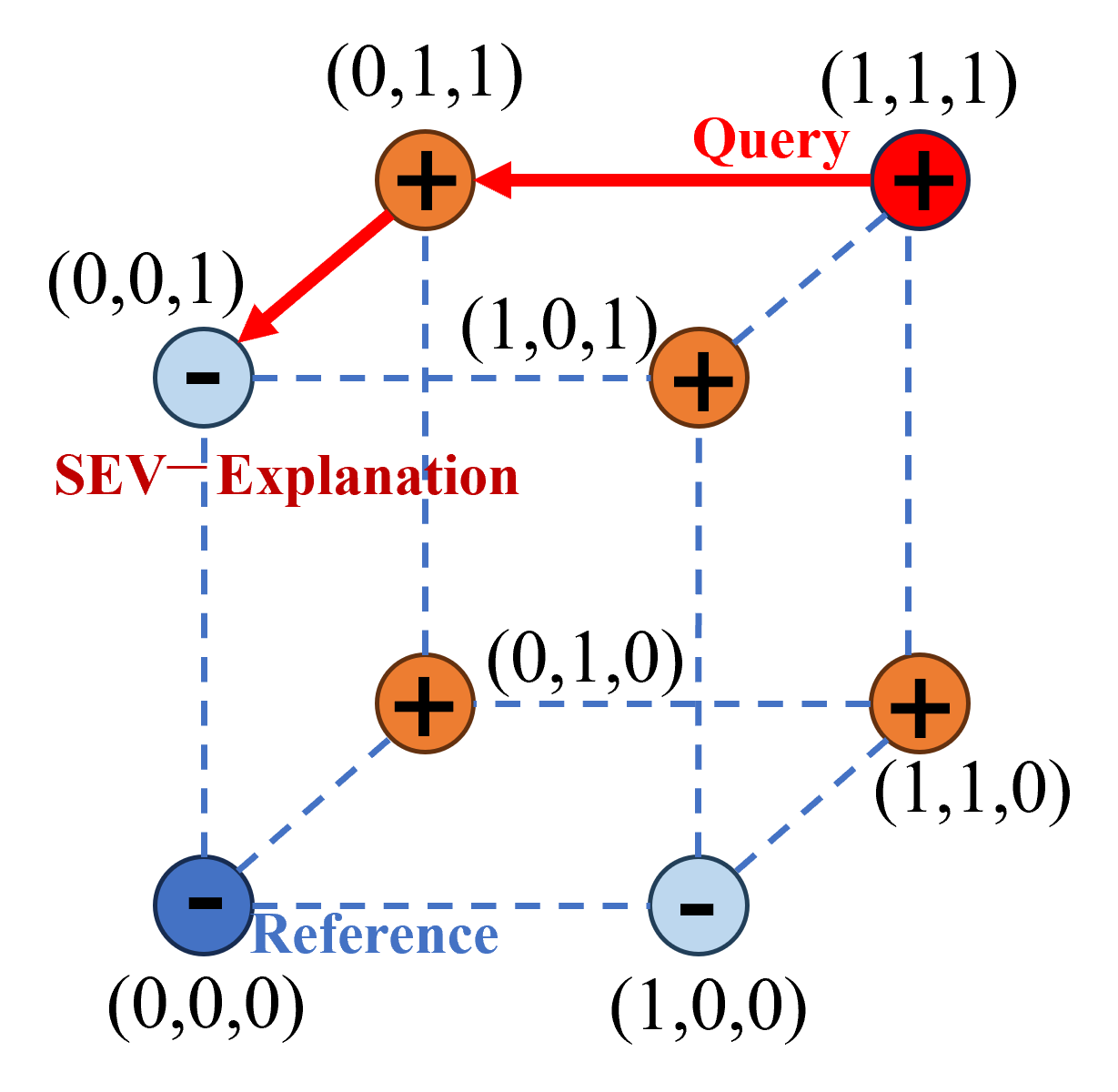}
         \caption{\sev{-}=2 calculation in \sev{} Hypercube}
         \label{fig:boolean_minus}
     \end{subfigure}
     \begin{subfigure}[5]{0.5\textwidth}
        \centering
        \caption{\sev{+}=1 example}
        \scalebox{0.7}{
        \begin{tabular}{cccccc}
\hline
 & \textbf{Location} & \textbf{Housing} & \textbf{Loan} & \textbf{Education} & \textbf{Credit Risk} \\ \hline
\textbf{Reference} & (0,0,0) & Mortgage & 5-10k & Bachelor & Low \\
\textbf{\makecell[c]{\sev{+}\\ Explanation}} & (0,1,0) & {\color[HTML]{000000} Mortgage} & {\color[HTML]{FE0000} >\$10k} & Bachelor & {\color[HTML]{FE0000} High} \\ \hline
\textbf{Query} & (1,1,1) & {\color[HTML]{000000} Rent} & {\color[HTML]{000000} >\$10k} & {\color[HTML]{000000} High School} & {\color[HTML]{000000} High} \\ \hline
\end{tabular}}
        
         \label{tab:sev_plus example}
     \end{subfigure}

     \begin{subfigure}[5]{0.5\textwidth}
        \centering
        \caption{\sev{-}=2 example}
        \scalebox{0.7}{
        \begin{tabular}{cccccc}
\hline
\textbf{} & \textbf{Location} & \textbf{Housing} & \textbf{Loan} & \textbf{Education} & \textbf{Credit Risk} \\ \hline
\textbf{Query} & (1,1,1) & Rent & >\$10k & High School & High \\
\textbf{} & (0,1,1) & {\color[HTML]{FE0000} Mortgage} & >\$10k & High School & High \\
\textbf{\makecell[c]{\sev{-}\\Explanation}} & (0,0,1) & {\color[HTML]{FE0000} Mortgage} & {\color[HTML]{FE0000} 5-10k} & High School & \textcolor{blue}{Low} \\ \hline
\textbf{Reference} & (0,0,0) & Mortgage & 5-10k & Bachelor & Low \\ \hline
\end{tabular}}
        
         \label{tab:sev_minus example}
     \end{subfigure}
     \begin{subfigure}[b]{0.23\textwidth}
         \centering
         \includegraphics[width=\textwidth]{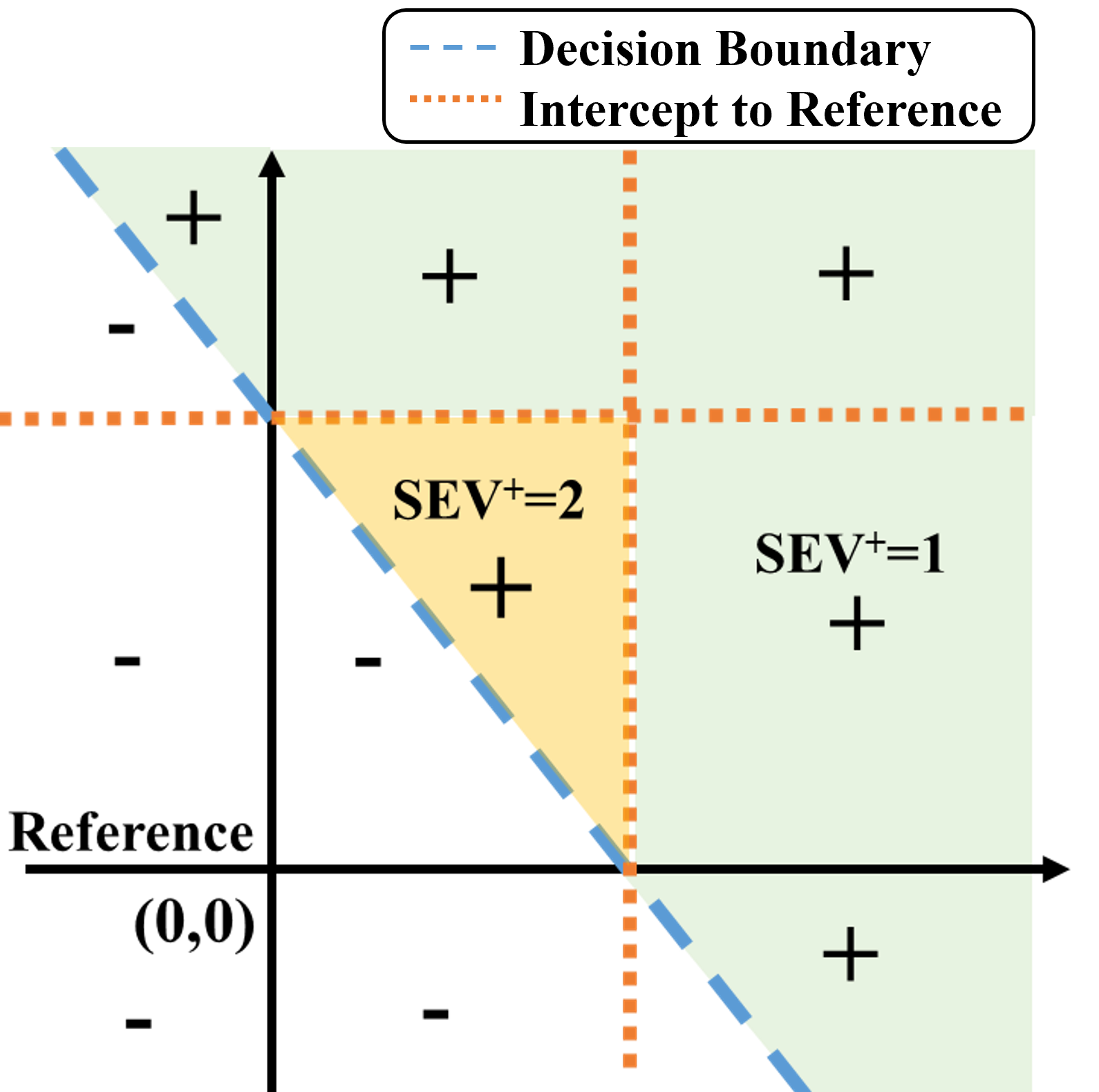}
         \caption{\sev{+} in linear classifier}
         \label{fig:sev_plus}
     \end{subfigure}
    \begin{subfigure}[b]{0.23\textwidth}
         \centering
         \includegraphics[width=\textwidth]{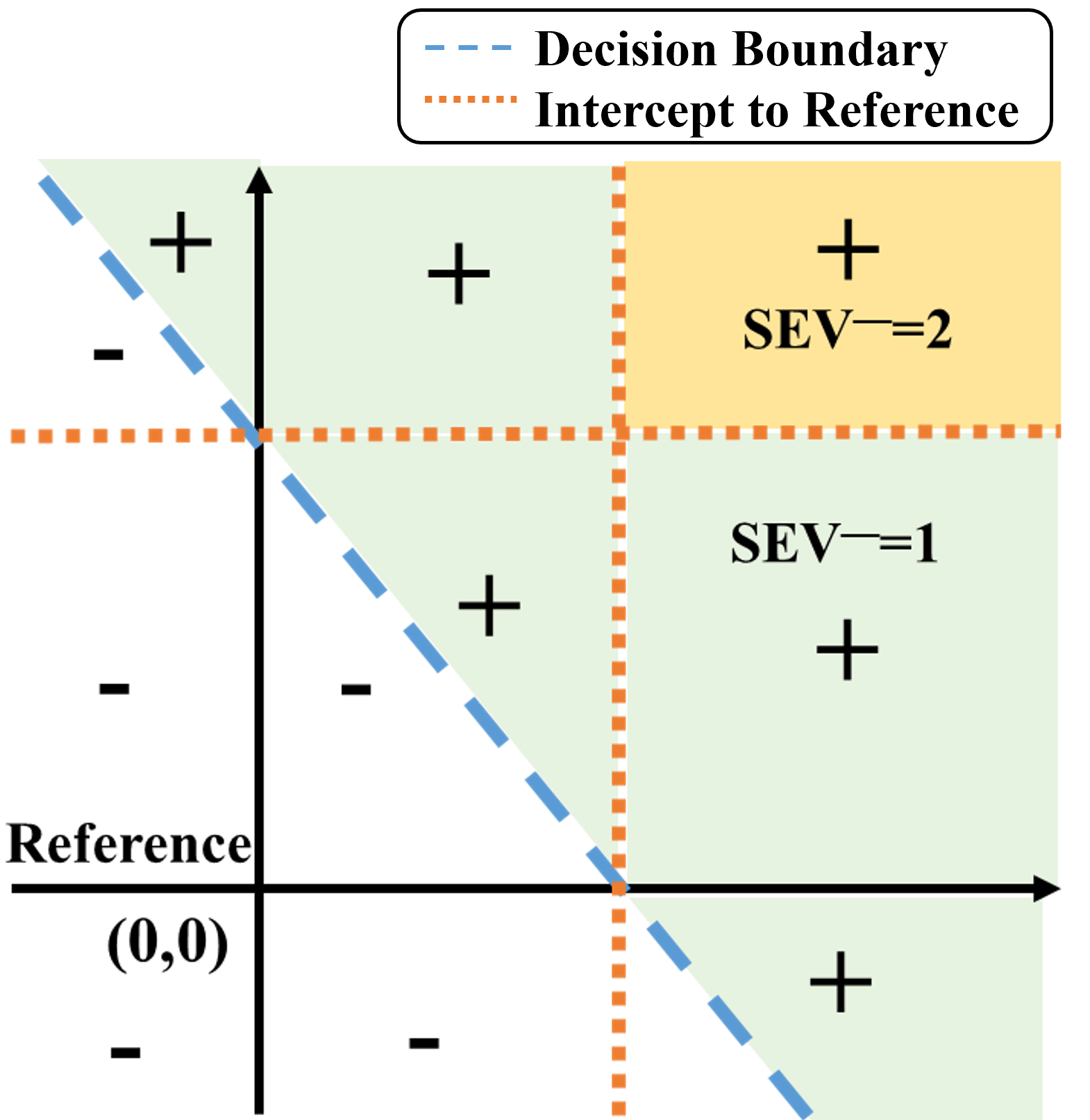}
         \caption{\sev{-} in linear classifier}
         \label{fig:sev_minus}
     \end{subfigure}
        \caption{Visual illustrations of \sev{} definitions}
        \label{fig:sev_definition}
\end{figure}

In real-world applications, some features might be immutable or should not contribute to explanations for legal or ethical reasons. Suppose a bank uses a machine learning model for making credit loan decisions. The bank may want to ensure that the explanation for a denied loan does not involve age since no one can adjust their age. We can accommodate this via a straightforward extension of \sev{-}, which we call the Restricted Sparse Explanation Value (\sev{}$^\circledR$):

\begin{definition}[Restricted SEV, denoted \sev{}$^\circledR$]
\label{def:restricted_sev}
For a query, predicted as positive ($f(\bx^{\q})=1$), and a set of restricted features $\mathcal{S}_r \subseteq \{1, 2, \dots, p\}$, the Restricted Sparse Explanation Value (\sev{}$^\circledR$) is the length of the shortest pathway from the sample's feature values to a negatively predicted vertex, considering only non-restricted features, i.e.,
\begin{equation}
\begin{aligned}
\min_{\bb \in \{0,1\}^p} & \quad \|\mathbf{1}-\bb\|_0 \quad \\
\textrm{s.t.} & \quad \mathcal{L}_{f, \q}(\bb)=0,  \quad \textrm{ where } b_j = 1  \;\forall j \in \mathcal{S}_r.
\end{aligned}
\end{equation}
\end{definition}


\sev{+}, \sev{-}, and \sev{\circledR} can all be computed by a breadth-first search along the \sev{} hypercube. Since we care about sparse and faithful explanations (i.e., no more than 5-7 features) we can set up search depth limits, which also have the effect of lowering the computational time. SEVs are usually very low in practice, and thus can generally be computed in milliseconds. Detailed results are shown in Section \ref{subsec:low_sev}. Appendix \ref{sec:sev_scability} also shows that SEV requires equal or less time to generate explanations than other methods.

\section{Optimizing Sparse Explanation Values}
\label{sec:methods}

In this section, we describe how to obtain classifiers that optimize the average \sev{-} or \sev{+}. As we mentioned, calculating SEVs is theoretically a combinatorial problem. Thus directly optimizing it is intractable. Therefore, we propose some easy-to-optimize surrogate objectives that obtain classifiers with very low SEVs in practice. 

\subsection{Volume-based SEV$^+$ Optimization for Linear Models}
Consider a linear classifier, $f(\bx):= \mathbf{1}[(\beta_0 + \sum_{j=1}^p \beta_j x_j)>0]$. In this case, there exists an analytical surrogate objective for optimizing \sev{+}. Given this classifier, the positively classified region of the space $\{\bx \in \mathbb{R}^p | f(\bx)=1\}$ can be partitioned into regions with \sev{+}=~1, \sev{+}=~2, ..., \sev{+}=~$p$. Approximating the data to be uniformly distributed, we can minimize the average \sev{+} by minimizing the volume of the regions with SEV${}^{+}\geq 2$.

\begin{theorem}\label{thm:volume}
Consider a linear classifier, $
    f(\bx):= \mathbf{1}[(\beta_0 + \sum_{j=1}^p \beta_j x_j)>0]
$, where $\forall j$, $\beta_j\neq 0$, and for reference $\tilde{\bx}$, we have $f(\tilde \bx) = 0$ (i.e., reference predicted as negative). Let $g^{\textrm{\rm ref}}(\boldsymbol{\beta}) = \beta_0+\sum_{j=1}^{p} \beta_{j}\tilde{x}_j$ be the raw score of the classifier $f$ at reference $\tilde \bx$. For all $k\in\{2,3,\cdots,p\}$, the volume of region $V_k$ in the input space $\mathbb{R}^p$ with SEV${}^{+}= k$ is: 
\begin{equation}
V_k=c_k \cdot \prod_{j=1}^{p} \left\lvert \frac{g^{\textrm{\rm ref}}(\boldsymbol{\beta})}{\beta_j}\right\rvert,
\end{equation}
where $c_k$ is a finite constant unrelated to the $\beta$'s.
\end{theorem}
See the proof in Appendix \ref{sec:theorem41_proof}. 

If some $\beta_j$'s are zero, we can either calculate volume on the subspace where the $\beta_j$'s are non-zero, or add an $\epsilon$ to the terms in the denominator.
According to Theorem \ref{thm:volume}, the volume of the region with SEV${}^{+}\geq 2$ is proportional to the volume of a hyperrectangle, given by $\prod_{j=1}^{p} |\frac{g^{\textrm{ref}}(\boldsymbol{\beta})}{\beta_j}|$. Therefore, to minimize the expected SEV in the region, we minimize the volume of this hyperrectangle. For better optimization stability, we optimize the logarithm of the volume, i.e.,
\begin{equation} \label{eqn:volume_loss}
    \min_{\beta_0,\beta_1,\cdots,\beta_p} \frac{1}{p}\sum_{j=1}^{p} \log\left(\min\left(\left|\frac{g^{\textrm{\rm ref}}(\boldsymbol{\beta})}{\beta_j}\right|,\epsilon\right)\right)
\end{equation}
where $\epsilon$ is used to ensure numerical stability. 

This volume-based SEV loss, which we call \textbf{\volopt{}}, is a simple and theoretically justified method that is effective for optimizing the \sev{+} of linear models. However, it is limited to linear models and cannot be applied to more complex models. In the next subsection, we introduce individual-based SEV loss, which can be applied to other types of models as well.

\subsection{Individual-based SEV Optimization}
The volume-based SEV loss above approximates queries as arising uniformly over $\mathcal{X}$, which might not be true. Making the less restrictive assumption that the distribution of a training dataset is close to that of test data, we can instead focus on minimizing the average training SEV. The difficulty here lies in the combinatorial number of steps required to calculate SEV. However, we have found that most trained classifiers yield low average SEVs (roughly $<3$) on real datasets. With the majority of training data having low SEV, we can design the optimization objective to penalize queries with \sev{+} not equal to 1. On the other hand, we maximize the fraction of points where \sev{+} is exactly 1. 
Given a class $\mathcal{F}$ of classification models that output the probability of belonging to the positive class, the loss term for minimizing \sev{+} on a training set is:
\begin{equation}\label{eqn:individual_loss}
    \begin{aligned}
        &\ell_{\text{SEV\_All\_Opt}+} (g) :=\\ &-\frac{1}{n^{+}}\sum_{i=1}^{n^{+}}\min\left(\max_{j =1, \dots, p} g(\be_j\odot \bx_{i} + (\mathbf{1} - \be_j) \odot \tilde{\bx}), ~\mathcal{T}\right)
    \end{aligned}
\end{equation}
where $\bx_{i}$ is the $i$-th query in the training data, $n^{+}$ represents the number of positively predicted queries in the dataset, and $\be_j$ denotes the vector with a 1 in the $j^{th}$ coordinate and 0's elsewhere, and $g:\mathbb{R}^p \to [0,1]$ is the continuous estimator associated with classifier $f$ (e.g., in logistic regression, $g$ is a sigmoid function), and $\mathcal{T}$ is the threshold value for classification, which is usually 0.5 for balanced data. Here, we consider all vertices 1 step away from the reference and see whether at least one of them is predicted as positive (i.e., \sev{+}=1). The min operation taken with the threshold $\mathcal{T}$ is used to fix the value of the objective when the prediction is positive since we do not care where on the positive side of the decision boundary a prediction is, as long as it is positive. 
Similarly, we propose the following loss term for minimizing \sev{-}:
\begin{equation}
\begin{aligned}
    &\ell_{\text{SEV\_All\_Opt}-}(g) := \\ &\frac{1}{n^{+}}\sum_{i=1}^{n^{+}}\max\left(\min_{j = 1, \dots, p}g((\mathbf{1} - \be_j)\odot \bx_i + \be_j \odot \tilde{\bx}),~\mathcal{T} \right)
\end{aligned}
\end{equation}

We call these individual-based SEV losses \textbf{\allopt{+}} and \textbf{\allopt{-}}. The \allopt{-} loss can also be used to target \sev{\circledR} by replacing the minimization over $j = 1, \dots, p$ with a minimization over $j \in \mathcal{S}_r$, which we denote as \textbf{\allopt{\circledR}}. The experiments section will show that these losses are not only effective in shrinking the average SEV but often result in attaining the minimum possible SEV value of 1 for most or all queries. 


One thing to note is that both \allopt{} and \volopt{} assume that the reference is predicted negative; that is, $\mathcal{L}_{f,\q}(\mathbf{0}) = 0$. Thus, if the reference is predicted as positive, then optimization for \sev{+} leads to a trivial solution where \sev{+}$=0$ for all samples. Similarly \sev{-} would not find a shortest path (or any path for that matter) from the observation to the negative class, since the reference is positive.
Thus, to ensure the reference is predicted to be negative, we add a term that penalizes the reference receiving a positive prediction:
\begin{equation}
    \ell_{\text{Pos\_base}}:= \max(g(\tilde \bx), \mathcal{T}-\theta)
\end{equation}
where $\theta>0$ is a margin parameter, for example, $\theta=0.05$. This term is $(\mathcal{T}-\theta)$ as long as the reference is predicted negative. As soon as it exceeds that amount, it is penalized (increasing linearly in $g(\tilde \bx)$).

Combining the loss terms above, we optimize a linear combination of them,
\begin{equation}\label{eqn:totalloss}
    \min_{f\in \mathcal{F}} \ell_{\text{BCE}} + C_1 \ell_{\text{SEV}} + C_2 \ell_{\text{Pos\_base}}
\end{equation}
where $\ell_{\text{BCE}}$ is the Binary Cross Entropy Loss to control the accuracy of the training model. $\ell_{\text{SEV}}$ can be any of the SEV-based terms we introduced, and $\ell_{\text{Pos\_base}}$ is for ensuring the reference is predicted as negative. 
$C_1$ and $C_2$ are the strengths of the terms and can be tuned through cross-validation. Minimizing \eqref{eqn:totalloss} gives a low-SEV model.

\section{Experiments}
\label{sec:experiments}

We present experiments on six real-world datasets: (i) the UCI Adult Income dataset for predicting income levels \citep{adult_income_data, uci}, (ii) the FICO Home Equity Line of Credit Dataset for assessing credit risk, used for the Explainable Machine Learning Challenge \citep{fico}, (iii) the UCI German Credit dataset for determining creditworthiness \citep{german_credit_data, uci}, (iv) the MIMIC-III dataset for predicting patient outcomes in intensive care units \citep{mimic3data,mimic3article}, (v) the COMPAS dataset \citep{propublica,WangHanEtAl2022} for predicting recidivism, and (vi) the Diabetes dataset \citep{diab_data,uci} for predicting if patients will be re-admitted within two years. Additional details on data and preprocessing are in Appendix \ref{subsec:dataset_details}. 

We train four reference binary classifiers: (i, ii) Logistic Regression classifiers with $\ell_1$ (\textbf{L1 LR}) and $\ell_2$ (\textbf{L2 LR}) penalties, (iii) a Gradient Boosting Decision Tree classifier (\textbf{GBDT}), and (iv) a 2-layer Multi-Layer Perceptron (\textbf{MLP}). In addition, we train ``optimized'' variants of these models, in which the SEV penalties described in the previous sections are implemented. Details on training the methods are in Appendix \ref{sec:app:model_training}. Below, we report the average test SEV; i.e. averaged across queries in the test set.


\subsection{Most models already have low SEVs.}
\label{subsec:low_sev}
We first compute \sev{+} and \sev{-} for our four reference classifiers. 
We emphasize that these models are not optimized for SEV, they were created with standard machine-learning algorithms. All algorithms were run for 10 different train-test splits and the mean and standard deviation across these splits of the average test accuracy, AUC, \sev{+} and \sev{-} on the German Credit data are shown in Table \ref{tab:german_sev_non_optimized}. Results for the other datasets are listed in Appendix \ref{sec:appendix_results}.


\begin{table}[!ht]
\centering
\small
\caption{SEV for reference classifiers in German Credit}
\label{tab:german_sev_non_optimized}
\setlength\tabcolsep{3pt}
\scalebox{1}{%
\begin{tabular}{ccccc}
\hline
\textbf{Model} & \textbf{L1 LR} & \textbf{L2 LR} & \textbf{MLP} & \textbf{GBDT}\\ \hline
\textbf{\makecell{Acc}} & \textcolor{gray}{$0.74\pm0.02$} & \textcolor{gray}{$0.73\pm0.03$} & \textcolor{gray}{$0.76\pm0.03$} & \textcolor{gray}{$0.75\pm0.03$}\\\rule{0pt}{2ex}    
\textbf{\makecell{AUC}} & \textcolor{gray}{$0.77\pm0.03$} & \textcolor{gray}{$0.76\pm0.03$} & \textcolor{gray}{$0.77\pm0.08$} & \textcolor{gray}{$0.78\pm0.02$}\\\rule{0pt}{2ex}    
\textbf{\sev{+}} & $1.00\pm0.00$ & $1.00\pm0.00$ & $1.04\pm0.08$ & $1.10\pm0.24$ \\
\textbf{\sev{-}} & $1.29\pm0.14$ & $1.74\pm0.16$ & $1.58\pm0.22$ & $1.58\pm0.22$\\ \hline
\end{tabular}}
\end{table}

 Even though the German Credit models for L2LR, MLP and GBDT used up to all 20 features (they are not sparse), we observe from Table \ref{tab:german_sev_non_optimized} that most of the queries need only to align one reference feature value with the query value to change to a positive prediction (\sev{+}), and we need to align fewer than 2 feature values from the query to the reference for a negative prediction (\sev{-}). Thus, \textit{even without optimization for SEV at all, most models already have sparse explanations}. However, there is still room for improvement, as we now show.
 

\subsection{SEV can be reduced with no loss in performance.}
While existing models already have low SEVs on real datasets, our SEV optimization algorithm proposed in the previous section can further decrease SEVs -- creating accurate models with extreme decision sparsity.
To illustrate, Figure \ref{fig:sev+comparison} shows the impact on \sev{+} of \allopt{} and \volopt{} on a reference linear classifier trained on the Adult dataset. Figure \ref{fig:adult_lr_sev_freq} shows the SEV distribution across queries; while L1 LR and L2 LR have reasonably low SEVs (per the discussion in the previous subsection), both of the proposed optimization algorithms are able to find models for which the \sev{+} of \emph{all} queries is 1 (the minimum possible SEV). Table \ref{tab:lr_adult_perf} in the Appendix shows that this incurs negligible performance loss.  Figure \ref{fig:adult_lr_sankey} shows how the optimization affects the \sev{+} of each query. Its biggest impact on the mean is from reducing the explanations of many queries from involving 2 to 5 terms to a single one, with no query's explanation becoming more complicated. Additionally, the predicted class distribution is comparable, and the impact of the optimization is not simply to predict a greater number of negative labels; it actually targets the explanation sparsity. 

\begin{figure}[!ht]
     \centering
     \begin{subfigure}[b]{0.238\textwidth}
         \centering
         \includegraphics[width=\textwidth]{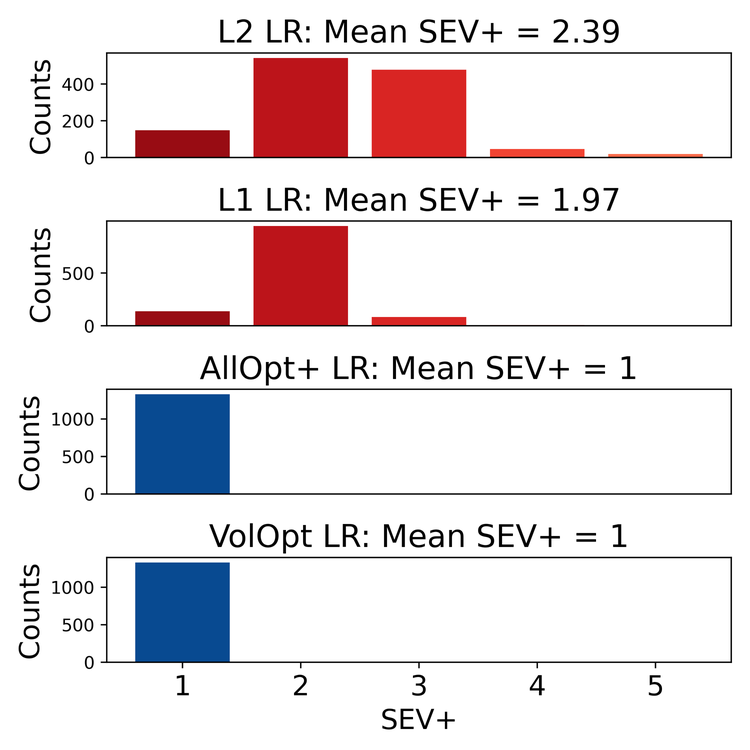}
         \caption{\sev{+} counts across methods}
         \label{fig:adult_lr_sev_freq}
     \end{subfigure}
     \hfill
     \begin{subfigure}[b]{0.238\textwidth}
         \centering
         \includegraphics[width=\textwidth]{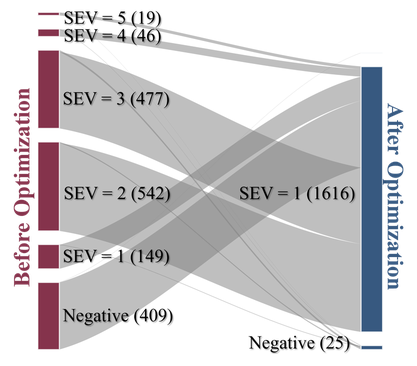}
         \caption{\sev{+} Optimization Result}
         \label{fig:adult_lr_sankey}
     \end{subfigure}
        \caption{\sev{+} performance for linear classifiers in Adult}
        \label{fig:sev+comparison}
\end{figure}


\allopt{+} and \allopt{-} can also be applied to nonlinear models. To illustrate, we apply them to baseline \textbf{MLP} and \textbf{GBDT} classifiers, again using the Adult data. For \textbf{GBDT}, we use our methods to fine-tune the weights for each tree generated by the original method. Both are optimized by Stochastic Gradient Descent. Figure \ref{fig:adult_nonlinear_sev_freq} shows that across both model classes, and for both \sev{-} and \sev{+}, \allopt{-} and \allopt{+} are able to reduce the SEV of all queries to 1, as in the linear case. Appendix \ref{sec:appendix_results} has other performance metrics, as well as results for other datasets, showing that we can consistently reduce SEV with negligible loss in accuracy, across various model classes and datasets.

\begin{figure}[!ht]
     \centering
     \begin{subfigure}[b]{0.238\textwidth}
         \centering
         \includegraphics[width=\textwidth]{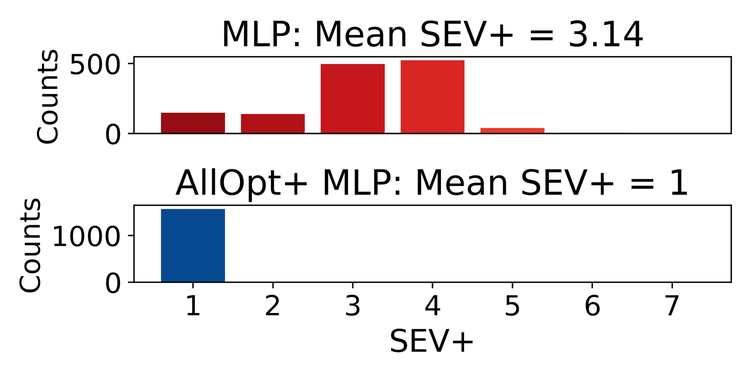}
         \caption{\allopt{+} in MLP}
         \label{fig:adult_mlp_plus}
     \end{subfigure}
     \hfill
     \begin{subfigure}[b]{0.238\textwidth}
         \centering
         \includegraphics[width=\textwidth]{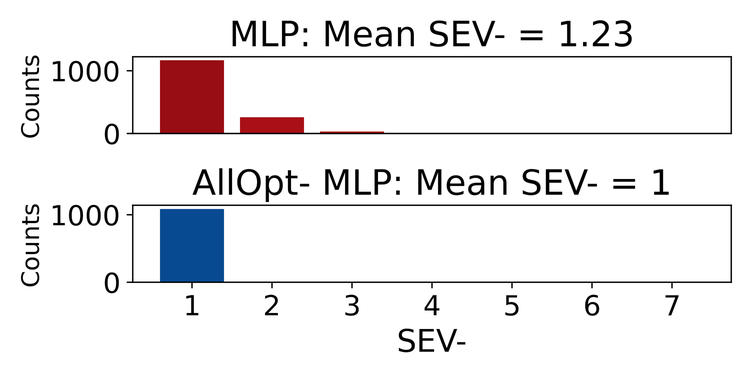}
         \caption{\allopt{-} in MLP}
         \label{fig:adult_mlp_minus}
     \end{subfigure}
      \begin{subfigure}[b]{0.238\textwidth}
         \centering
         \includegraphics[width=\textwidth]{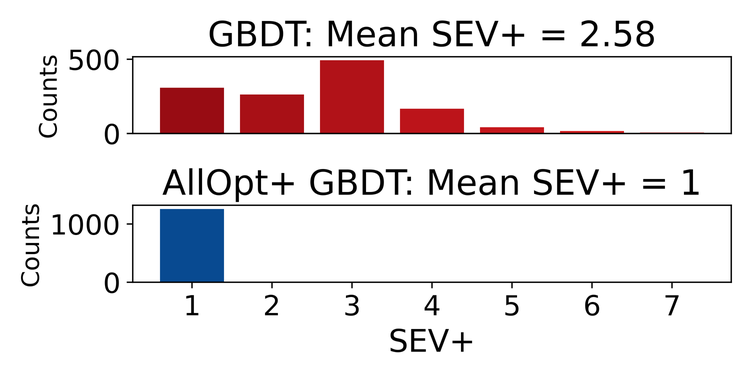}
         \caption{\allopt{+} in GBDT}
         \label{fig:adult_gbdt_plus}
     \end{subfigure}
     \hfill
     \begin{subfigure}[b]{0.238\textwidth}
         \centering
         \includegraphics[width=\textwidth]{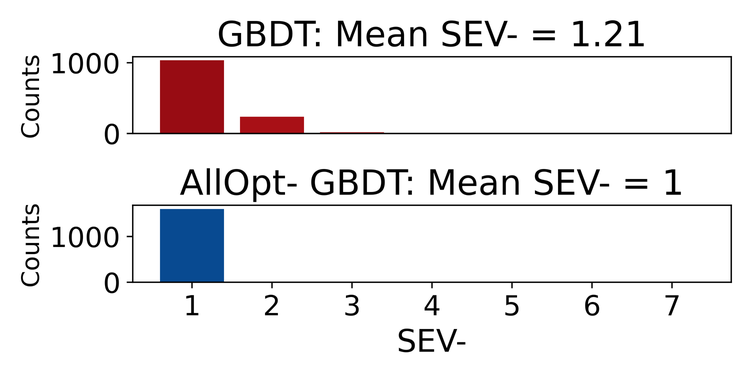}
         \caption{\allopt{-} in GBDT}
         \label{fig:adult_gbdt_minus}
     \end{subfigure}
     \caption{Performance of \allopt{+} and \allopt{-} in Adult}
     \label{fig:adult_nonlinear_sev_freq}
\end{figure}

Moreover, to evaluate the time consumption of our methods, we have tried different dataset sizes. The detailed information for each dataset is shown in Appendix \ref{subsec:dataset_details}. To better compare the training time consumption, we have used \textbf{MLP} as the baseline model and compared its running time with and without \allopt{} for different datasets. All the models were run 10 times and we report the mean and standard deviation in Table \ref{tab:time_compare_optimization}.

\begin{table}[ht]
\centering
\small
\begin{tabular}{cccc}
\hline
\textbf{Dataset}& \textbf{Baseline} & \textbf{\allopt{+}} & \textbf{\allopt{-}} \\ \hline
COMPAS&$ 10.5\pm 0.69$&$26.9\pm 1.06$&$ 21.7\pm 0.03$\\
Adult&$61.1\pm 0.15$&$ 164\pm 1.56$&$ 136\pm 1.22$\\
MIMIC III&$78.2\pm 1.09 $&$ 207\pm 2.71$&$ 167\pm 2.42$\\
German Credit &$1.81\pm 0.01$&$5.01\pm 0.01$&$ 4.08\pm 0.04$\\
FICO&$16.9\pm 0.78$&$ 47.6\pm  0.34$&$ 37.4\pm 1.30$\\
Diabetes&$188\pm 3.08$&$ 606\pm 3.65$&$ 517\pm 3.31$\\
\hline
\end{tabular}
\caption{The training time (seconds) comparison with and without \allopt{} in MLP}
\label{tab:time_compare_optimization}
\end{table}

Based on the run time results shown above, we can observe that even when optimizing a complex black box model, the increase in time consumption associated with optimizing SEV is quite low.

\subsection{It is unnecessary to have global sparsity to have sparse explanations.}
\label{subsec:sex_complexity}


In this section, we discuss whether global sparsity is needed for sparse local explanations.
We use the proportion of zero coefficients for linear models as a global sparsity measure. We will use the Adult income dataset as an example to see how the model distributes coefficients.


\begin{table}[!ht]
\centering
\small
\caption{SEV and zero-coefficient \% of LR in Adult}
\label{tab:sparsity_sev}
\setlength\tabcolsep{3pt}
\scalebox{1}{%
\begin{tabular}{ccccc}
\hline
 \textbf{Methods} & \textbf{L1 LR} & \textbf{L2 LR} & \textbf{\makecell{\allopt{+}\\ LR}} & \textbf{\makecell{\volopt{}\\ LR}} \\ \hline
\makecell{Mean \\\sev{+}} & $1.95$$\pm0.01$ & $2.30$$\pm0.08$ & $1.00$$\pm0.00$ & $1.00$$\pm0.00$ \\
\begin{tabular}[c]{@{}c@{}}\% of Zero \\ Coefficients\end{tabular} & $87.9$$\pm0.00$ & $0.09$$\pm0.00$ & $0.00$$\pm0.00$ & $0.00\pm0.00$\\ \hline
\end{tabular}}

\end{table}
The \textbf{L1 LR}, \textbf{L2 LR}, \textbf{\allopt{+} LR}, and \textbf{\volopt{} LR} methods are all compared in Table \ref{tab:sparsity_sev}. L1 LR has already attained low SEVs, possibly because its models are globally sparse. Nevertheless, even without global sparsity, L2 penalized linear regression also generates a low SEV model. This suggests that if our objective is to construct models with low SEVs, we should not just focus on producing globally sparse models, as this may be unnecessarily restrictive. Moreover, after \sev{+} optimization, neither the \allopt{+} LR nor the \volopt{} LR optimization approaches produce sparse models and yet have perfect SEV's. Thus, it is not necessary to have global sparsity to achieve sparse explanations.


\subsection{Explaining a fixed model: comparing SEV to other local explanation methods}
\label{subsec:sev_xai}

Here, we use GBDT to generate a model on the COMPAS dataset, and compare \sev{-} with various post hoc explanation methods, specifically: SHAP-C (using treeSHAP and kernelSHAP) \citep{shapley,ramon2020comparison}, LIME-C \citep{lime,ramon2020comparison}, and Diverse Counterfactual Explanations (DiCE) \citep{dice}. LIME and SHAP are post-hoc explainers that provide local feature importance scores to the prediction. The idea of SHAP-C and LIME-C is to order the features by importance, and based on that ordering, progressively (from the most important feature to less important ones) align the queries' features with the reference values. We then calculate the number of alignments needed for the point to change from being positively predicted to negatively predicted (shown as Flip Number in Figure \ref{fig:local_comparison_compas}). We also compare our methods to DiCE, counting the number of features it modifies to generate a counterfactual explanation.

\begin{figure}[!ht]
    \centering
    \begin{subfigure}[b]{0.35\textwidth}
    \centering
    \includegraphics[width=\textwidth]{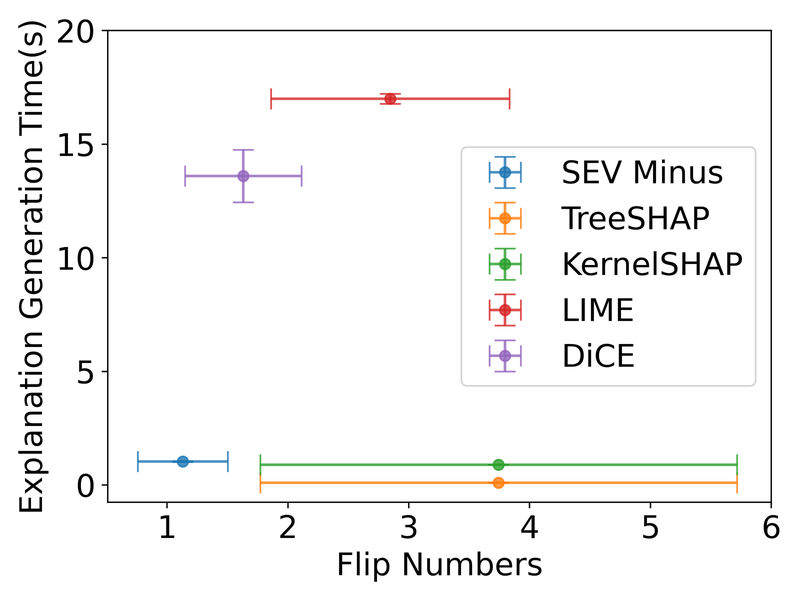}
    \caption{Flip Numbers and Mean Explanation Time Consumption}
    \label{fig:flip_compas}
    \end{subfigure}
    \hfill
    \begin{subfigure}[b]{0.35\textwidth}
    \centering
    \includegraphics[width=\textwidth]{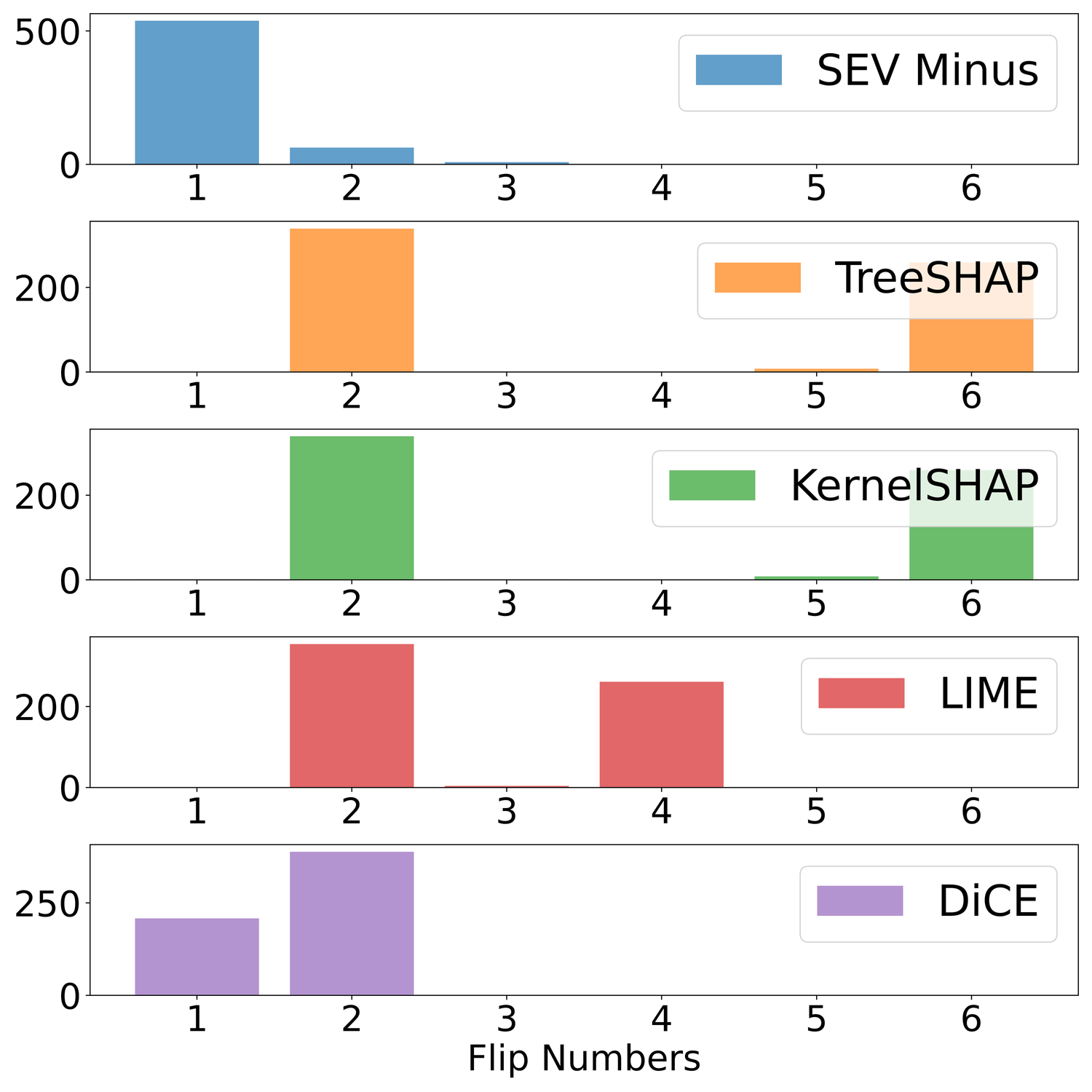}
    \caption{Flip Number Distributions}
    \label{fig:flip_xai}
    \end{subfigure}
    \caption{Local Explanation Methods' Performance Comparison in COMPAS}
    \label{fig:local_comparison_compas}
\end{figure}

Figure \ref{fig:local_comparison_compas} shows that, on average, LIME-C and SHAP-C flip about half of the features in the model (which uses 7 features in total) and do not induce sparse explanations. This conclusion aligns with \citet{fernandez2022explaining}'s observation that high locally important features may not always influence model decisions.

Even though DiCE applies backward feature selection after generating explanations to control sparsity \citep{dice}, Figure \ref{fig:local_comparison_compas} shows that \sev{-} provides sparser explanations than DiCE. Moreover, DiCE provides less interpretable explanations than \sev{-}.
Recall that DiCE uses minimal edit distance to generate counterfactual explanations, whereas we find a minimal number of flips between original values and reference values.
As we discussed in Related Works, the edit-distance-based counterfactual explanations may not be as interpretable to humans  \citep{counterfac_img} because the counterfactual often ends up in a low-density region where one cannot determine the class with certainty. The counterfactual explanations generated by DiCE can even lead  \textit{away} from the general direction of the other class. For instance, we found that on a criminal recidivism dataset, DiCE stated that a high-risk criminal offender who commits \textit{more} crimes would be classified as \textit{lower} risk, which is paradoxical. 
This unreasonable explanation arises because DiCE can reference \emph{any} negatively-predicted point, including outliers.
In contrast, the reference point for \sev{} is the center of the negative population, thus the direction of the \sev{} is generally meaningful: towards the negatives.
For example, the reference value used in our approach for the number of prior misdemeanors is 0, which is more natural.

 Table \ref{tab:example_2_compas} shows counterfactual explanations for two different queries. For the first, DiCE suggests that the number of juvenile crimes should go from 1 to 5 to change the prediction from `will recidivate' to `won't recidivate.' This is nonsensical, as the number of crimes should be positively correlated with the propensity to recidivate. In contrast, \sev{-} suggests that the number of prior crimes should go from 2 to 1, which makes sense. 
Similarly, for the second query, DiCE suggests that the number of juvenile crimes should go from 3 to 7 and the number of juvenile misdemeanors from 2 to 11 in order to change the prediction from `will recidivate' to `won't recidivate.' Again, this explanation is nonsensical. In contrast, \sev{-} provides two possible explanations: (i) that the number of prior crimes should go from 3 to 1 or (ii) that the number of juvenile misdemeanors should go from 2 to 0. Again, this makes more sense. Given that DiCE uses a random initialization to find a counterfactual explanation whereas \sev{-} provides the \emph{minimum} number of feature alignments to cross the decision boundary, we would expect \sev{-} to be more stable and to generate useful explanations.

\begin{table*}[!ht]
\centering

\scalebox{0.76}{%
\begin{tabular}{ccccccccc}
\hline
\textbf{Explanations} & \textbf{Sex} & \textbf{Age} & \textbf{\begin{tabular}[c]{@{}c@{}}Juv fel\\ counts\end{tabular}} & \textbf{\begin{tabular}[c]{@{}c@{}}Juv misd\\ count\end{tabular}} & \textbf{\begin{tabular}[c]{@{}c@{}}Juvenile \\ crimes\end{tabular}} & \textbf{\begin{tabular}[c]{@{}c@{}}Priors\\ counts\end{tabular}} & \textbf{\begin{tabular}[c]{@{}c@{}}Current Charge \\ Degree\end{tabular}} & \textbf{\begin{tabular}[c]{@{}c@{}}Predicted\\ Label\end{tabular}} \\ \hline
\textbf{Query 1} & Male & 23 & 0 & 1 & 1 & 2 & Not Felony & 1 \\
\arrayrulecolor{gray}\hline
\textbf{DiCE} & -- & -- & -- & -- & 5 & -- & -- & 0 \\
\textbf{\sev{-}} & -- & -- & -- & -- & -- & 1 & -- & 0 \\ \hline\hline
\textbf{Query 2} & Male & 19 & 1 & 2 & 3 & 3 & Felony & 1 \\
\arrayrulecolor{gray}\hline
\textbf{DiCE} & -- & -- & -- & 11 & 7 & -- & -- & 0 \\
\textbf{\sev{-}}  & -- & -- & -- & 0 & -- & -- & -- & 0 \\
&   --         & -- & -- & -- & -- & 1 & --& 0\\\hline
\end{tabular}}
\caption{DiCE and \sev{-} explanations for two different queries in COMPAS}
\label{tab:example_2_compas}
\end{table*}

In addition, \sev{} has shown competitive speed in generating local explanations. Figure \ref{fig:flip_compas} shows that DiCE takes more time to generate stable sparse counterfactual explanations than \sev{-}. Table \ref{fig:local_explanation_time} shows the detailed time required to generate a local explanation for our method and baselines SHAP, LIME, and DiCE, for several datasets. We use \textbf{GBDT} as a baseline model and evaluate the median explanation runtime on queries in the test set. The table shows that \sev{-} has a reasonable running time for black-box models on most datasets. We can observe that the explanation time is better than DiCE in all the datasets since it only flips from feature values to the reference. The reason why the FICO Dataset takes a long time is that its average \sev{-} is about 4, which means that the search in the \sev{} hypercube starting from the query is too deep. However, if the time consumption for generating explanations is too long, it means that this model cannot be sparsely explained, but we can use \allopt{} or \volopt{} methods to optimize our model. Importantly, other types of explanations (SHAP, LIME, DiCE) do not have optimization methods such as \allopt{} or \volopt{}. The detailed comparison between KernelSHAP and \sev{-} is shown in Appendix \ref{sec:sev_scability}.

\begin{table*}[ht]
\centering
\scalebox{0.9}{
\begin{tabular}{cccccc}
\hline
\textbf{Dataset} & \textbf{TreeSHAP} & \textbf{KernelSHAP} & \textbf{LIME} & \textbf{DiCE} & \textbf{\sev{-}} \\ \hline
Adult & $1.08\pm0.00$ & $4.82\pm0.01$ & $20.27\pm0.22$ & $43.41\pm0.16$ & $1.39\pm0.00$ \\
COMPAS & $0.50\pm0.01$ & $0.22\pm0.00$ & $2.68\pm0.10$ & $20.86\pm0.39$ & $0.19\pm0.02$ \\
MIMIC & $1.13\pm0.01$ & $17.30\pm0.10$ & $6.10\pm0.05$ & Crashed & $0.45\pm0.03$ \\
\begin{tabular}[c]{@{}c@{}}German\\ Credit\end{tabular} & $0.53\pm0.00$ & $9.70\pm0.17$ & $7.11\pm0.07$ & $26.70\pm0.91$ & $1.37\pm0.05$ \\
FICO & $0.56\pm0.02$ & $8.25\pm0.27$ & $6.05\pm0.14$ & $27.82\pm0.05$ & $15.75\pm1.01$ \\
Diabetes & $0.55\pm0.01$ & $9.42\pm0.10$ & $22.49\pm0.26$ & $34.66\pm1.33$ & $2.24\pm0.01$ \\ \hline
\end{tabular}}
\caption{The median explanation runtime (in $10^{-2}$ s) for query units predicted as positive. The values in the table differ from those in Figure \ref{fig:flip_compas} since Figure \ref{fig:flip_compas} reports the mean runtime instead of median.}
\label{fig:local_explanation_time}
\end{table*}

\subsection{Restricted SEV in practice}
\label{subsev:restrict_sev}

In real applications, some features are not changeable, such as race, gender, or age. To evaluate \sev{\circledR} and our optimization algorithm under restricted feature sets, we apply our loss function to COMPAS and set \texttt{gender} and \texttt{age} as restricted features when training a linear classifier.

\begin{figure}[th]
\centering
\small
\begin{subfigure}[b]{0.5\textwidth}
\caption{\sev{\circledR} Difference before and after optimization}
\label{tab:sev_r_difference}
\centering
\scalebox{1}{%
\begin{tabular}{lcc}
\hline
 & \textbf{\begin{tabular}[c]{@{}c@{}}Before \\ Optimization\end{tabular}} & \textbf{\begin{tabular}[c]{@{}c@{}}After\\ Optimization\end{tabular}} \\ \hline
\textbf{Test Acc} & \textcolor{gray}{$0.68\pm0.01$} & \textcolor{gray}{$0.65\pm0.01$} \\
\textbf{Test AUC} & \textcolor{gray}{$0.73\pm0.01$} & \textcolor{gray}{$0.72\pm0.01$} \\
\textbf{Mean \sev{\circledR}} & $3.20\pm0.26$ & $1.28\pm0.38$ \\
\textbf{Unexplained (\%)} & $34.40\pm4.40$ & $3.35\pm6.20$ \\ \hline
\end{tabular}
}
\end{subfigure}
\hfill
\begin{subfigure}[b]{0.5\textwidth}
\centering
\includegraphics[width=\textwidth]{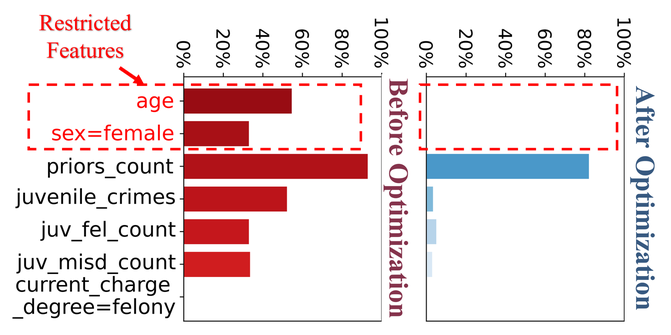}
\caption{Feature Prevalence in Explanations}
\label{fig:explanations_features}
\end{subfigure}

\caption{The performance of \allopt{\circledR} on COMPAS data}
\label{fig:sev_r_performance}

\end{figure}

In this restricted setting, there may exist queries whose predictions do not flip even if all nonrestricted features are flipped. We call these queries ``unexplainable.'' Figure \ref{tab:sev_r_difference} shows the proportion of unexplainable queries in the restricted setting and their relevant \sev{\circledR} before and after restriction and optimization of the \allopt{\circledR} objective. Figure \ref{fig:explanations_features} shows the proportion of queries using each feature for sparse explanations before (Red) and after restriction and optimization (Blue). After introducing restricted features, over one-third of the queries are unable to be explained by the baseline model. However, after \allopt{\circledR} is applied, the percentage of unexplainable queries decreases to 3\% without much performance loss. Moreover, Figure \ref{fig:explanations_features} shows that after applying \allopt{\circledR} with \texttt{age} and \texttt{sex=female} as restricted features, \texttt{prior\_count} becomes the most important factor for explanations for almost all queries, which is more actionable.
In real cases, we might want to replace this feature with a dynamic feature, e.g., the number of priors over the last 5 years, to allow individuals to reduce that feature over time through their actions. In that case, one's explanation would often be simply: \textit{you are predicted as high risk because you committed a crime in the last 5 years}. 
More results on the performance of \allopt{\circledR} are shown in Appendix \ref{sec:sev_r_explanations}. 
These results demonstrate that  \allopt{\circledR} works well in restricted cases.





\section{Discussion and Limitations}

In this work, we introduce the Sparse Explanation Value (SEV) as a new way to measure and optimize interpretability in machine learning models. SEV measures decision sparsity, which focuses on how simply predictions can be explained, instead of global sparsity. Because users care about explanations for their individual predictions (rather than everyone else's predictions), SEV better connects models to their real-world applications. Crucially, we have found (i) that many current models naturally have low SEV, (ii) that global sparsity is not necessary for decision sparsity, and (iii) that SEV can be directly optimized without sacrificing accuracy. The use of interpretable reference values and the development of optimization algorithms (\volopt{} and \allopt{}) further highlight SEV's potential in applications.

There are several limitations to our approach. One is that causal relationships may exist among features, invalidating certain transitions across the \sev{} hypercube. This can be addressed by adapting the definition of vertices to change features in tandem, though it requires knowledge of the causal graph. Another practical consideration is choosing an appropriate reference and SEV variant. However, the flexibility in reference and variant choice actually helps tailor analyses to specific problems. For example, individuals or businesses might consider \sev{+} to understand how they can make their profiles more appealing to lenders. In contrast, examining \sev{-} can help lenders understand the fragility of their approvals and can provide useful information about important variables. Lastly, \sev{\circledR}, which excludes unchangeable characteristics, offers more actionable explanations to users. However, restricting explanations may obscure biases inherent in the model since the excluded features cannot contribute to explanations. 

Ultimately, SEV shifts the burden for interpretability in machine learning models from prioritizing sparse models to prioritizing sparse decisions. SEV is distinct from existing black-box explanation methods, can be computed for pre-trained models, and can be optimized via efficient optimization algorithms. These qualities make it a  candidate for real-world modeling pipelines. 

\newpage
\bibliography{aaai24}

\section*{Checklist}

 \begin{enumerate}

 \item For all models and algorithms presented, check if you include:
 \begin{enumerate}
   \item A clear description of the mathematical setting, assumptions, algorithm, and/or model. [Yes]
   \item An analysis of the properties and complexity (time, space, sample size) of any algorithm. [Yes]
   \item (Optional) Anonymized source code, with specification of all dependencies, including external libraries. [Yes]
 \end{enumerate}

 \item For any theoretical claim, check if you include:
 \begin{enumerate}
   \item Statements of the full set of assumptions of all theoretical results. [Yes]
   \item Complete proofs of all theoretical results. [Yes]
   \item Clear explanations of any assumptions. [Yes]     
 \end{enumerate}

 \item For all figures and tables that present empirical results, check if you include:
 \begin{enumerate}
   \item The code, data, and instructions needed to reproduce the main experimental results (either in the supplemental material or as a URL). [Yes]
   \item All the training details (e.g., data splits, hyperparameters, how they were chosen). [Yes]
         \item A clear definition of the specific measure or statistics and error bars (e.g., with respect to the random seed after running experiments multiple times). [Yes]
         \item A description of the computing infrastructure used. (e.g., type of GPUs, internal cluster, or cloud provider). [Yes]
 \end{enumerate}

 \item If you are using existing assets (e.g., code, data, models) or curating/releasing new assets, check if you include:
 \begin{enumerate}
   \item Citations of the creator If your work uses existing assets. [Yes]
   \item The license information of the assets, if applicable. [Yes]
   \item New assets either in the supplemental material or as a URL, if applicable. [Yes]
   \item Information about consent from data providers/curators. [Not Applicable]
   \item Discussion of sensible content if applicable, e.g., personally identifiable information or offensive content. [Not Applicable]
 \end{enumerate}

 \item If you used crowdsourcing or conducted research with human subjects, check if you include:
 \begin{enumerate}
   \item The full text of instructions given to participants and screenshots. [Not Applicable]
   \item Descriptions of potential participant risks, with links to Institutional Review Board (IRB) approvals if applicable. [Not Applicable]
   \item The estimated hourly wage paid to participants and the total amount spent on participant compensation. [Not Applicable]
 \end{enumerate}

 \end{enumerate}

}\fi

\clearpage
\appendix
\onecolumn
\ifappendix{
\section{Datasets Description}
\label{subsec:dataset_details}
All datasets were partitioned into training and test subsets according to an 80-20 stratified split. Numeric features were linearly rescaled to have mean zero and variance one. Categorical features with $k$ unique levels were one-hot encoded into $k$ dummy variables. Binary features were one-hot encoded into a single dummy variable. Dataset sizes pre- and post-encoding are displayed in Table \ref{tab:datasets}. The reference for features is given by the feature mean for continuous features and the mode for binary and categorical features.

\begin{table}[!ht]
\centering
\setlength\tabcolsep{3pt}
\begin{tabular}{cccc}\toprule
           & Observations & \makecell{Pre-Encoded \\ Features} & \makecell{Post-Encoded \\ Features} \\\midrule
COMPAS & 6,907 & 7 & 7  \\
Adult & 32,561 & 14 & 107  \\
MIMIC-III & 48,786 & 14 & 14 \\
Diabetes & 101,766 & 33 & 101 \\
\makecell{German\\ Credit} & 1,000 & 20 & 59 \\
FICO & 10,459 & 23 & 23 \\
\bottomrule
\end{tabular}
\caption{Training Dataset Sizes}
\label{tab:datasets}
\end{table}
\subsection*{COMPAS}
The COMPAS dataset consists of data on criminal recidivism in Broward County, Florida \citep{propublica}. The objective is to predict two-year recidivism using the following features: sex, age, number of priors, number of juvenile felonies/misdemeanors/crimes, and whether the current charge is a felony. 

\subsection*{Adult}
The Adult data is based on U.S. census data including demographics, education and job information, marital status, and capital gain/loss \citep{adult_income_data, uci}. The target outcome is whether an individual's salary is above \$50,000.

\subsection*{MIMIC-III}
MIMIC-III is a database containing a variety of medical information for patients during ICU stays in the Beth Israel Deaconess Medical Center \citep{mimic3data,mimic3article}. We take the `hospital expires flag', a binary indicator of whether or not a patient died in the given hospitalization, to be the outcome of interest and select the following subset of variables as features:  
\texttt{age}, \texttt{preiculos} (pre-ICU length of stay), \texttt{gcs} (Glasgow Coma Scale), \texttt{heartrate\_min}, \texttt{heartrate\_max}, \texttt{meanbp\_min} (min blood pressure), \texttt{meanbp\_max} (max blood pressure), \texttt{resprate\_min}, \texttt{resprate\_max}, \texttt{tempc\_min}, \texttt{tempc\_max}, \texttt{urineoutput}, \texttt{mechvent} (whether the patient is on mechanical ventilation), and \texttt{electivesurgery} (whether the patient had elective surgery).

\subsection*{Diabetes}

The Diabetes dataset is derived from 10 years (1999-2008) of clinical care at 130 US hospitals and integrated delivery networks \citep{diab_data,uci}. It comprises over 50 features representing patient and hospital outcomes. The dataset includes attributes such as \texttt{race}, \texttt{gender}, \texttt{age}, \texttt{admission type}, \texttt{time in hospital}, the \texttt{medical specialty of the admitting physician}, \texttt{number of lab tests performed}, \texttt{number of medications}, etc. We take whether the patient will \texttt{revisit} the hospital in 2 years as a binary indicator.

\subsection*{German Credit}
The German Credit data \citep{german_credit_data, uci} uses financial and demographic markers (checking account status, credit history, employment/marital status, etc.) to predict whether an individual is at risk for defaulting on a loan. 
\subsection*{FICO}
The FICO Home Equity Line of Credit (HELOC) dataset used for the Explainable Machine Learning Challenge contains a variety of financial markers (number of inquiries into a user's account, max delinquency, number of satisfactory trades, etc.) for various loan applicants \citep{fico}. The target outcome is whether a consumer was ever 90+ days past due in any 2-year period since they opened their account.  

\section{Model Training}
\label{sec:app:model_training}
Baseline models were fit using \texttt{sklearn} \citep{sklearn} implementations in \textsf{Python}. The logistic regression models L1 LR and L2 LR were fit using regularization parameter $C = 0.01$. The 2-layer MLP used ReLU activation and consisted of two fully-connected layers with 128 nodes each. It was trained with early stopping. The gradient-boosted classifier used 200 trees with a max depth of 3. 

The optimized models were trained by adding the SEV losses from Section \ref{sec:methods} to the standard loss term for the models. For GBDT, the training goal is to reweight the trees from the reference GBDT model. The resulting loss was minimized via gradient descent in \texttt{PyTorch} \citep{pytorch}, with a batch size of 128, a learning rate of 0.1, and the Adam optimizer. To maintain high accuracy, the first 70 training epochs are warm-up epochs optimizing just Binary Cross Entropy Loss for classification (\texttt{BCELoss}). The next 30 epochs add the \allopt{} and \volopt{} terms and the reference positive penalty term to encourage low \sev{} values.

\clearpage
\section{Model Results}
\label{sec:appendix_results}

In this section, we compare our optimized classifiers to the reference classifiers in terms of accuracy/AUC and \sev{+}/\sev{-}.

\subsection{Results Overview}
\label{subsec:results_overview}

Here we report training and testing accuracy, AUC, mean \sev{+}, and mean \sev{-}. Table entries are means and standard deviations across 10 random train-test splits for each dataset. The optimized mean \sev{+} and \sev{-} are marked in \textcolor{red}{red} from Table \ref{tab:lr_adult_perf} to Table \ref{tab:gbdt_diabetes_perf}. More detailed SEV counts for each dataset before and after optimization are shown in Appendix \ref{subsec:sev_distributions}. Blank entries are caused by incompatible loss functions and objectives (e.g., \allopt{+}, designed to target \sev{+}, and \sev{-}). From these tables, we observe that with the application of \allopt{+} and \volopt{} methods for \sev{+} and \allopt{-} for \sev{-}, we can obtain models with lower \sev{+} or \sev{-}  without significant performance loss. This means that we are able to find models with sparse explanations for more queries, without reducing the predictive power of the model. 

\ifshow
\begin{table*}[ht]
\centering
\begin{tabular}{lcccccc}\toprule
& \multicolumn{2}{c}{Train} & \multicolumn{4}{c}{Test}
\\\cmidrule(lr){2-3}\cmidrule(lr){4-7}
           & Accuracy  & AUC & Accuracy    & AUC  & Mean \sev{+} & Mean \sev{-}\\\midrule
L2 & $0.85\pm0.00$ & $0.90\pm0.00$ & $0.85\pm0.02$ & $0.90\pm0.00$ & $2.27\pm0.11$ & $1.18\pm0.15$\\
L1 & $0.85\pm0.00$ & $0.90\pm0.00$ & $0.85\pm0.01$ & $0.90\pm0.00$ & $1.96\pm 0.07$ & $1.13\pm0.01$\\
\allopt{+} & $0.85\pm0.00$ & $0.90\pm0.00$ & $0.85\pm0.01$ & $0.90\pm0.01$ & \textcolor{red}{$1.10\pm0.26$}\\
\allopt{-} & $0.85\pm0.00$ & $0.90\pm0.00$ & $0.84\pm0.01$ & $0.90\pm0.00$ & & \textcolor{red}{$1.03\pm0.02$}\\
\volopt{} & $0.84\pm0.01$ & $0.90\pm0.02$ & $0.84\pm0.01$ & $0.89\pm0.02$ & \textcolor{red}{$1.01\pm0.00$} \\\bottomrule
\end{tabular}
    \caption{SEV Optimization Performance of linear classifiers on Adult}
\label{tab:lr_adult_perf}
\end{table*}

\begin{table*}[ht]
\centering
\scalebox{1}{%
\begin{tabular}{lcccccc}\toprule
& \multicolumn{2}{c}{Train} & \multicolumn{4}{c}{Test}
\\\cmidrule(lr){2-3}\cmidrule(lr){4-7}
           & Accuracy  & AUC & Accuracy    & AUC  & Mean \sev{+} & Mean \sev{-}\\\midrule
MLP & $0.87\pm0.00$ & $0.93\pm0.00$ & $0.86\pm0.00$ & $0.92\pm0.00$ & $2.97\pm0.01$ & $1.21\pm0.04$\\
\allopt{+} & $0.86\pm0.01$ & $0.92\pm0.01$ & $0.84\pm0.00$ & $0.90\pm0.00$ & \textcolor{red}{$1.01\pm0.01$}\\
\allopt{-} & $0.86\pm0.00$ & $0.92\pm0.00$ & $0.85\pm0.00$ & $0.93\pm0.00$ & & \textcolor{red}{$1.03\pm0.03$}\\\bottomrule
\end{tabular}}
\caption{SEV Optimization Performance of multi-layer perceptrons on Adult}
\label{tab:mlp_adult_perf}
\end{table*}

\begin{table*}[ht]
\centering
\scalebox{1}{%
\begin{tabular}{lcccccc}\toprule
& \multicolumn{2}{c}{Train} & \multicolumn{2}{c}{Test}
\\\cmidrule(lr){2-3}\cmidrule(lr){4-5}
           & Accuracy  & AUC & Accuracy    & AUC  & Mean \sev{+} & Mean \sev{-}\\\midrule
GBDT & $0.88\pm0.00$ & $0.93\pm0.00$ & $0.87\pm0.00$ & $0.93\pm0.00$ & $2.66\pm0.17$ & $1.21\pm0.02$\\
\allopt{+} & $0.88\pm0.00$ & $0.93\pm0.01$ & $0.87\pm0.00$ & $0.90\pm0.00$ & \textcolor{red}{$1.03\pm0.01$}\\
\allopt{-} & $0.87\pm0.01$ & $0.93\pm0.01$ & $0.85\pm0.02$ & $0.91\pm0.01$ & & \textcolor{red}{$1.00\pm0.03$}\\\bottomrule
\end{tabular}}
\caption{SEV Optimization Performance of gradient boosting decision trees on Adult}
\label{tab:gbdt_adult_perf}
\end{table*}

\begin{table*}[ht]
\centering
\scalebox{1}{%
\begin{tabular}{lcccccc}\toprule
& \multicolumn{2}{c}{Train} & \multicolumn{4}{c}{Test}
\\\cmidrule(lr){2-3}\cmidrule(lr){4-7}
           & Accuracy  & AUC & Accuracy    & AUC  & Mean \sev{+} & Mean \sev{-}\\\midrule
L2 & $0.68\pm0.00$ & $0.73\pm0.01$ & $0.68\pm0.01$ & $0.73\pm0.02$ & $1.04\pm0.01$ & $1.25\pm0.03$\\
L1 & $0.68\pm0.00$ & $0.73\pm0.00$ & $0.68\pm0.01$ & $0.73\pm0.02$ & $1.06\pm 0.01$ & $1.25\pm0.03$\\
\allopt{+} & $0.67\pm0.01$ & $0.72\pm0.01$ & $0.66\pm0.01$ & $0.71\pm0.02$ & \textcolor{red}{$1.03\pm0.03$}\\
\allopt{-} & $0.65\pm0.01$ & $0.72\pm0.01$ & $0.65\pm0.02$ & $0.72\pm0.01$ & & \textcolor{red}{$1.05\pm0.07$}\\
\volopt{} & $0.65\pm0.01$ & $0.72\pm0.02$ & $0.65\pm0.01$ & $0.71\pm0.02$ & \textcolor{red}{$1.03\pm0.05$} \\\bottomrule
\end{tabular}}
\caption{SEV Optimization Performance of linear classifiers on COMPAS}
\label{tab:lr_compas_perf}
\end{table*}

\begin{table*}[ht]
\centering
\scalebox{1}{%
\begin{tabular}{lcccccc}\toprule
& \multicolumn{2}{c}{Train} & \multicolumn{4}{c}{Test}
\\\cmidrule(lr){2-3}\cmidrule(lr){4-7}
           & Accuracy  & AUC & Accuracy    & AUC  & Mean \sev{+} & Mean \sev{-}\\\midrule
MLP & $0.68\pm0.01$ & $0.74\pm0.00$ & $0.68\pm0.01$ & $0.73\pm0.02$ & $1.01\pm0.01$ & $1.53\pm0.27$\\
\allopt{+} & $0.68\pm0.01$ & $0.73\pm0.00$ & $0.68\pm0.01$ & $0.73\pm0.01$ & \textcolor{red}{$1.00\pm0.00$}\\
\allopt{-} & $0.68\pm0.00$ & $0.73\pm0.00$ & $0.68\pm0.00$ & $0.74\pm0.01$ & & \textcolor{red}{$1.24\pm0.14$}\\\bottomrule
\end{tabular}}
\caption{SEV Optimization Performance of multi-layer perceptrons on COMPAS}
\label{tab:mlp_compas_perf}
\end{table*}

\begin{table*}[ht]
\centering
\scalebox{1}{%
\begin{tabular}{lcccccc}\toprule
& \multicolumn{2}{c}{Train} & \multicolumn{4}{c}{Test}
\\\cmidrule(lr){2-3}\cmidrule(lr){4-7}
           & Accuracy  & AUC & Accuracy    & AUC  & Mean \sev{+} & Mean \sev{-}\\\midrule
GBDT & $0.70\pm0.00$ & $0.76\pm0.00$ & $0.68\pm0.00$ & $0.73\pm0.00$ & $1.12\pm0.03$ & $1.13\pm0.03$\\
\allopt{+} & $0.70\pm0.01$ & $0.76\pm0.01$ & $0.66\pm0.01$ & $0.70\pm0.01$ & \textcolor{red}{$1.01\pm0.01$}\\
\allopt{-} & $0.70\pm0.01$ & $0.77\pm0.01$ & $0.66\pm0.01$ & $0.70\pm0.01$ & & \textcolor{red}{$1.01\pm0.01$}\\\bottomrule
\end{tabular}}
\caption{SEV Optimization Performance of gradient boosting decision trees on COMPAS}
\label{tab:gbdt_compas_perf}
\end{table*}

\begin{table*}[ht]
\centering
\scalebox{1}{%
\begin{tabular}{lcccccc}\toprule
& \multicolumn{2}{c}{Train} & \multicolumn{4}{c}{Test}
\\\cmidrule(lr){2-3}\cmidrule(lr){4-7}
           & Accuracy  & AUC & Accuracy    & AUC  & Mean \sev{+} & Mean \sev{-}\\\midrule
L2 & $0.89\pm0.00$ & $0.80\pm0.00$ & $0.89\pm0.01$ & $0.80\pm0.02$ & $4.27\pm0.04$ & $1.16\pm0.02$\\
L1 & $0.89\pm0.00$ & $0.80\pm0.00$ & $0.89\pm0.00$ & $0.80\pm0.01$ & $4.36\pm 0.01$ & $1.14\pm0.03$\\
\allopt{+} & $0.89\pm0.00$ & $0.78\pm0.01$ & $0.89\pm0.00$ & $0.78\pm0.01$ & \textcolor{red}{$3.58\pm0.33$}\\
\allopt{-} & $0.89\pm0.00$ & $0.78\pm0.01$ & $0.89\pm0.00$ & $0.78\pm0.01$ & & \textcolor{red}{$1.06\pm0.06$}\\
\volopt{} & $0.88\pm0.02$ & $0.76\pm0.01$ & $0.88\pm0.03$ & $0.76\pm0.01$ & \textcolor{red}{$3.70\pm0.71$} \\\bottomrule
\end{tabular}}
\caption{SEV Optimization Performance of linear classifiers on MIMIC III}
\label{tab:lr_mimic_perf}
\end{table*}

\begin{table*}[ht]
\centering
\scalebox{1}{%
\begin{tabular}{lcccccc}\toprule
& \multicolumn{2}{c}{Train} & \multicolumn{4}{c}{Test}
\\\cmidrule(lr){2-3}\cmidrule(lr){4-7}
           & Accuracy  & AUC & Accuracy    & AUC  & Mean \sev{+} & Mean \sev{-}\\\midrule
MLP & $0.90\pm0.00$ & $0.86\pm0.01$ & $0.90\pm0.00$ & $0.85\pm0.01$ & $4.27\pm0.10$ & $1.21\pm0.03$\\
\allopt{+} & $0.89\pm0.00$ & $0.81\pm0.01$ & $0.89\pm0.00$ & $0.81\pm0.01$ & \textcolor{red}{$2.34\pm0.77$}\\
\allopt{-} & $0.89\pm0.00$ & $0.82\pm0.01$ & $0.89\pm0.00$ & $0.82\pm0.01$ & & \textcolor{red}{$1.02\pm0.01$}\\\bottomrule
\end{tabular}}
\caption{SEV Optimization Performance of multi-layer perceptrons on the MIMIC III}
\label{tab:mlp_mimic_perf}
\end{table*}

\begin{table*}[ht]
\centering
\scalebox{1}{%
\begin{tabular}{lcccccc}\toprule
& \multicolumn{2}{c}{Train} & \multicolumn{4}{c}{Test}
\\\cmidrule(lr){2-3}\cmidrule(lr){4-7}
           & Accuracy  & AUC & Accuracy    & AUC  & Mean \sev{+} & Mean \sev{-}\\\midrule
GBDT & $0.91\pm0.00$ & $0.87\pm0.00$ & $0.90\pm0.00$ & $0.85\pm0.01$ & $5.26\pm0.08$ & $1.19\pm0.03$\\
\allopt{+} & $0.90\pm0.00$ & $0.87\pm0.00$ & $0.89\pm0.01$ & $0.83\pm0.01$ & \textcolor{red}{$3.27\pm0.02$}\\
\allopt{-} & $0.90\pm0.00$ & $0.87\pm0.00$ & $0.89\pm0.01$ & $0.83\pm0.01$ & & \textcolor{red}{$1.02\pm0.02$}\\\bottomrule
\end{tabular}}
\caption{SEV Optimization Performance of gradient boosting decision trees on MIMIC III}
\label{tab:gbdt_mimic_perf}
\end{table*}

\begin{table*}[ht]
\centering
\scalebox{1}{%
\begin{tabular}{lcccccc}\toprule
& \multicolumn{2}{c}{Train} & \multicolumn{4}{c}{Test}
\\\cmidrule(lr){2-3}\cmidrule(lr){4-7}
           & Accuracy  & AUC & Accuracy    & AUC  & Mean \sev{+} & Mean \sev{-}\\\midrule
L2 & $0.79\pm0.01$ & $0.85\pm0.00$ & $0.73\pm0.03$ & $0.76\pm0.03$ & $1.00\pm0.00$ & $1.75\pm0.16$\\
L1 & $0.76\pm0.01$ & $0.80\pm0.01$ & $0.73\pm0.02$ & $0.77\pm0.03$ & $1.00\pm 0.00$ & $1.13\pm0.01$\\
\allopt{+} & $0.79\pm0.01$ & $0.83\pm0.01$ & $0.75\pm0.05$ & $0.77\pm0.05$ & \textcolor{red}{$1.00\pm0.00$}\\
\allopt{-} & $0.78\pm0.01$ & $0.83\pm0.01$ & $0.75\pm0.05$ & $0.77\pm0.05$ & & \textcolor{red}{$1.04\pm0.05$}\\
\volopt{} & $0.77\pm0.01$ & $0.83\pm0.01$ & $0.70\pm0.05$ & $0.75\pm0.05$ & & \textcolor{red}{$1.00\pm0.00$} \\\bottomrule
\end{tabular}}
\caption{SEV Optimization Performance of linear classifiers on German Credit}
\label{tab:lr_german_perf}
\end{table*}

\begin{table*}[ht]
\centering
\scalebox{1}{%
\begin{tabular}{lcccccc}\toprule
& \multicolumn{2}{c}{Train} & \multicolumn{4}{c}{Test}
\\\cmidrule(lr){2-3}\cmidrule(lr){4-7}
           & Accuracy  & AUC & Accuracy    & AUC  & Mean \sev{+} & Mean \sev{-}\\\midrule
MLP & $0.80\pm0.00$ & $0.86\pm0.02$ & $0.76\pm0.03$ & $0.79\pm0.03$ & $1.04\pm0.08$ & $1.58\pm0.22$\\
\allopt{+} & $1.00\pm0.03$ & $1.00\pm0.03$ & $0.73\pm0.02$ & $0.80\pm0.04$ & \textcolor{red}{$1.00\pm0.00$}\\
\allopt{-} & $1.00\pm0.01$ & $1.00\pm0.02$ & $0.76\pm0.04$ & $0.81\pm0.05$ & & \textcolor{red}{$1.18\pm0.12$}\\\bottomrule
\end{tabular}}
\caption{SEV Optimization Performance of multi-layer perceptrons on German Credit}
\label{tab:mlp_german_perf}
\end{table*}

\begin{table*}[!ht]
\centering
\scalebox{1}{%
\begin{tabular}{lcccccc}\toprule
& \multicolumn{2}{c}{Train} & \multicolumn{4}{c}{Test}
\\\cmidrule(lr){2-3}\cmidrule(lr){4-7}
           & Accuracy  & AUC & Accuracy    & AUC  & Mean \sev{+} & Mean \sev{-}\\\midrule
GBDT & $0.96\pm0.00$ & $0.99\pm0.00$ & $0.74\pm0.03$ & $0.78\pm0.02$ & $1.10\pm0.24$ & $1.50\pm0.06$\\
\allopt{+} & $1.00\pm0.00$ & $1.00\pm0.00$ & $0.73\pm0.02$ & $0.76\pm0.02$ & \textcolor{red}{$1.01\pm0.01$}\\
\allopt{-} & $1.00\pm0.00$ & $1.00\pm0.00$ & $0.74\pm0.02$ & $0.76\pm0.03$ & & \textcolor{red}{$1.17\pm0.07$}\\\bottomrule
\end{tabular}}
\caption{SEV Optimization Performance of gradient boosting decision trees on German Credit}
\label{tab:gbdt_german_perf}
\end{table*}

\begin{table*}[!ht]
\centering

\scalebox{1}{%
\begin{tabular}{lcccccc}\toprule
& \multicolumn{2}{c}{Train} & \multicolumn{4}{c}{Test}
\\\cmidrule(lr){2-3}\cmidrule(lr){4-7}
           & Accuracy  & AUC & Accuracy    & AUC  & Mean \sev{+} & Mean \sev{-}\\\midrule
L2 & $0.72\pm0.00$ & $0.78\pm0.00$ & $0.71\pm0.01$ & $0.78\pm0.01$ & $1.28\pm0.02$ & $2.76\pm0.08$\\
L1 & $0.71\pm0.00$ & $0.78\pm0.00$ & $0.71\pm0.01$ & $0.78\pm0.01$ & $1.12\pm 0.01$ & $2.46\pm0.07$\\
\allopt{+} & $0.71\pm0.00$ & $0.77\pm0.00$ & $0.70\pm0.01$ & $0.77\pm0.01$ & \textcolor{red}{$1.03\pm0.02$}\\
\allopt{-} & $0.68\pm0.01$ & $0.75\pm0.01$ & $0.68\pm0.01$ & $0.75\pm0.01$ & & \textcolor{red}{$1.14\pm0.04$}\\
\volopt{} & $0.70\pm0.02$ & $0.77\pm0.01$ & $0.69\pm0.01$ & $0.76\pm0.02$ & \textcolor{red}{$1.04\pm0.04$} \\\bottomrule
\end{tabular}}
\caption{SEV Optimization Performance of linear classifiers on FICO}
\label{tab:lr_fico_perf}
\end{table*}

\begin{table*}[!ht]
\centering
\scalebox{1}{%
\begin{tabular}{lcccccc}\toprule
& \multicolumn{2}{c}{Train} & \multicolumn{4}{c}{Test}
\\\cmidrule(lr){2-3}\cmidrule(lr){4-7}
           & Accuracy  & AUC & Accuracy    & AUC  & Mean \sev{+} & Mean \sev{-}\\\midrule
MLP & $0.73\pm0.01$ & $0.81\pm0.01$ & $0.72\pm0.01$ & $0.79\pm0.01$ & $1.01\pm0.01$ & $2.83\pm0.22$\\
\allopt{+} & $0.68\pm0.01$ & $0.78\pm0.01$ & $0.67\pm0.01$ & $0.78\pm0.02$ & \textcolor{red}{$1.01\pm0.04$}\\
\allopt{-} & $0.77\pm0.00$ & $0.85\pm0.00$ & $0.67\pm0.01$ & $0.75\pm0.01$ & & \textcolor{red}{$1.52\pm0.44$}\\\bottomrule
\end{tabular}}
\caption{SEV Optimization Performance of multi-layer perceptrons on FICO}
\label{tab:mlp_fico_perf}
\end{table*}

\begin{table*}[!ht]
\centering

\scalebox{0.95}{%
\begin{tabular}{lcccccc}\toprule
& \multicolumn{2}{c}{Train} & \multicolumn{4}{c}{Test}
\\\cmidrule(lr){2-3}\cmidrule(lr){4-7}
           & Accuracy  & AUC & Accuracy    & AUC  & Mean \sev{+} & Mean \sev{-}\\\midrule
GBDT & $0.77\pm0.00$ & $0.85\pm0.00$ & $0.73\pm0.03$ & $0.80\pm0.01$ & $1.00\pm0.00$ & $3.58\pm0.13$\\
\allopt{+} & $0.80\pm0.00$ & $0.98\pm0.01$ & $0.70\pm0.01$ & $0.75\pm0.01$ & \textcolor{red}{$1.00\pm0.00$}\\
\allopt{-} & $0.80\pm0.01$ & $0.89\pm0.01$ & $0.69\pm0.01$ & $0.75\pm0.01$ & & \textcolor{red}{$1.84\pm0.10$}\\\bottomrule
\end{tabular}}
\caption{SEV Optimization Performance of gradient boosting decision trees on FICO}
\label{tab:gbdt_fico_perf}
\end{table*}

\begin{table*}[ht]
\centering
\scalebox{1}{%
\begin{tabular}{lcccccc}\toprule
& \multicolumn{2}{c}{Train} & \multicolumn{4}{c}{Test}
\\\cmidrule(lr){2-3}\cmidrule(lr){4-7}
           & Accuracy  & AUC & Accuracy    & AUC  & Mean \sev{+} & Mean \sev{-}\\\midrule
L2 & $0.62\pm0.00$ & $0.67\pm0.00$ & $0.62\pm0.00$ & $0.67\pm0.00$ & $1.00\pm0.00$ & $1.53\pm0.01$\\
L1 & $0.62\pm0.00$ & $0.67\pm0.00$ & $0.62\pm0.00$ & $0.67\pm0.00$ & $1.00\pm 0.00$ & $1.49\pm0.01$\\
\allopt{+} & $0.56\pm0.00$ & $0.63\pm0.00$ & $0.56\pm0.00$ & $0.63\pm0.00$ & \textcolor{red}{$1.00\pm0.00$}\\
\allopt{-} & $0.56\pm0.00$ & $0.63\pm0.00$ & $0.56\pm0.00$ & $0.63\pm0.00$ & & \textcolor{red}{$1.00\pm0.00$}\\
\volopt{} & $0.57\pm0.04$ & $0.62\pm0.03$ & $0.57\pm0.03$ & $0.61\pm0.03$ & \textcolor{red}{$1.00\pm0.00$} \\\bottomrule
\end{tabular}}
\caption{SEV Optimization Performance of linear classifiers on Diabetes}
\label{tab:lr_diabetes_perf}
\end{table*}

\begin{table*}[ht]
\centering

\scalebox{1}{%
\begin{tabular}{lcccccc}\toprule
& \multicolumn{2}{c}{Train} & \multicolumn{4}{c}{Test}
\\\cmidrule(lr){2-3}\cmidrule(lr){4-7}
           & Accuracy  & AUC & Accuracy    & AUC  & Mean \sev{+} & Mean \sev{-}\\\midrule
MLP & $0.73\pm0.01$ & $0.81\pm0.01$ & $0.72\pm0.01$ & $0.79\pm0.01$ & $1.01\pm0.01$ & $2.83\pm0.22$\\
\allopt{+} & $0.68\pm0.01$ & $0.78\pm0.01$ & $0.67\pm0.01$ & $0.78\pm0.02$ & \textcolor{red}{$1.01\pm0.04$}\\
\allopt{-} & $0.77\pm0.00$ & $0.85\pm0.00$ & $0.67\pm0.01$ & $0.75\pm0.01$ & & \textcolor{red}{$1.52\pm0.44$}\\\bottomrule
\end{tabular}}
\caption{SEV Optimization Performance of multi-layer perceptrons on Diabetes}
\label{tab:mlp_diabetes_perf}
\end{table*}

\begin{table*}[!ht]
\centering
\scalebox{1}{%
\begin{tabular}{lcccccc}\toprule
& \multicolumn{2}{c}{Train} & \multicolumn{4}{c}{Test}
\\\cmidrule(lr){2-3}\cmidrule(lr){4-7}
           & Accuracy  & AUC & Accuracy    & AUC  & Mean \sev{+} & Mean \sev{-}\\\midrule
GBDT & $0.77\pm0.00$ & $0.85\pm0.00$ & $0.73\pm0.03$ & $0.80\pm0.01$ & $1.00\pm0.00$ & $3.58\pm0.13$\\
\allopt{+} & $0.80\pm0.00$ & $0.98\pm0.01$ & $0.70\pm0.01$ & $0.75\pm0.01$ & \textcolor{red}{$1.00\pm0.00$}\\
\allopt{-} & $0.80\pm0.01$ & $0.89\pm0.01$ & $0.69\pm0.01$ & $0.75\pm0.01$ & & \textcolor{red}{$1.84\pm0.10$}\\\bottomrule
\end{tabular}}
\caption{SEV Optimization Performance of gradient boosting decision trees on Diabetes}
\label{tab:gbdt_diabetes_perf}
\end{table*}
\fi

\clearpage
\subsection{Detailed SEV Distributions Plots}
\label{subsec:sev_distributions}

In this section, we will display the \sev{+} and \sev{-} counts distributions for optimized and unoptimized models for a single train-test split of each dataset. The distribution of the \sev{+} and \sev{-} counts are shown in subfigures \textit{(a)}, while the accompanying Sankey plot shows how counts change due to optimization. The red on the left side of the Sankey Plots shows the distribution of the SEV before optimization, while the red on the right shows the SEV distribution after optimization.v
\ifshow
\begin{figure}[ht]
     \centering
     \begin{subfigure}[b]{0.4\textwidth}
         \centering
         \includegraphics[width=\textwidth]{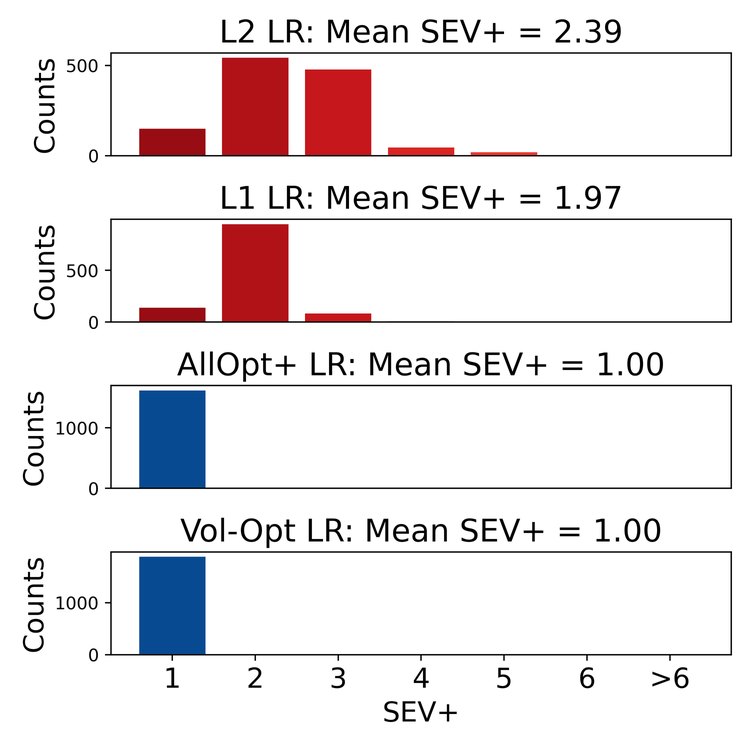}
         \caption{\sev{+} counts across methods}
     \end{subfigure}
     \hfill
     \begin{subfigure}[b]{0.4\textwidth}
         \centering
         \includegraphics[width=\textwidth]{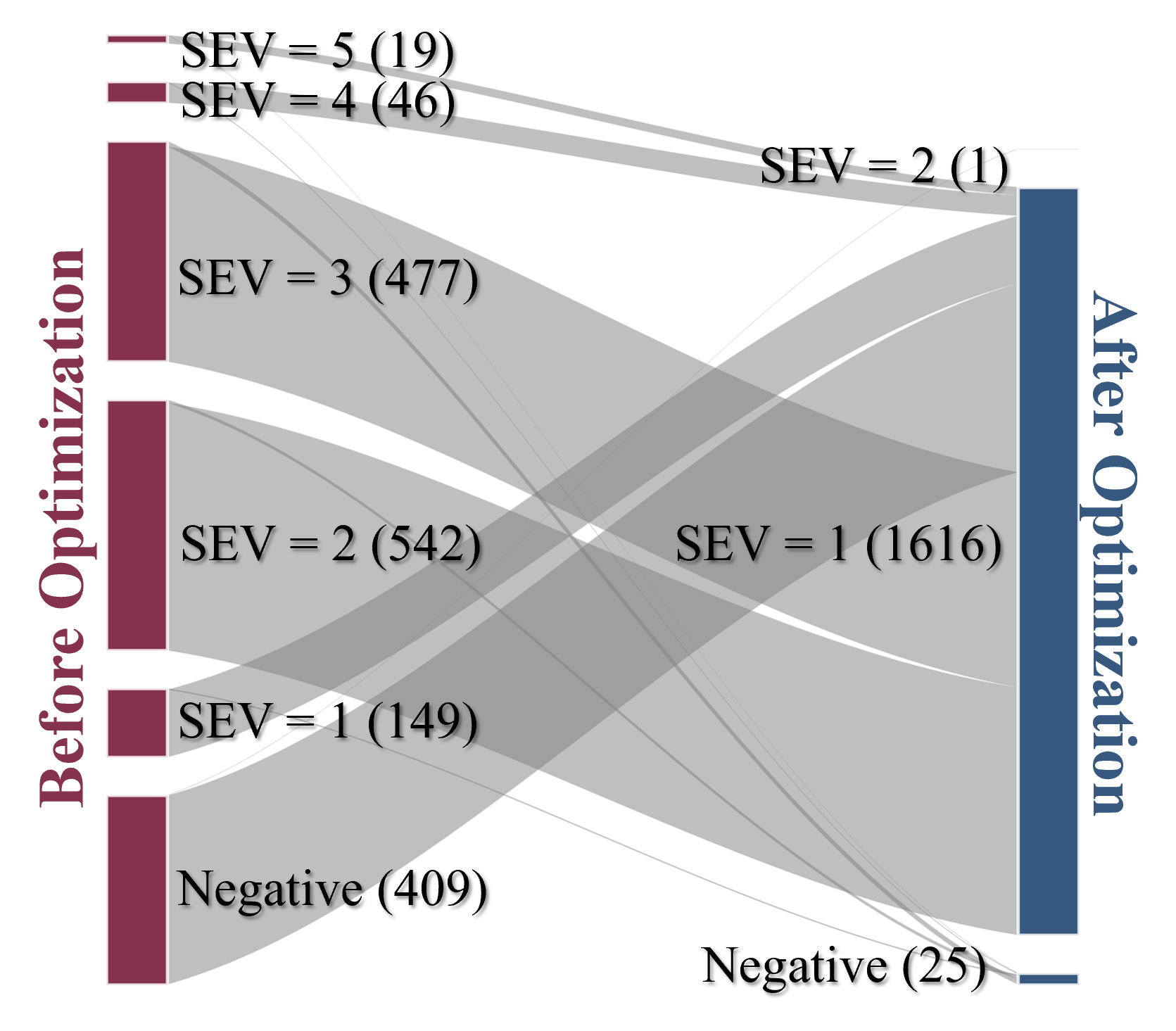}
         \caption{L2 LR \sev{+} to \allopt{+} LR}
     \end{subfigure}
     \begin{subfigure}[b]{0.4\textwidth}
         \centering
         \includegraphics[width=\textwidth]{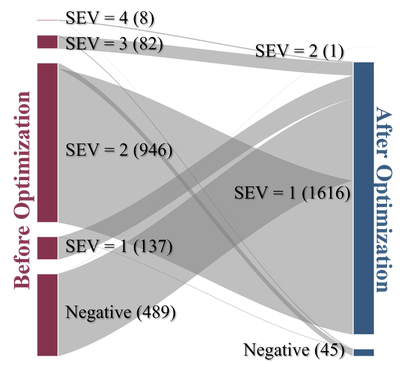}
         \caption{L1 LR \sev{+} to \allopt{+} LR}
     \end{subfigure}
     \hfill
     \begin{subfigure}[b]{0.4\textwidth}
         \centering
         \includegraphics[width=\textwidth]{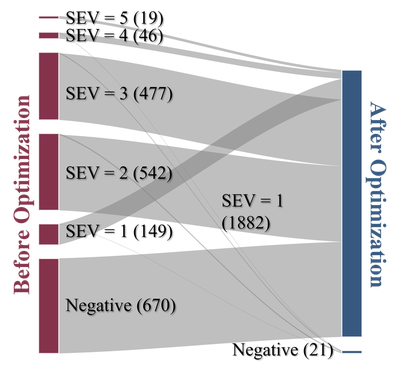}
         \caption{L2 LR \sev{+} to \volopt{} LR}
     \end{subfigure}
    \caption{\sev{+} Optimization performance for linear classifiers on Adult}
    \label{fig:adult_lr_sev+}
\end{figure}

\begin{figure}[!ht]
    \centering
    \begin{subfigure}[b]{0.3\textwidth}
         \centering
         \includegraphics[width=\textwidth]{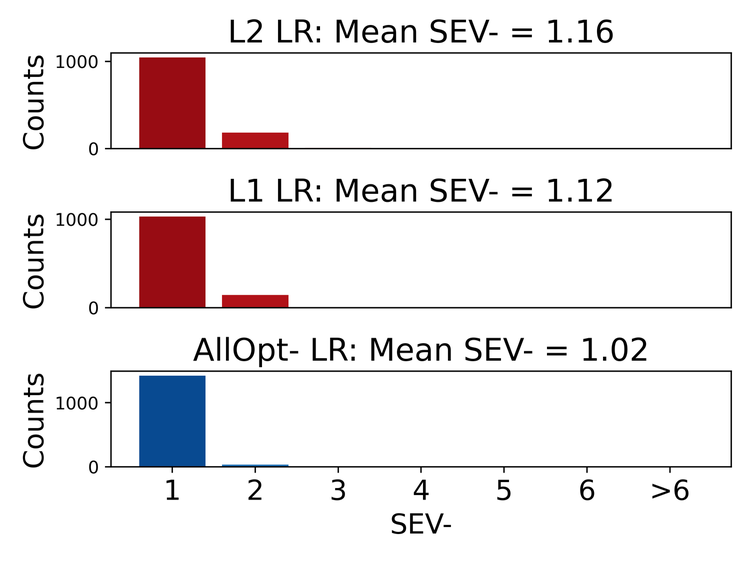}
         \caption{\sev{-} counts across methods}
     \end{subfigure}
     \hfill
    \begin{subfigure}[b]{0.3\textwidth}
         \centering
         \includegraphics[width=\textwidth]{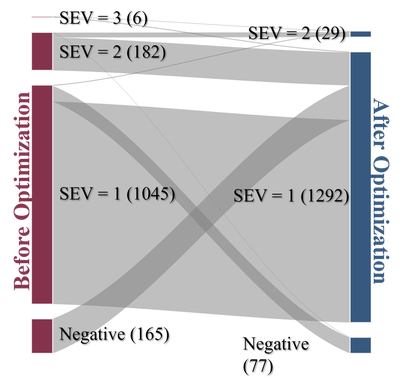}
         \caption{L2 LR \sev{-} to \allopt{-} LR}
     \end{subfigure}
     \hfill
     \begin{subfigure}[b]{0.3\textwidth}
         \centering
         \includegraphics[width=\textwidth]{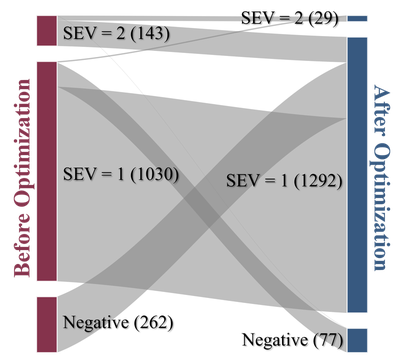}
         \caption{L1 LR \sev{-} to \allopt{-} LR}
     \end{subfigure}
     \caption{\sev{-} optimization performance for linear classifiers on Adult}
    \label{fig:adult_lr_sev-}
\end{figure}

\begin{figure}[!ht]
    \centering
    \begin{subfigure}[b]{0.45\textwidth}
         \centering
         \includegraphics[width=\textwidth]{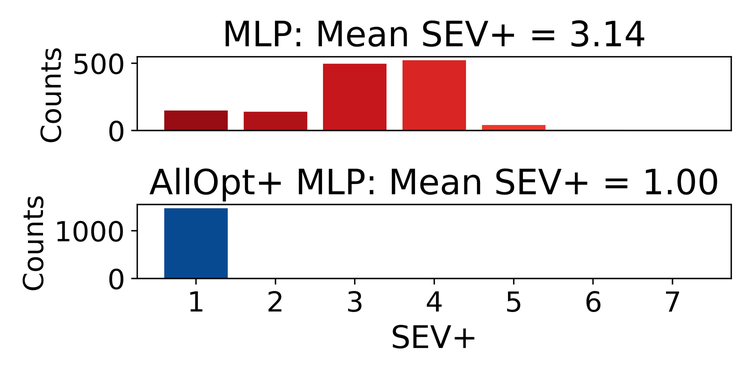}
         \caption{\sev{+} counts across methods}
     \end{subfigure}
     \hfill
     \begin{subfigure}[b]{0.35\textwidth}
         \centering
         \includegraphics[width=\textwidth]{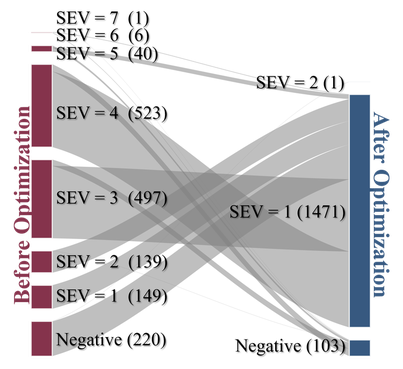}
         \caption{MLP to \allopt{+} MLP}
     \end{subfigure}
     \caption{\sev{+} optimization performance for multi-layer perceptions on Adult}
\end{figure}

\begin{figure}[ht]
    \centering
    \begin{subfigure}[b]{0.45\textwidth}
         \centering
         \includegraphics[width=\textwidth]{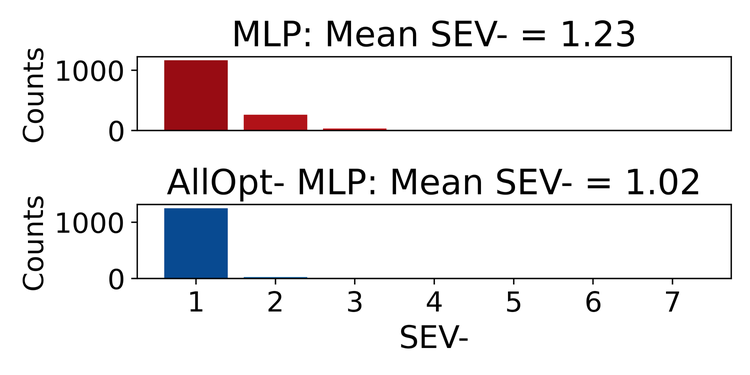}
         \caption{\sev{-} counts across methods}
     \end{subfigure}
     \hfill
     \begin{subfigure}[b]{0.35\textwidth}
         \centering
         \includegraphics[width=\textwidth]{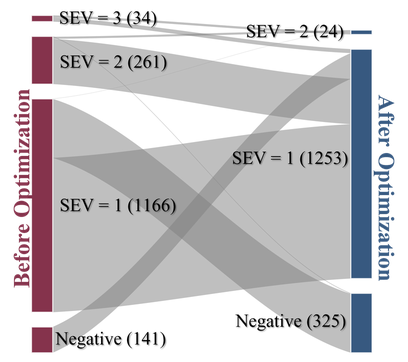}
         \caption{MLP to \allopt{-} MLP}
     \end{subfigure}
     \caption{\sev{-} optimization performance for multi-layer perceptions on Adult}
\end{figure}

\begin{figure}[!ht]
    \centering
    \begin{subfigure}[b]{0.45\textwidth}
         \centering
         \includegraphics[width=\textwidth]{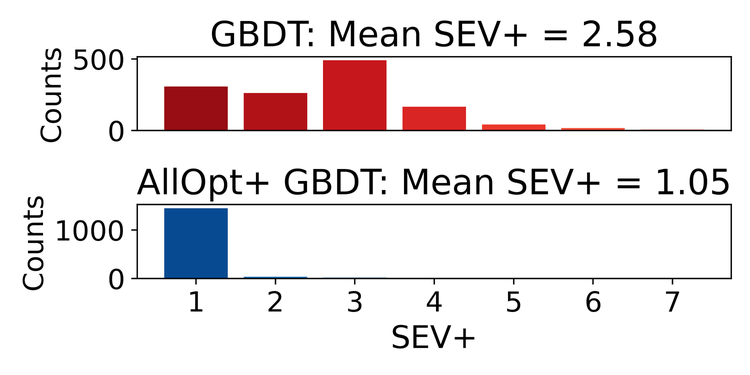}
         \caption{\sev{+} counts across methods}
     \end{subfigure}
     \hfill
     \begin{subfigure}[b]{0.35\textwidth}
         \centering
         \includegraphics[width=\textwidth]{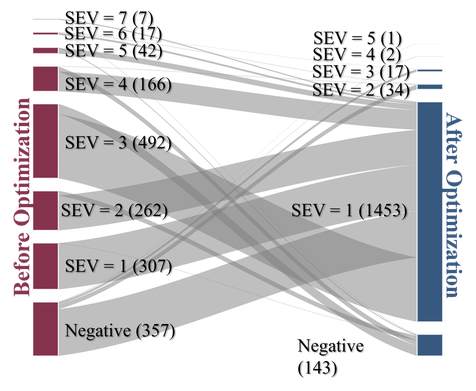}
         \caption{GBDT to \allopt{+} GBDT}
     \end{subfigure}
     \caption{\sev{+} optimization performance for gradient boosting trees on Adult}
\end{figure}

\begin{figure}[ht]
    \centering
    \begin{subfigure}[b]{0.45\textwidth}
         \centering
         \includegraphics[width=\textwidth]{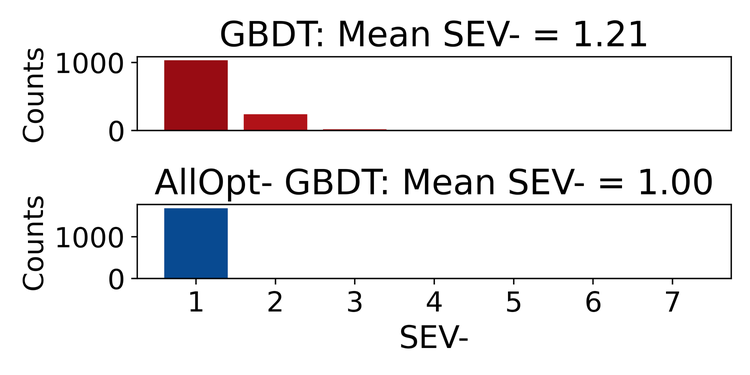}
         \caption{\sev{-} counts across methods}
     \end{subfigure}
     \hfill
     \begin{subfigure}[b]{0.35\textwidth}
         \centering
         \includegraphics[width=\textwidth]{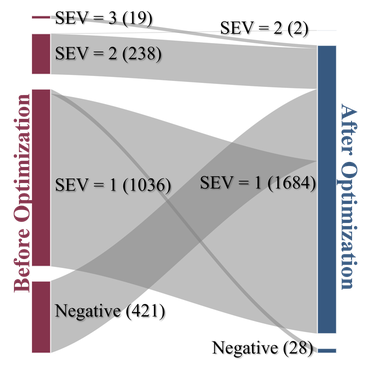}
         \caption{GBDT to \allopt{-} GBDT}
     \end{subfigure}
     \caption{\sev{-} optimization performance for gradient boosting trees on Adult}
\end{figure}

\begin{figure}[!ht]
     \centering
     \begin{subfigure}[b]{0.4\textwidth}
         \centering
         \includegraphics[width=\textwidth]{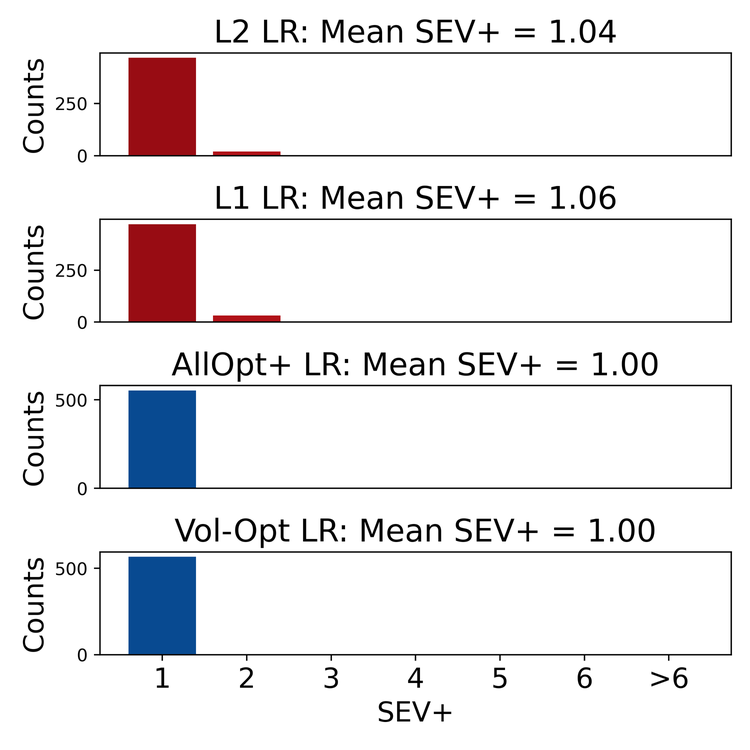}
         \caption{\sev{+} counts across methods}
     \end{subfigure}
     \hfill
     \begin{subfigure}[b]{0.4\textwidth}
         \centering
         \includegraphics[width=\textwidth]{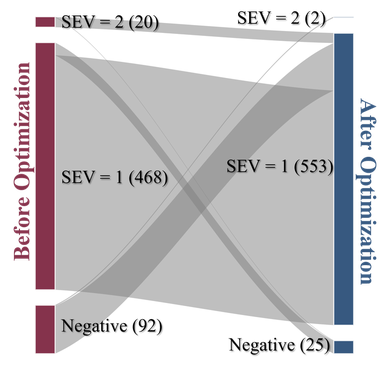}
         \caption{L2 LR \sev{+} to \allopt{+} LR}
     \end{subfigure}
     \caption{\sev{+} Optimization performance for linear classifiers on COMPAS}
    \label{fig:compas_lr_sev+}
\end{figure}
\begin{figure}[ht]
\centering
     \begin{subfigure}[b]{0.4\textwidth}
         \centering
         \includegraphics[width=\textwidth]{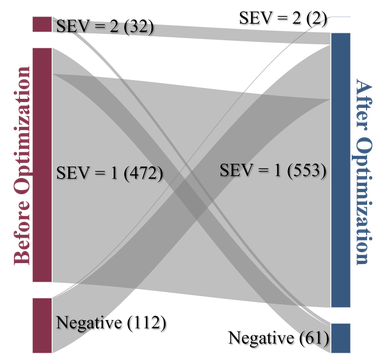}
         \caption{L1 LR \sev{+} to \allopt{+} LR}
     \end{subfigure}
     \hfill
     \begin{subfigure}[b]{0.4\textwidth}
         \centering
         \includegraphics[width=\textwidth]{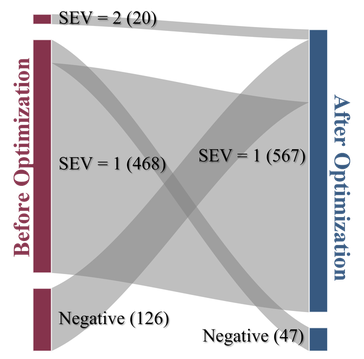}
         \caption{L2 LR \sev{+} to \volopt{} LR}
     \end{subfigure}
        \caption{\sev{+} Optimization performance for linear classifiers on COMPAS(cont.)}
        \label{fig:compas_lr_sev+_2}
\end{figure}

\begin{figure}[ht]
    \centering
    \begin{subfigure}[b]{0.3\textwidth}
         \centering
         \includegraphics[width=\textwidth]{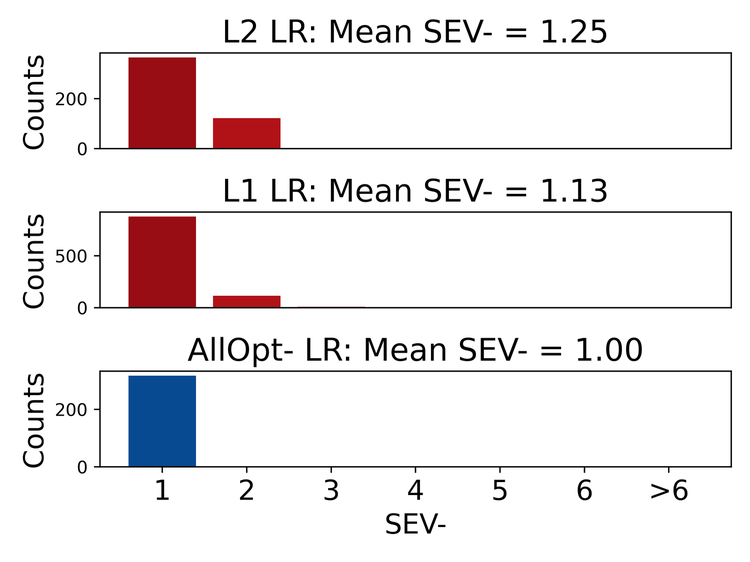}
         \caption{\sev{-} counts across methods}
     \end{subfigure}
     \hfill
    \begin{subfigure}[b]{0.3\textwidth}
         \centering
         \includegraphics[width=\textwidth]{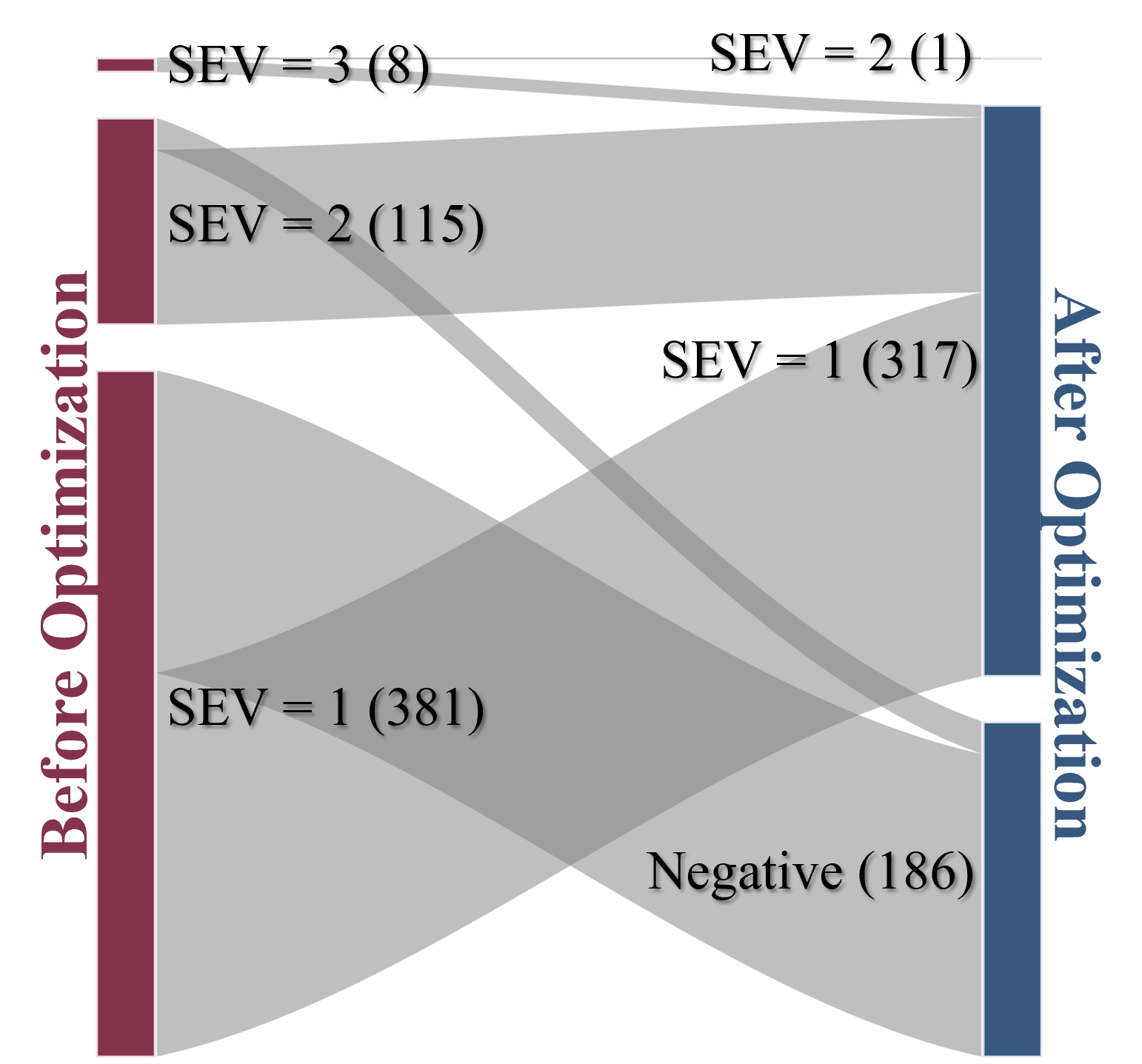}
         \caption{L2 LR \sev{-} to \allopt{-} LR}
     \end{subfigure}
     \hfill
     \begin{subfigure}[b]{0.3\textwidth}
         \centering
         \includegraphics[width=\textwidth]{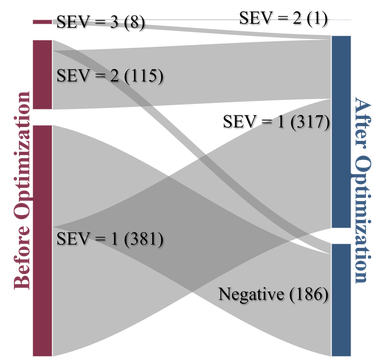}
         \caption{L1 LR \sev{-} to \allopt{-} LR}
     \end{subfigure}
     \caption{\sev{-} optimization performance for linear classifiers on COMPAS}
    \label{fig:compas_lr_sev-}
\end{figure}

\begin{figure}[!ht]
    \centering
    \begin{subfigure}[b]{0.45\textwidth}
         \centering
         \includegraphics[width=\textwidth]{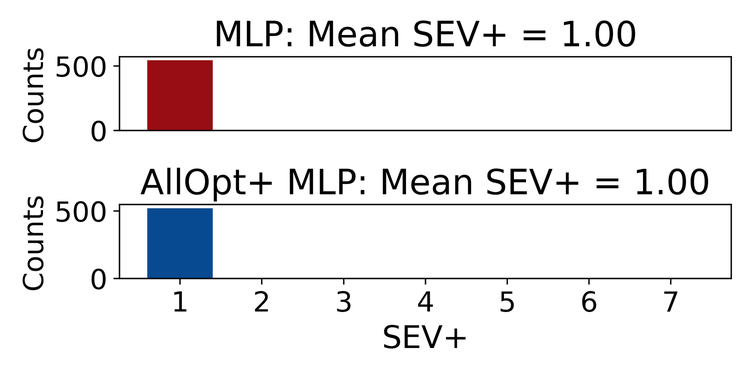}
         \caption{\sev{+} counts across methods}
     \end{subfigure}
     \hfill
     \begin{subfigure}[b]{0.35\textwidth}
         \centering
         \includegraphics[width=\textwidth]{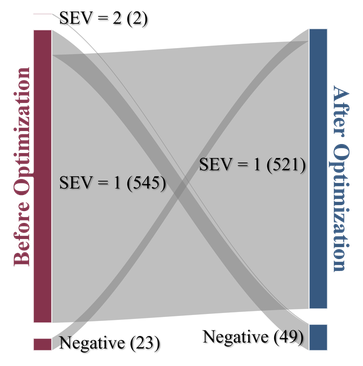}
         \caption{MLP to \allopt{+} MLP}
     \end{subfigure}
     \caption{\sev{+} optimization performance for multi-layer perceptions on COMPAS}
\end{figure}

\begin{figure}[!ht]
    \centering
    \begin{subfigure}[b]{0.45\textwidth}
         \centering
         \includegraphics[width=\textwidth]{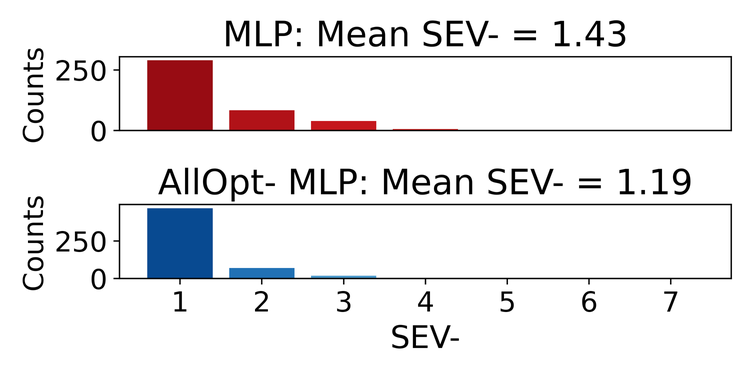}
         \caption{\sev{-} counts across methods}
     \end{subfigure}
     \hfill
     \begin{subfigure}[b]{0.35\textwidth}
         \centering
         \includegraphics[width=\textwidth]{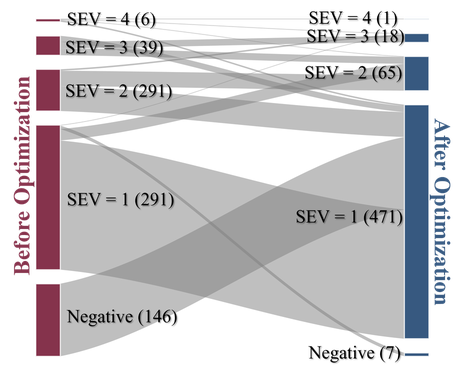}
         \caption{MLP to \allopt{-} MLP}
     \end{subfigure}
     \caption{\sev{-} optimization performance for multi-layer perceptions on COMPAS}
\end{figure}

\begin{figure}[!ht]
    \centering
    \begin{subfigure}[b]{0.45\textwidth}
         \centering
         \includegraphics[width=\textwidth]{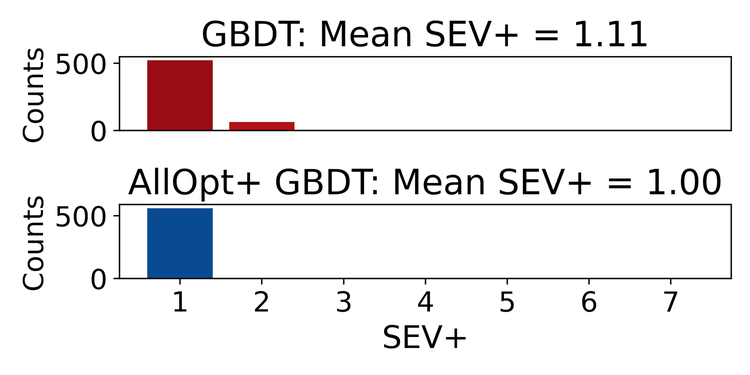}
         \caption{\sev{+} counts across methods}
     \end{subfigure}
     \hfill
     \begin{subfigure}[b]{0.35\textwidth}
         \centering
         \includegraphics[width=\textwidth]{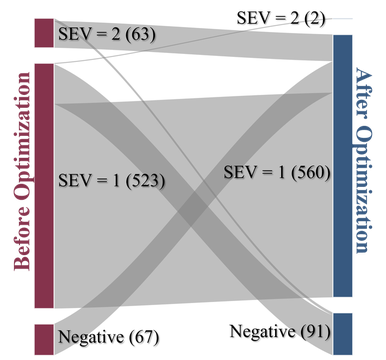}
         \caption{GBDT to \allopt{+} GBDT}
     \end{subfigure}
     \caption{\sev{+} optimization performance for gradient boosting trees on COMPAS}
\end{figure}

\begin{figure}[ht]
    \centering
    \begin{subfigure}[b]{0.45\textwidth}
         \centering
         \includegraphics[width=\textwidth]{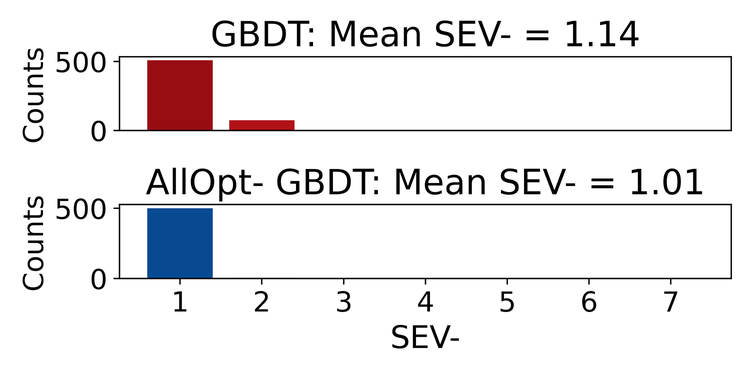}
         \caption{\sev{-} counts across methods}
     \end{subfigure}
     \hfill
     \begin{subfigure}[b]{0.35\textwidth}
         \centering
         \includegraphics[width=\textwidth]{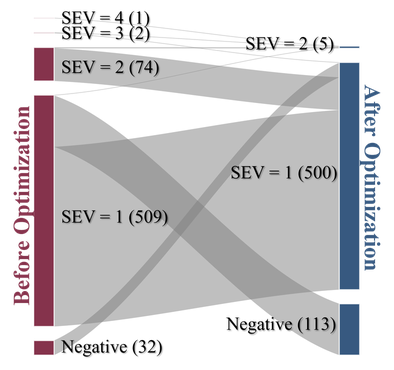}
         \caption{GBDT to \allopt{-} GBDT}
     \end{subfigure}
     \caption{\sev{-} optimization performance for gradient boosting trees on COMPAS}
\end{figure}

\begin{figure}[!ht]
     \centering
     \begin{subfigure}[b]{0.4\textwidth}
         \centering
         \includegraphics[width=\textwidth]{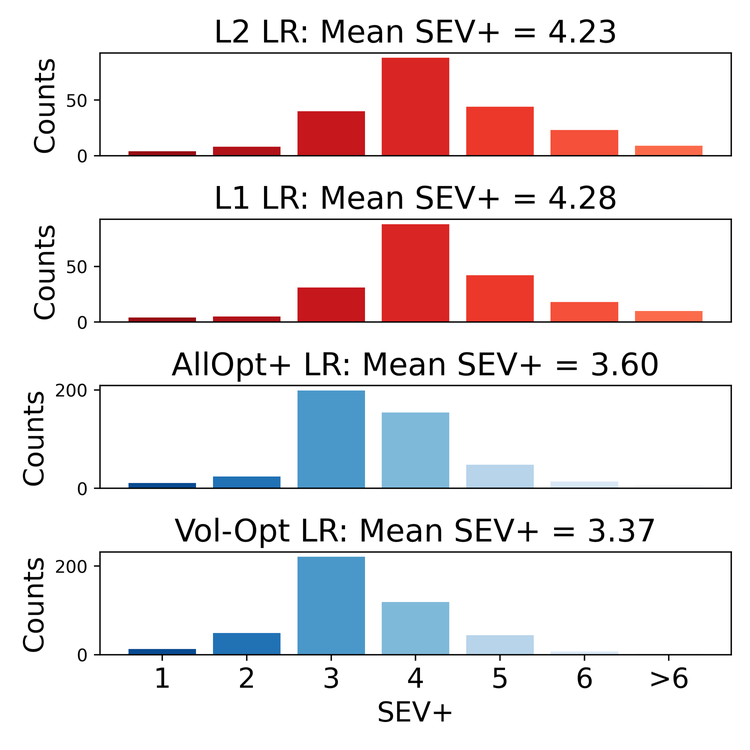}
         \caption{\sev{+} counts across methods}
     \end{subfigure}
     \hfill
     \begin{subfigure}[b]{0.4\textwidth}
         \centering
         \includegraphics[width=\textwidth]{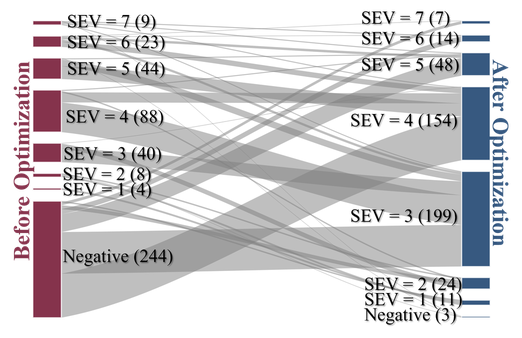}
         \caption{L2 LR \sev{+} to \allopt{+} LR}
     \end{subfigure}
     \caption{\sev{+} Optimization performance for linear classifiers on MIMIC}
        \label{fig:mimic_lr_sev+}
\end{figure}
\begin{figure}[ht]
    \centering
     \begin{subfigure}[b]{0.4\textwidth}
         \centering
         \includegraphics[width=\textwidth]{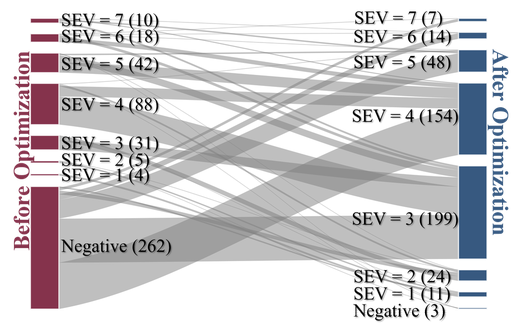}
         \caption{L1 LR \sev{+} to \allopt{+} LR}
     \end{subfigure}
     \hfill
     \begin{subfigure}[b]{0.4\textwidth}
         \centering
         \includegraphics[width=\textwidth]{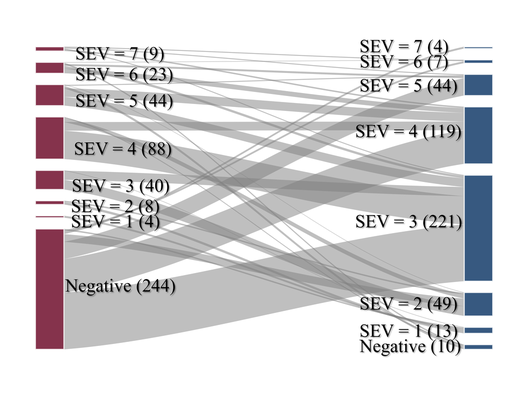}
         \caption{L2 LR \sev{+} to \volopt{} LR}
     \end{subfigure}
        \caption{\sev{+} Optimization performance for linear classifiers on MIMIC(cont.)}
        \label{fig:mimic_lr_sev+_2}
\end{figure}

\begin{figure}[ht]
    \centering
    \begin{subfigure}[b]{0.3\textwidth}
         \centering
         \includegraphics[width=\textwidth]{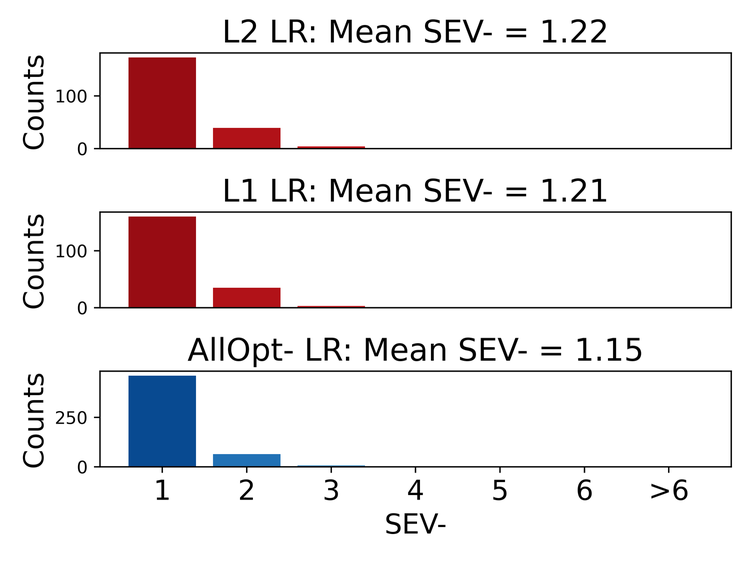}
         \caption{\sev{-} counts across methods}
     \end{subfigure}
     \hfill
    \begin{subfigure}[b]{0.3\textwidth}
         \centering
         \includegraphics[width=\textwidth]{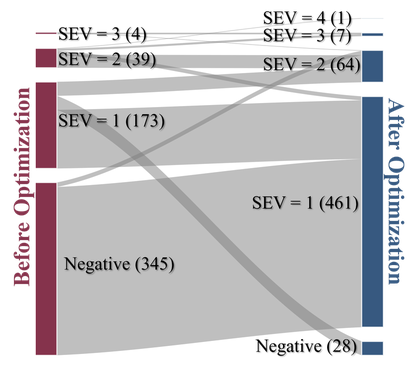}
         \caption{L2 LR \sev{-} to \allopt{-} LR}
     \end{subfigure}
     \hfill
     \begin{subfigure}[b]{0.3\textwidth}
         \centering
         \includegraphics[width=\textwidth]{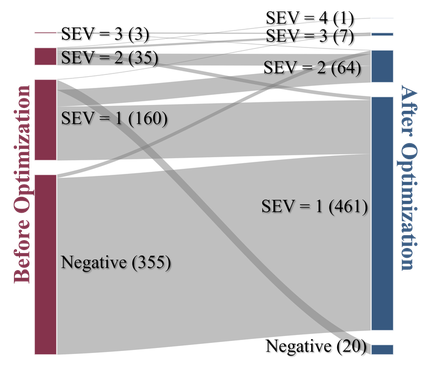}
         \caption{L1 LR \sev{-} to \allopt{-} LR}
     \end{subfigure}
     \caption{\sev{-} optimization performance for linear classifiers on MIMIC}
    \label{fig:mimic_lr_sev-}
\end{figure}

\begin{figure}[!ht]
    \centering
    \begin{subfigure}[b]{0.45\textwidth}
         \centering
         \includegraphics[width=\textwidth]{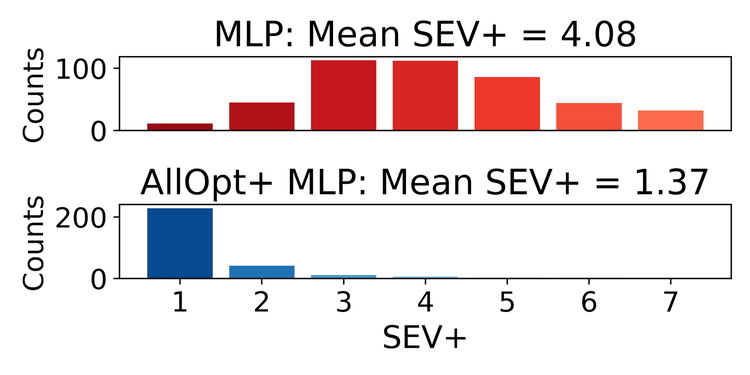}
         \caption{\sev{+} counts across methods}
     \end{subfigure}
     \hfill
     \begin{subfigure}[b]{0.35\textwidth}
         \centering
         \includegraphics[width=\textwidth]{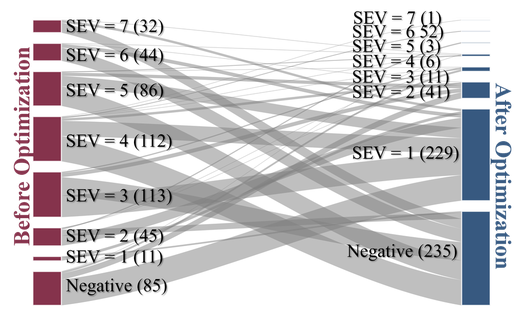}
         \caption{MLP to \allopt{+} MLP}
     \end{subfigure}
     \caption{\sev{+} optimization performance for multi-layer perceptions on MIMIC}
\end{figure}

\begin{figure}[ht]
    \centering
    \begin{subfigure}[b]{0.45\textwidth}
         \centering
         \includegraphics[width=\textwidth]{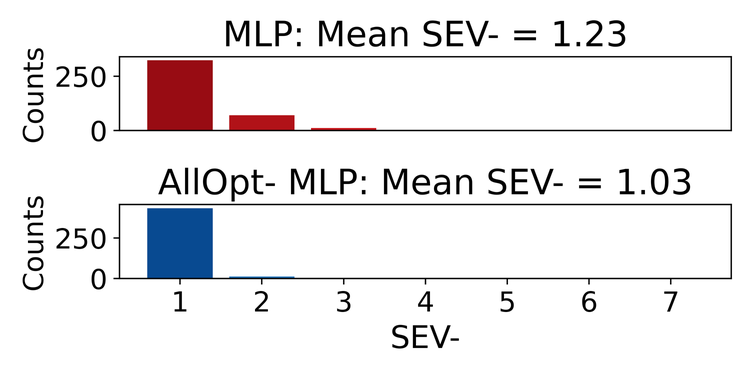}
         \caption{\sev{-} counts across methods}
     \end{subfigure}
     \hfill
     \begin{subfigure}[b]{0.35\textwidth}
         \centering
         \includegraphics[width=\textwidth]{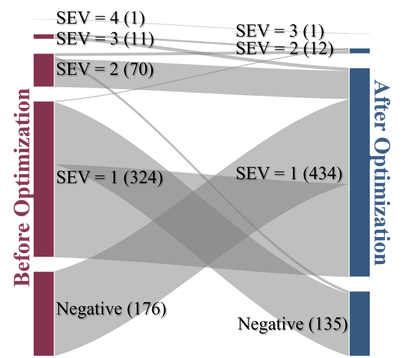}
         \caption{MLP to \allopt{-} MLP}
     \end{subfigure}
     \caption{\sev{-} optimization performance for multi-layer perceptions on MIMIC}
\end{figure}

\begin{figure}[!ht]
    \centering
    \begin{subfigure}[b]{0.45\textwidth}
         \centering
         \includegraphics[width=\textwidth]{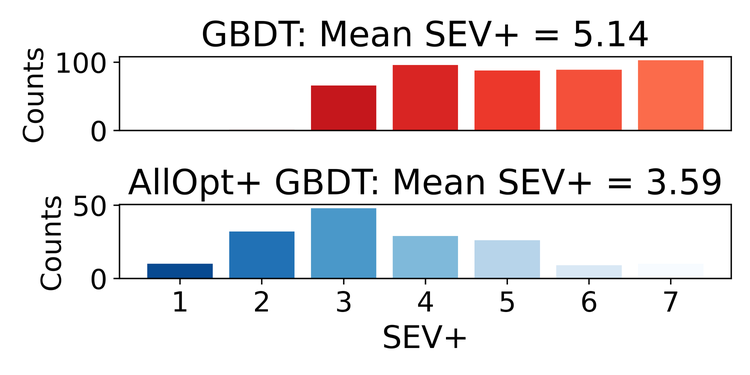}
         \caption{\sev{+} counts across methods}
     \end{subfigure}
     \hfill
     \begin{subfigure}[b]{0.35\textwidth}
         \centering
         \includegraphics[width=\textwidth]{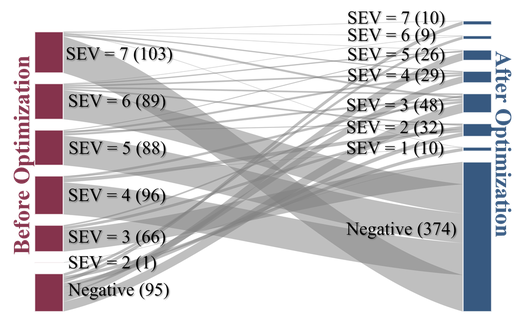}
         \caption{GBDT to \allopt{+} GBDT}
     \end{subfigure}
     \caption{\sev{+} optimization performance for gradient boosting trees on MIMIC}
\end{figure}

\begin{figure}[!ht]
    \centering
    \begin{subfigure}[b]{0.45\textwidth}
         \centering
         \includegraphics[width=\textwidth]{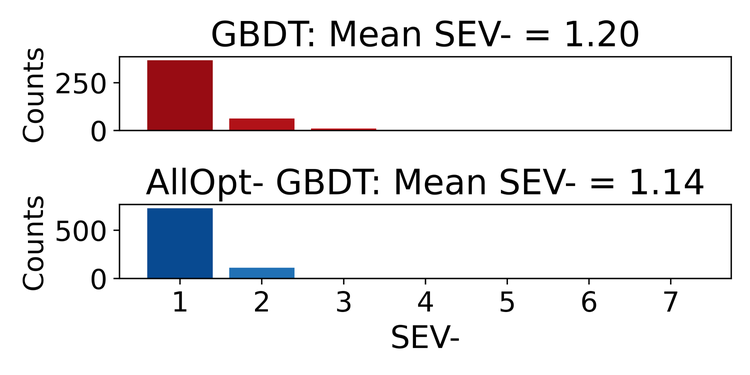}
         \caption{\sev{-} counts across methods}
     \end{subfigure}
     \hfill
     \begin{subfigure}[b]{0.35\textwidth}
         \centering
         \includegraphics[width=\textwidth]{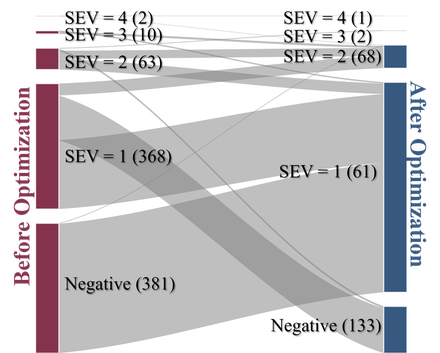}
         \caption{GBDT to \allopt{-} GBDT}
     \end{subfigure}
     \caption{\sev{-} optimization performance for gradient boosting trees on MIMIC}
\end{figure}

\begin{figure}[!ht]
     \centering
     \begin{subfigure}[b]{0.4\textwidth}
         \centering
         \includegraphics[width=\textwidth]{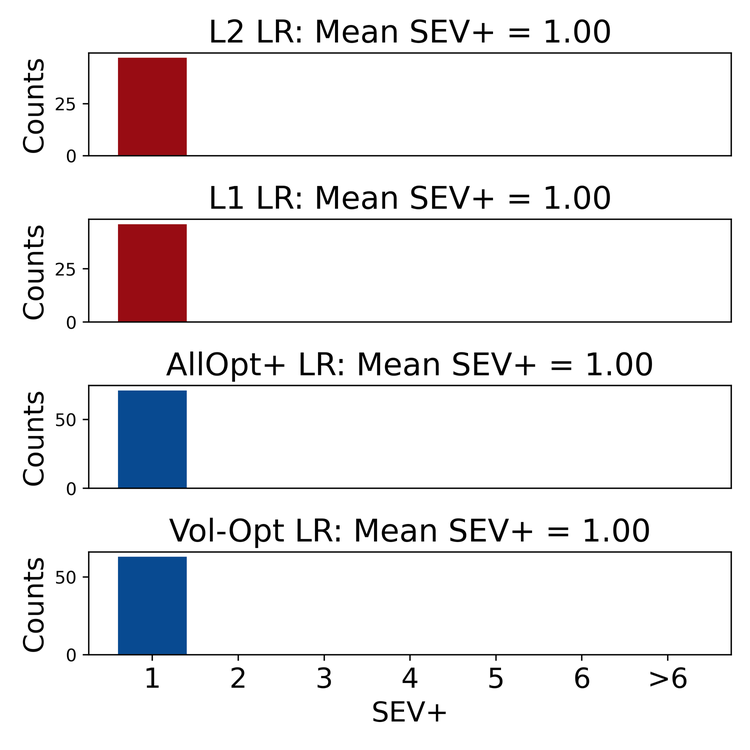}
         \caption{\sev{+} counts across methods}
     \end{subfigure}
     \hfill
     \begin{subfigure}[b]{0.4\textwidth}
         \centering
         \includegraphics[width=\textwidth]{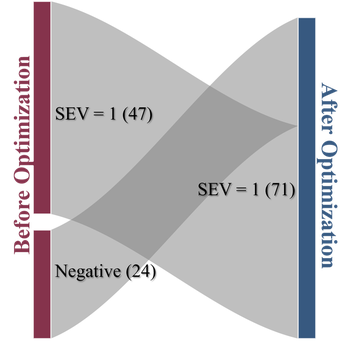}
         \caption{L2 LR \sev{+} to \allopt{+} LR}
     \end{subfigure}
     \caption{\sev{+} Optimization performance for linear classifiers on German Credit}
        \label{fig:german_lr_sev+}
\end{figure}
\begin{figure}[ht]
\centering
     \begin{subfigure}[b]{0.4\textwidth}
         \centering
         \includegraphics[width=\textwidth]{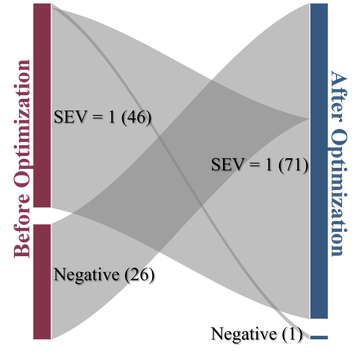}
         \caption{L1 LR \sev{+} to \allopt{+} LR}
     \end{subfigure}
     \hfill
     \begin{subfigure}[b]{0.4\textwidth}
         \centering
         \includegraphics[width=\textwidth]{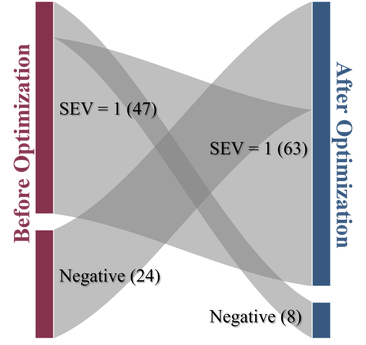}
         \caption{L2 LR \sev{+} to \volopt{} LR}
     \end{subfigure}
        \caption{\sev{+} Optimization performance for linear classifiers on German Credit (cont.)}
        \label{fig:german_lr_sev+_2}
\end{figure}

\begin{figure}[ht]
    \centering
    \begin{subfigure}[b]{0.3\textwidth}
         \centering
         \includegraphics[width=\textwidth]{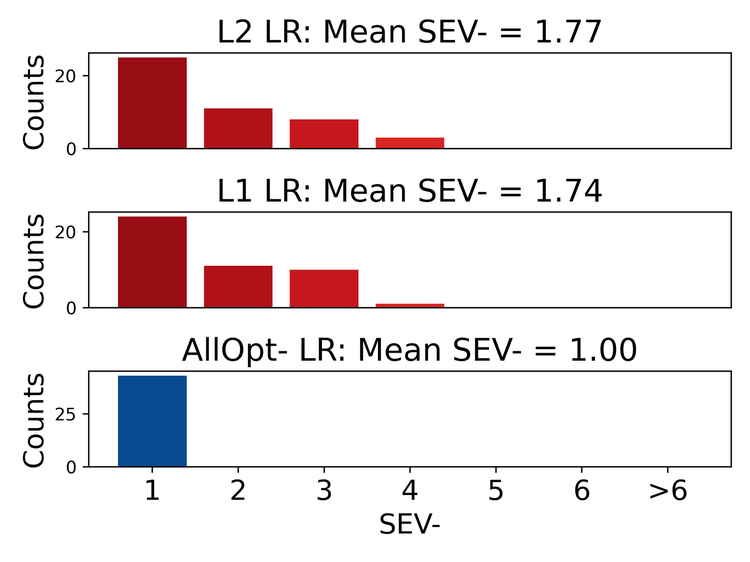}
         \caption{\sev{-} counts across methods}
     \end{subfigure}
     \hfill
    \begin{subfigure}[b]{0.3\textwidth}
         \centering
         \includegraphics[width=\textwidth]{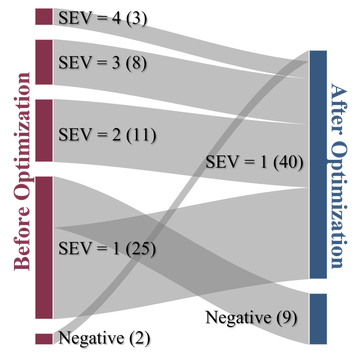}
         \caption{L2 LR \sev{-} to \allopt{-} LR}
     \end{subfigure}
     \hfill
     \begin{subfigure}[b]{0.3\textwidth}
         \centering
         \includegraphics[width=\textwidth]{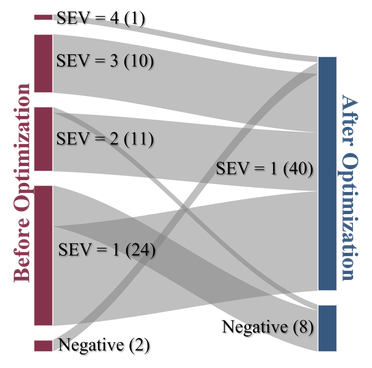}
         \caption{L1 LR \sev{-} to \allopt{-} LR}
     \end{subfigure}
     \caption{\sev{-} optimization performance for linear classifiers on German Credit}
    \label{fig:german_lr_sev-}
\end{figure}

\begin{figure}[!ht]
    \centering
    \begin{subfigure}[b]{0.45\textwidth}
         \centering
         \includegraphics[width=\textwidth]{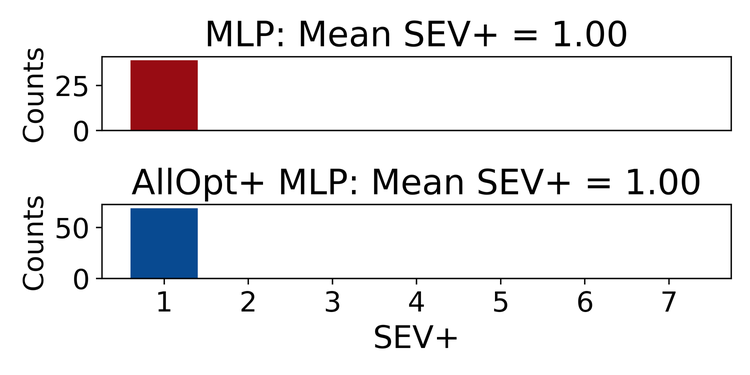}
         \caption{\sev{+} counts across methods}
     \end{subfigure}
     \hfill
     \begin{subfigure}[b]{0.35\textwidth}
         \centering
         \includegraphics[width=\textwidth]{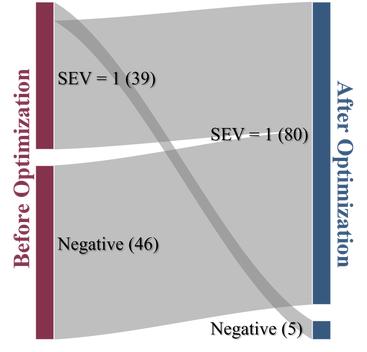}
         \caption{MLP to \allopt{+} MLP}
     \end{subfigure}
     \caption{\sev{+} optimization performance for multi-layer perceptions on German Credit}
\end{figure}

\begin{figure}[!ht]
    \centering
    \begin{subfigure}[b]{0.45\textwidth}
         \centering
         \includegraphics[width=\textwidth]{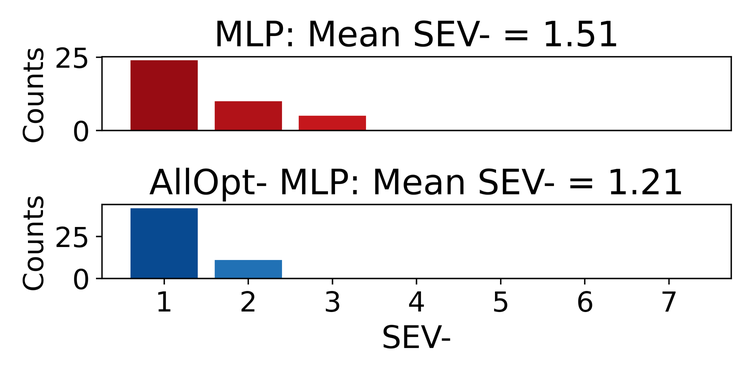}
         \caption{\sev{-} counts across methods}
     \end{subfigure}
     \hfill
     \begin{subfigure}[b]{0.35\textwidth}
         \centering
         \includegraphics[width=\textwidth]{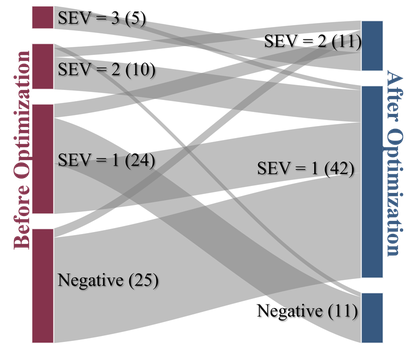}
         \caption{MLP to \allopt{-} MLP}
     \end{subfigure}
     \caption{\sev{-} optimization performance for multi-layer perceptions on German Credit}
\end{figure}

\begin{figure}[!ht]
    \centering
    \begin{subfigure}[b]{0.45\textwidth}
         \centering
         \includegraphics[width=\textwidth]{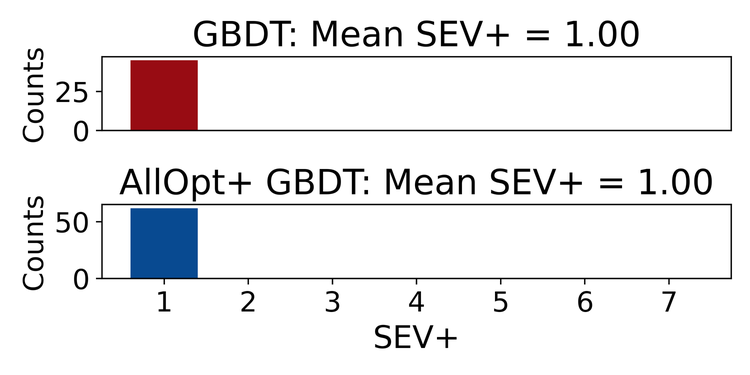}
         \caption{\sev{+} counts across methods}
     \end{subfigure}
     \hfill
     \begin{subfigure}[b]{0.35\textwidth}
         \centering
         \includegraphics[width=\textwidth]{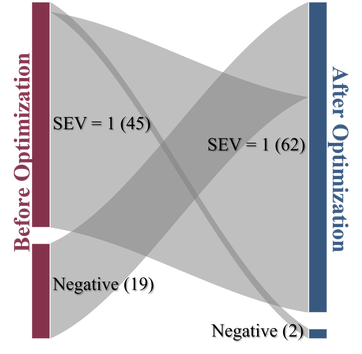}
         \caption{GBDT to \allopt{+} GBDT}
     \end{subfigure}
     \caption{\sev{+} optimization performance for gradient boosting trees on German Credit}
\end{figure}

\begin{figure}[!ht]
    \centering
    \begin{subfigure}[b]{0.45\textwidth}
         \centering
         \includegraphics[width=\textwidth]{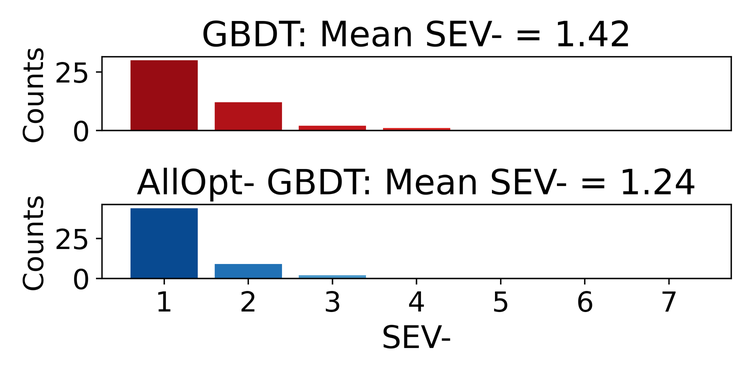}
         \caption{\sev{-} counts across methods}
     \end{subfigure}
     \hfill
     \begin{subfigure}[b]{0.35\textwidth}
         \centering
         \includegraphics[width=\textwidth]{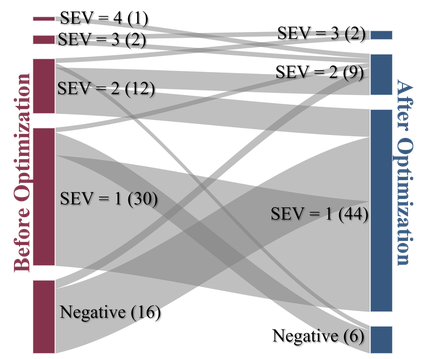}
         \caption{GBDT to \allopt{-} GBDT}
     \end{subfigure}
     \caption{\sev{-} optimization performance for gradient boosting trees on German Credit}
\end{figure}

\begin{figure}[!ht]
     \centering
     \begin{subfigure}[b]{0.4\textwidth}
         \centering
         \includegraphics[width=\textwidth]{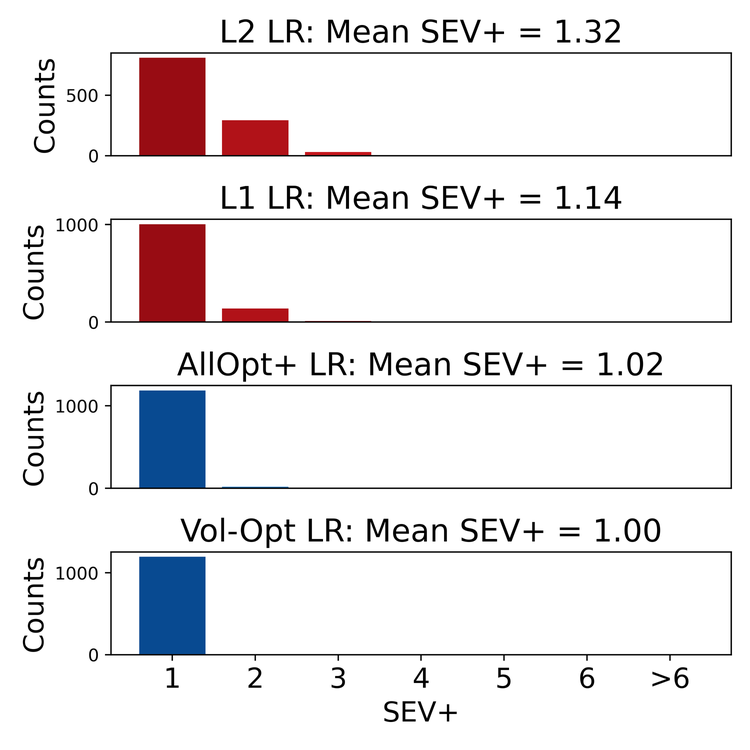}
         \caption{\sev{+} counts across methods}
     \end{subfigure}
     \hfill
     \begin{subfigure}[b]{0.4\textwidth}
         \centering
         \includegraphics[width=\textwidth]{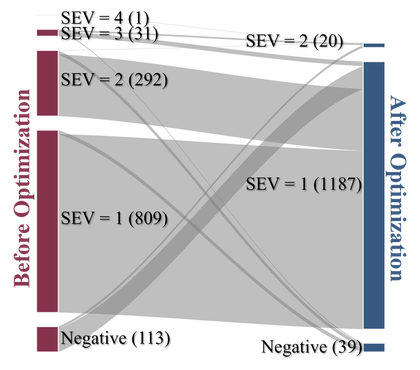}
         \caption{L2 LR \sev{+} to \allopt{+} LR}
     \end{subfigure}
     \caption{\sev{+} Optimization performance for linear classifiers on FICO}
        \label{fig:fico_lr_sev+}
\end{figure}
\begin{figure}[ht]
\centering
     \begin{subfigure}[b]{0.4\textwidth}
         \centering
         \includegraphics[width=\textwidth]{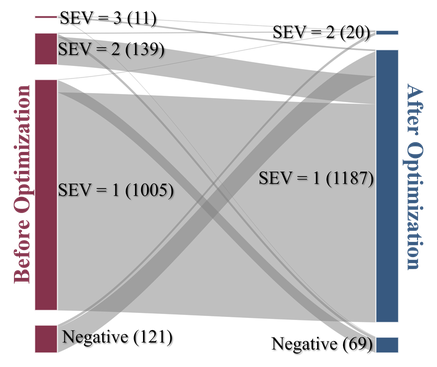}
         \caption{L1 LR \sev{+} to \allopt{+} LR}
     \end{subfigure}
     \hfill
     \begin{subfigure}[b]{0.4\textwidth}
         \centering
         \includegraphics[width=\textwidth]{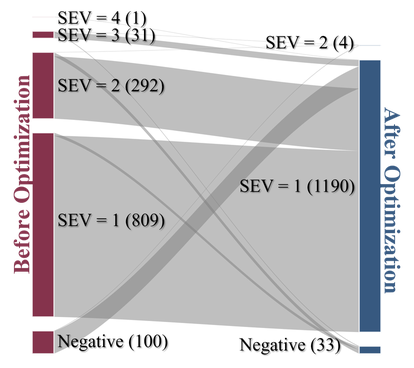}
         \caption{L2 LR \sev{+} to \volopt{} LR}
     \end{subfigure}
        \caption{\sev{+} Optimization performance for linear classifiers on FICO (cont.)}
        \label{fig:fico_lr_sev+_2}
\end{figure}

\begin{figure}[ht]
    \centering
    \begin{subfigure}[b]{0.3\textwidth}
         \centering
         \includegraphics[width=\textwidth]{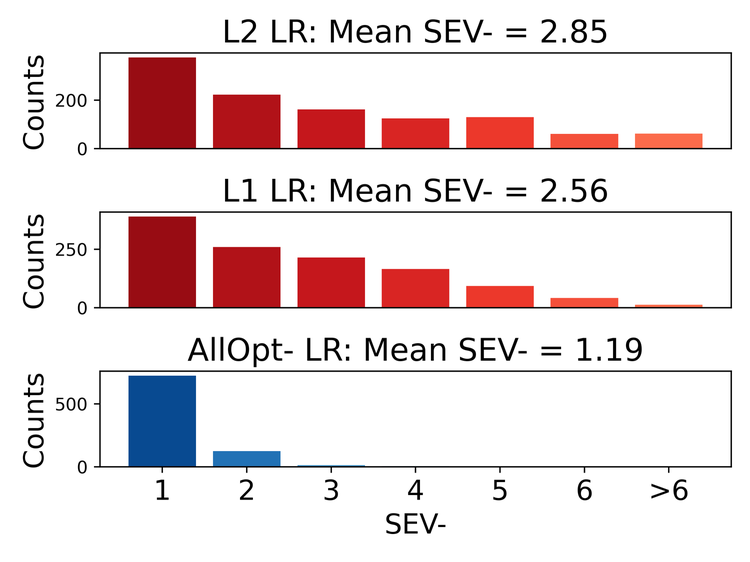}
         \caption{\sev{-} counts across methods}
     \end{subfigure}
     \hfill
    \begin{subfigure}[b]{0.3\textwidth}
         \centering
         \includegraphics[width=\textwidth]{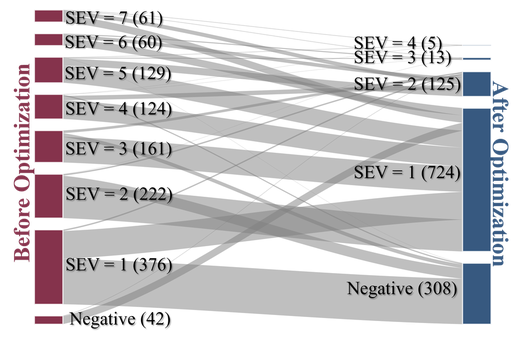}
         \caption{L2 LR \sev{-} to \allopt{-} LR}
     \end{subfigure}
     \hfill
     \begin{subfigure}[b]{0.3\textwidth}
         \centering
         \includegraphics[width=\textwidth]{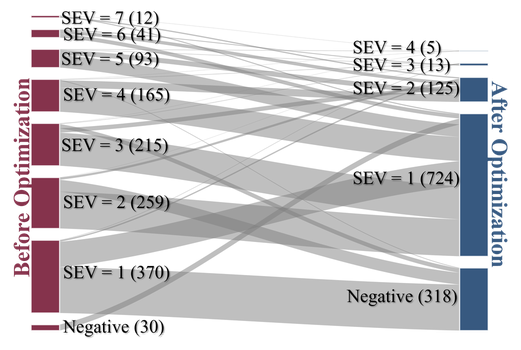}
         \caption{L1 LR \sev{-} to \allopt{-} LR}
     \end{subfigure}
     \caption{\sev{-} optimization performance for linear classifiers on FICO}
    \label{fig:fico_lr_sev-}
\end{figure}

\begin{figure}[!ht]
    \centering
    \begin{subfigure}[b]{0.45\textwidth}
         \centering
         \includegraphics[width=\textwidth]{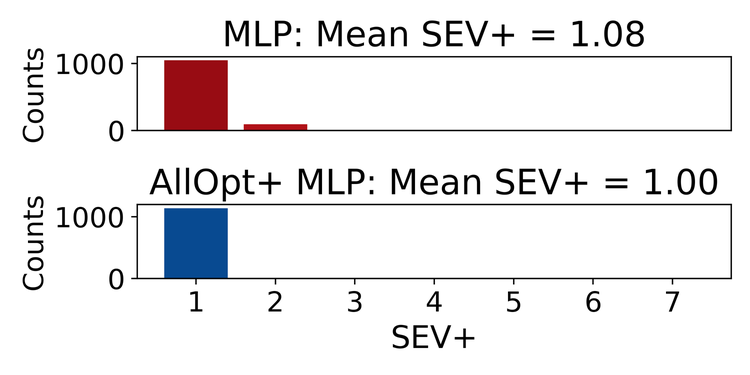}
         \caption{\sev{+} counts across methods}
     \end{subfigure}
     \hfill
     \begin{subfigure}[b]{0.35\textwidth}
         \centering
         \includegraphics[width=\textwidth]{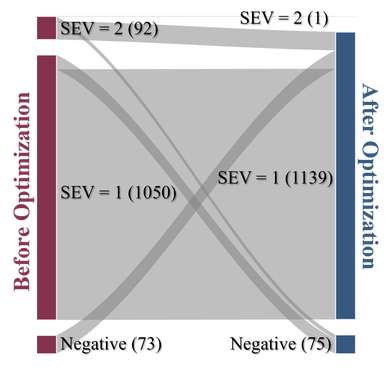}
         \caption{MLP to \allopt{+} MLP}
     \end{subfigure}
     \caption{\sev{+} optimization performance for multi-layer perceptions on FICO}
\end{figure}

\begin{figure}[ht]
    \centering
    \begin{subfigure}[b]{0.45\textwidth}
         \centering
         \includegraphics[width=\textwidth]{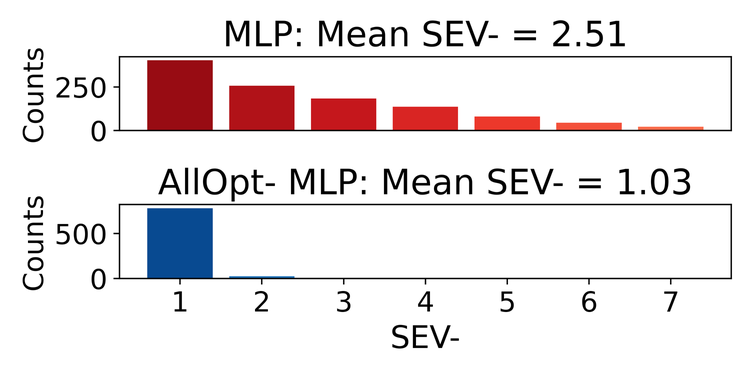}
         \caption{\sev{-} counts across methods}
     \end{subfigure}
     \hfill
     \begin{subfigure}[b]{0.35\textwidth}
         \centering
         \includegraphics[width=\textwidth]{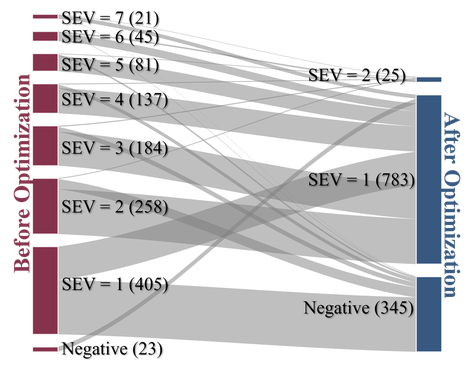}
         \caption{MLP to \allopt{-} MLP}
     \end{subfigure}
     \caption{\sev{-} optimization performance for multi-layer perceptions on FICO}
\end{figure}

\begin{figure}[!ht]
    \centering
    \begin{subfigure}[b]{0.45\textwidth}
         \centering
         \includegraphics[width=\textwidth]{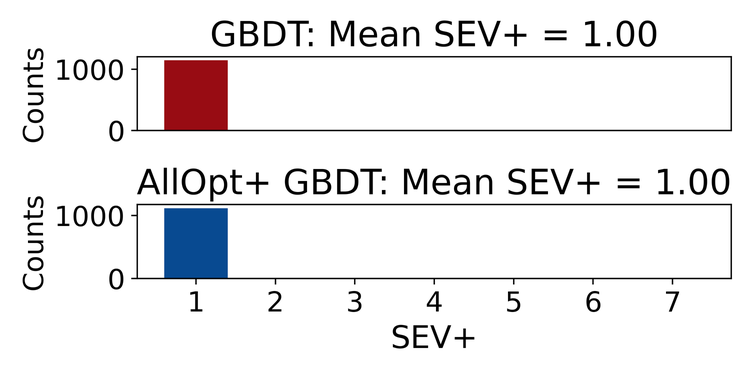}
         \caption{\sev{+} counts across methods}
     \end{subfigure}
     \hfill
     \begin{subfigure}[b]{0.35\textwidth}
         \centering
         \includegraphics[width=\textwidth]{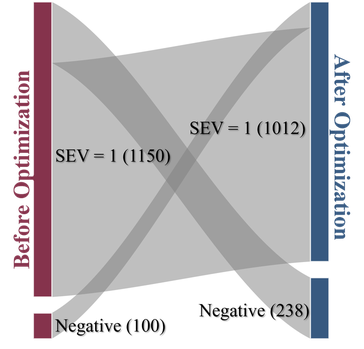}
         \caption{GBDT to \allopt{+} GBDT}
     \end{subfigure}
     \caption{\sev{+} optimization performance for gradient boosting trees on FICO}
\end{figure}

\begin{figure}[!ht]
    \centering
    \begin{subfigure}[b]{0.45\textwidth}
         \centering
         \includegraphics[width=\textwidth]{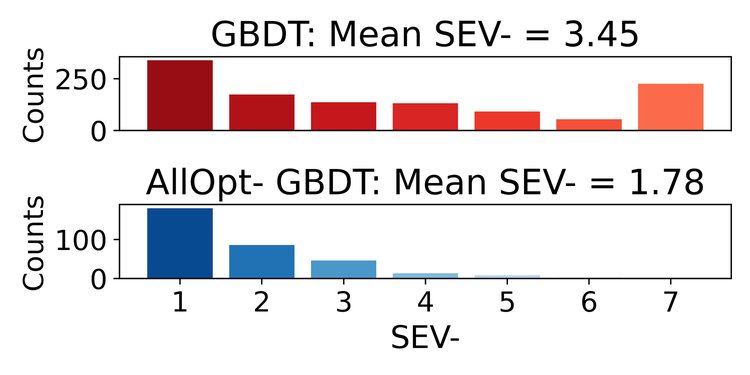}
         \caption{\sev{-} counts across methods}
     \end{subfigure}
     \hfill
     \begin{subfigure}[b]{0.35\textwidth}
         \centering
         \includegraphics[width=\textwidth]{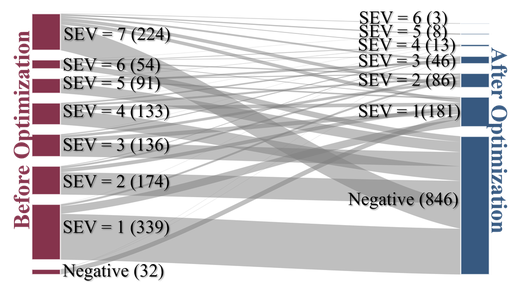}
         \caption{GBDT to \allopt{-} GBDT}
     \end{subfigure}
     \caption{\sev{-} optimization performance for gradient boosting trees on FICO}
\end{figure}

\begin{figure}[!ht]
     \centering
     \begin{subfigure}[b]{0.4\textwidth}
         \centering
         \includegraphics[width=\textwidth]{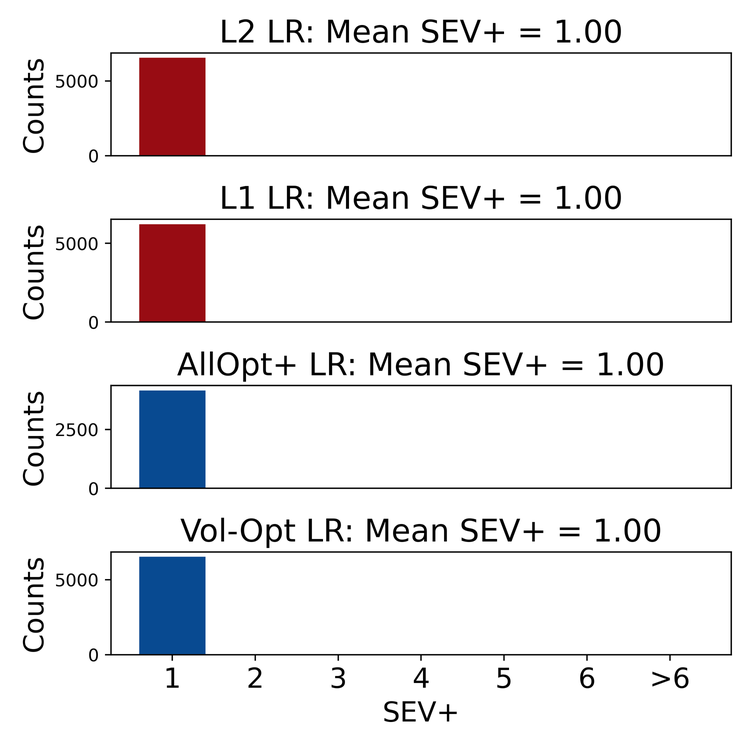}
         \caption{\sev{+} counts across methods}
     \end{subfigure}
     \hfill
     \begin{subfigure}[b]{0.4\textwidth}
         \centering
         \includegraphics[width=\textwidth]{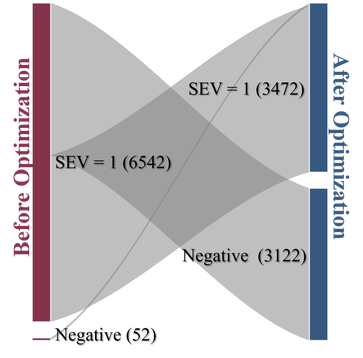}
         \caption{L2 LR \sev{+} to \allopt{+} LR}
     \end{subfigure}
     \caption{\sev{+} Optimization performance for linear classifiers on Diabetes}
        \label{fig:diabetes_lr_sev+}
\end{figure}
\begin{figure}[ht]
\centering
     \begin{subfigure}[b]{0.4\textwidth}
         \centering
         \includegraphics[width=\textwidth]{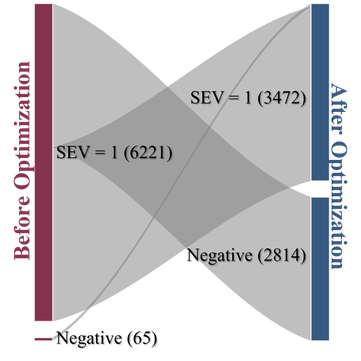}
         \caption{L1 LR \sev{+} to \allopt{+} LR}
     \end{subfigure}
     \hfill
     \begin{subfigure}[b]{0.4\textwidth}
         \centering
         \includegraphics[width=\textwidth]{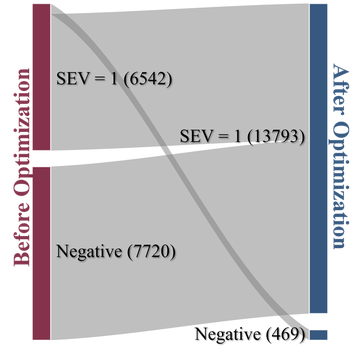}
         \caption{L2 LR \sev{+} to \volopt{} LR}
     \end{subfigure}
        \caption{\sev{+} Optimization performance for linear classifiers on Diabetes(cont.)}
        \label{fig:diabetes_lr_sev+_2}
\end{figure}

\begin{figure}[ht]
    \centering
    \begin{subfigure}[b]{0.3\textwidth}
         \centering
         \includegraphics[width=\textwidth]{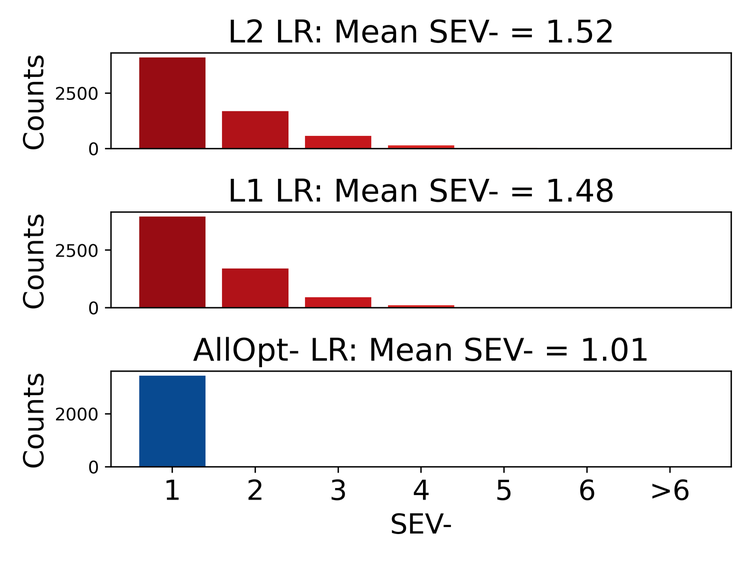}
         \caption{\sev{-} counts across methods}
     \end{subfigure}
     \hfill
    \begin{subfigure}[b]{0.3\textwidth}
         \centering
         \includegraphics[width=\textwidth]{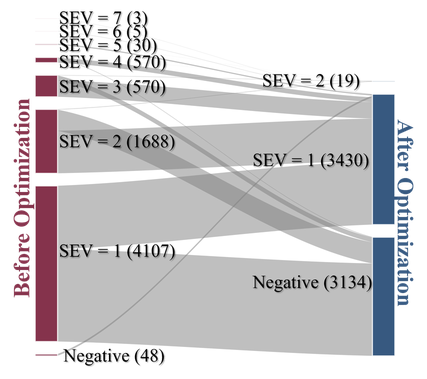}
         \caption{L2 LR \sev{-} to \allopt{-} LR}
     \end{subfigure}
     \hfill
     \begin{subfigure}[b]{0.3\textwidth}
         \centering
         \includegraphics[width=\textwidth]{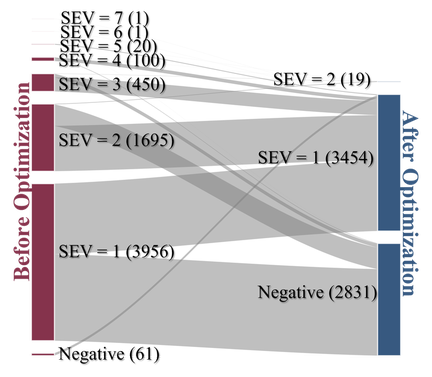}
         \caption{L1 LR \sev{-} to \allopt{-} LR}
     \end{subfigure}
     \caption{\sev{-} optimization performance for linear classifiers on Diabetes}
    \label{fig:diabetes_lr_sev-}
\end{figure}

\begin{figure}[!ht]
    \centering
    \begin{subfigure}[b]{0.45\textwidth}
         \centering
         \includegraphics[width=\textwidth]{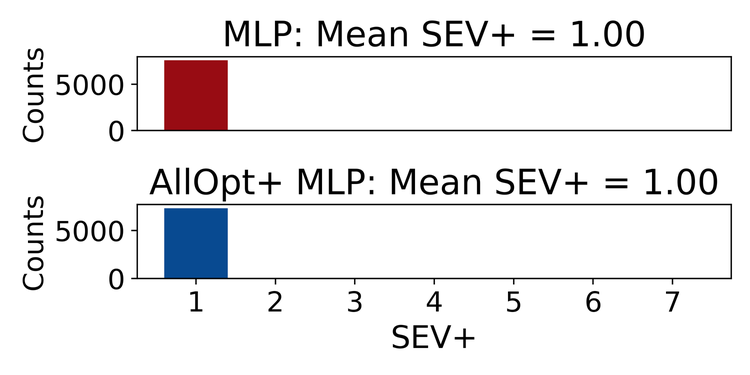}
         \caption{\sev{+} counts across methods}
     \end{subfigure}
     \hfill
     \begin{subfigure}[b]{0.35\textwidth}
         \centering
         \includegraphics[width=\textwidth]{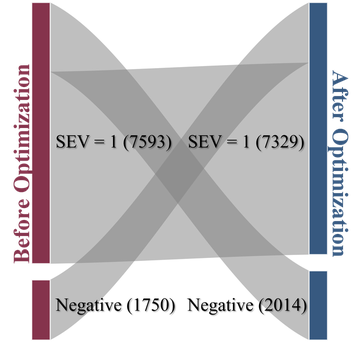}
         \caption{MLP to \allopt{+} MLP}
     \end{subfigure}
     \caption{\sev{+} optimization performance for multi-layer perceptions on Diabetes}
\end{figure}

\begin{figure}[ht]
    \centering
    \begin{subfigure}[b]{0.45\textwidth}
         \centering
         \includegraphics[width=\textwidth]{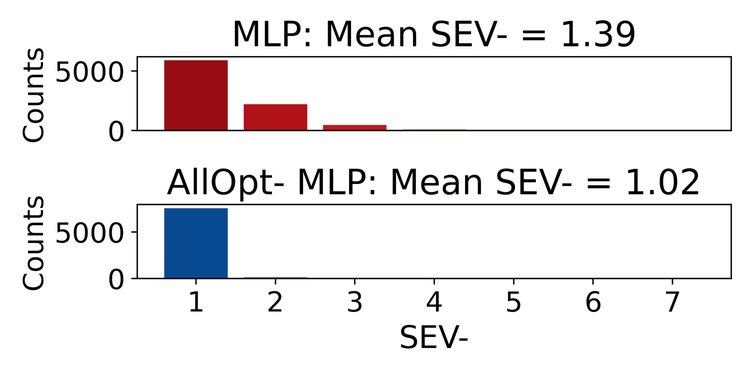}
         \caption{\sev{-} counts across methods}
     \end{subfigure}
     \hfill
     \begin{subfigure}[b]{0.35\textwidth}
         \centering
         \includegraphics[width=\textwidth]{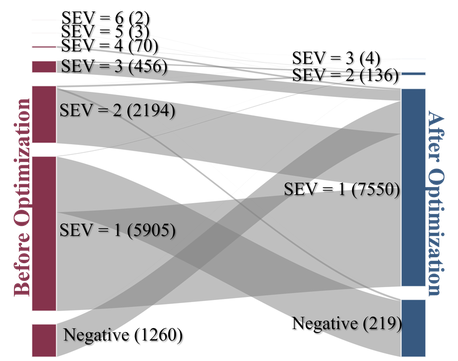}
         \caption{MLP to \allopt{-} MLP}
     \end{subfigure}
     \caption{\sev{-} optimization performance for multi-layer perceptions on Diabetes}
\end{figure}

\begin{figure}[!ht]
    \centering
    \begin{subfigure}[b]{0.45\textwidth}
         \centering
         \includegraphics[width=\textwidth]{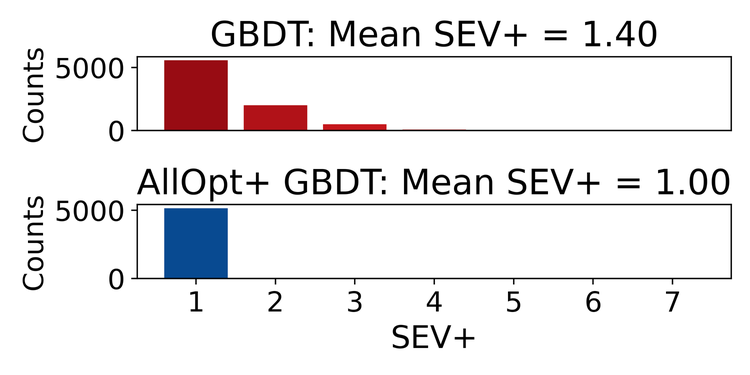}
         \caption{\sev{+} counts across methods}
     \end{subfigure}
     \hfill
     \begin{subfigure}[b]{0.35\textwidth}
         \centering
         \includegraphics[width=\textwidth]{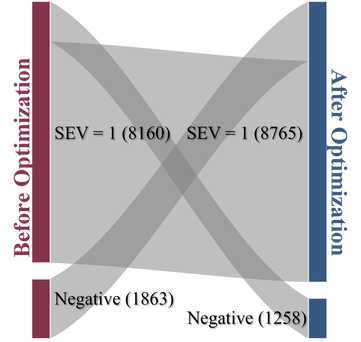}
         \caption{GBDT to \allopt{+} GBDT}
     \end{subfigure}
     \caption{\sev{+} optimization performance for gradient boosting trees on Diabetes}
\end{figure}

\begin{figure}[!ht]
    \centering
    \begin{subfigure}[b]{0.45\textwidth}
         \centering
         \includegraphics[width=\textwidth]{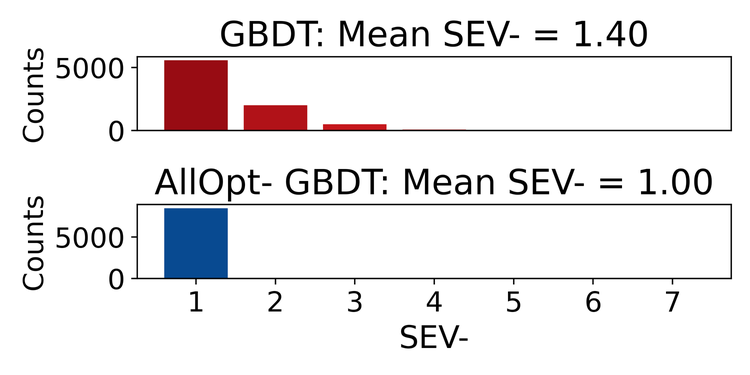}
         \caption{\sev{-} counts across methods}
     \end{subfigure}
     \hfill
     \begin{subfigure}[b]{0.35\textwidth}
         \centering
         \includegraphics[width=\textwidth]{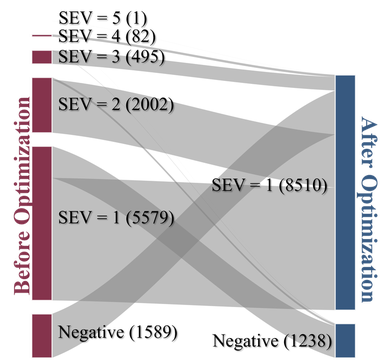}
         \caption{GBDT to \allopt{-} GBDT}
     \end{subfigure}
     \caption{\sev{-} optimization performance for gradient boosting trees on Diabetes}
\end{figure}

\fi
\clearpage
\subsection{Local Explanations Comparison}
\label{subsec:local_explanation_appendix}

In this section, we show how LIME-C, SHAP-C, DiCE, and \sev{-} each explain a fixed GBDT reference classifier across different datasets. These plots are analogous to those in Figure \ref{fig:flip_xai}. These figures do not include models optimized for \sev{}.

\begin{figure}[!ht]
    \centering
    \begin{subfigure}[b]{0.33\textwidth}
        \centering
        \includegraphics[width=\textwidth]{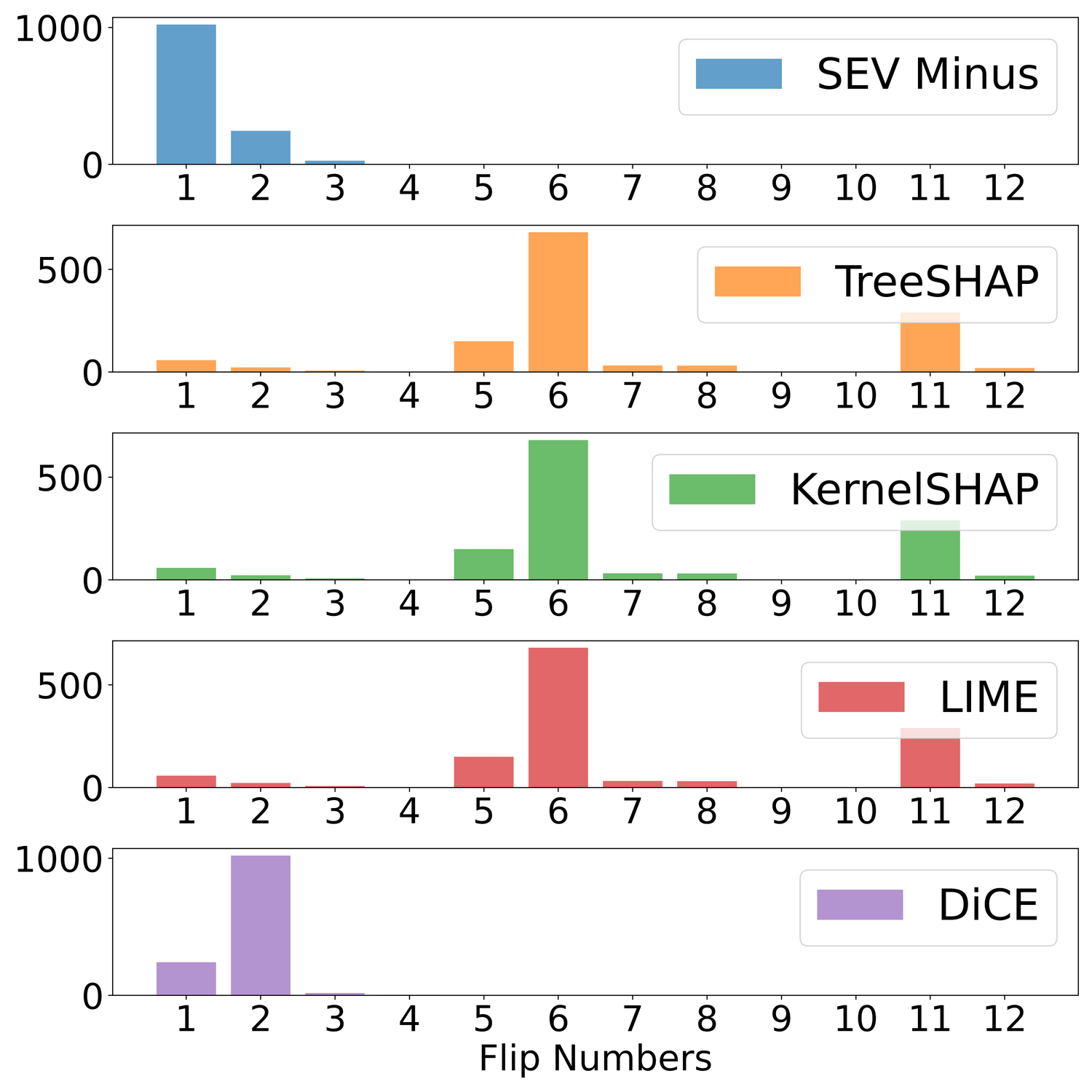}
        \caption{Adult Dataset}
    \end{subfigure}
    \hfill
    \begin{subfigure}[b]{0.33\textwidth}
        \centering
        \includegraphics[width=\textwidth]{AISTATS_Submission/fig_lowdpi/Experiment_C2/flip_compas_1.png}
        \caption{COMPAS Dataset}
    \end{subfigure}
    \begin{subfigure}[b]{0.33\textwidth}
        \centering
        \includegraphics[width=\textwidth]{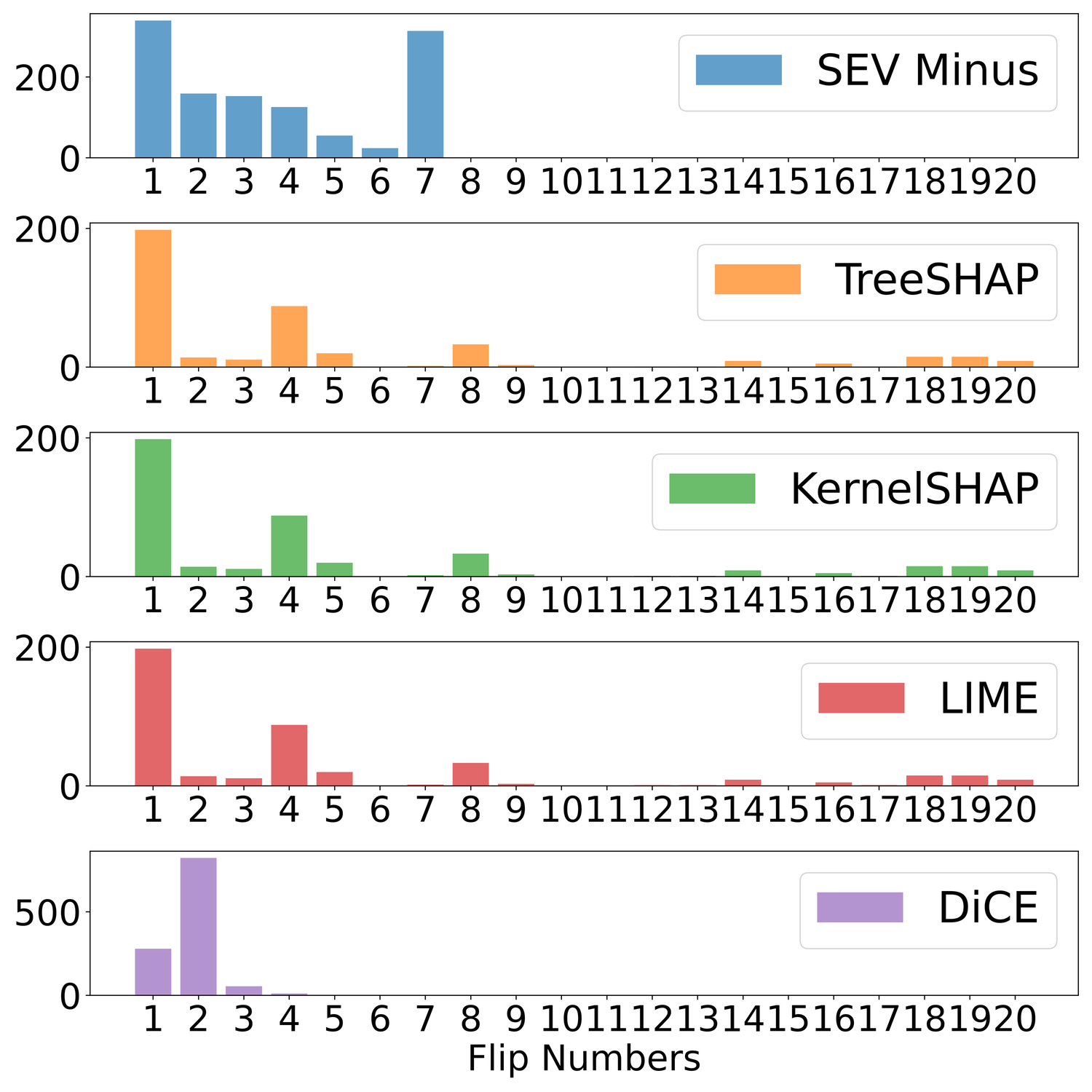}
        \caption{FICO Dataset}
    \end{subfigure}
    \label{fig:flip_total_counts_xai_1}
    \centering
    \begin{subfigure}[b]{0.32\textwidth}
        \centering
        \includegraphics[width=\textwidth]{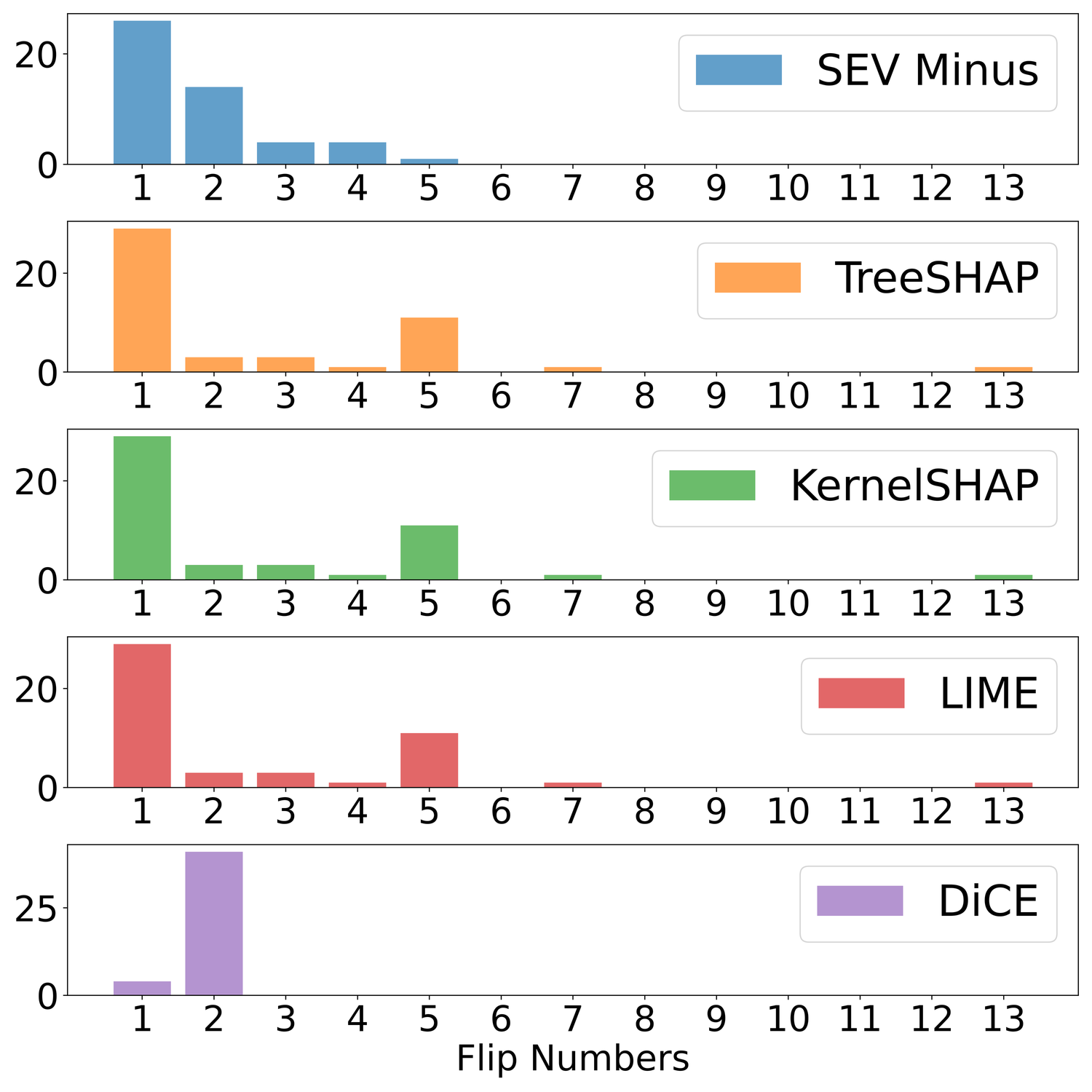}
        \caption{German Credit Dataset}
    \end{subfigure}
    \begin{subfigure}[b]{0.32\textwidth}
        \centering
        \includegraphics[width=\textwidth]{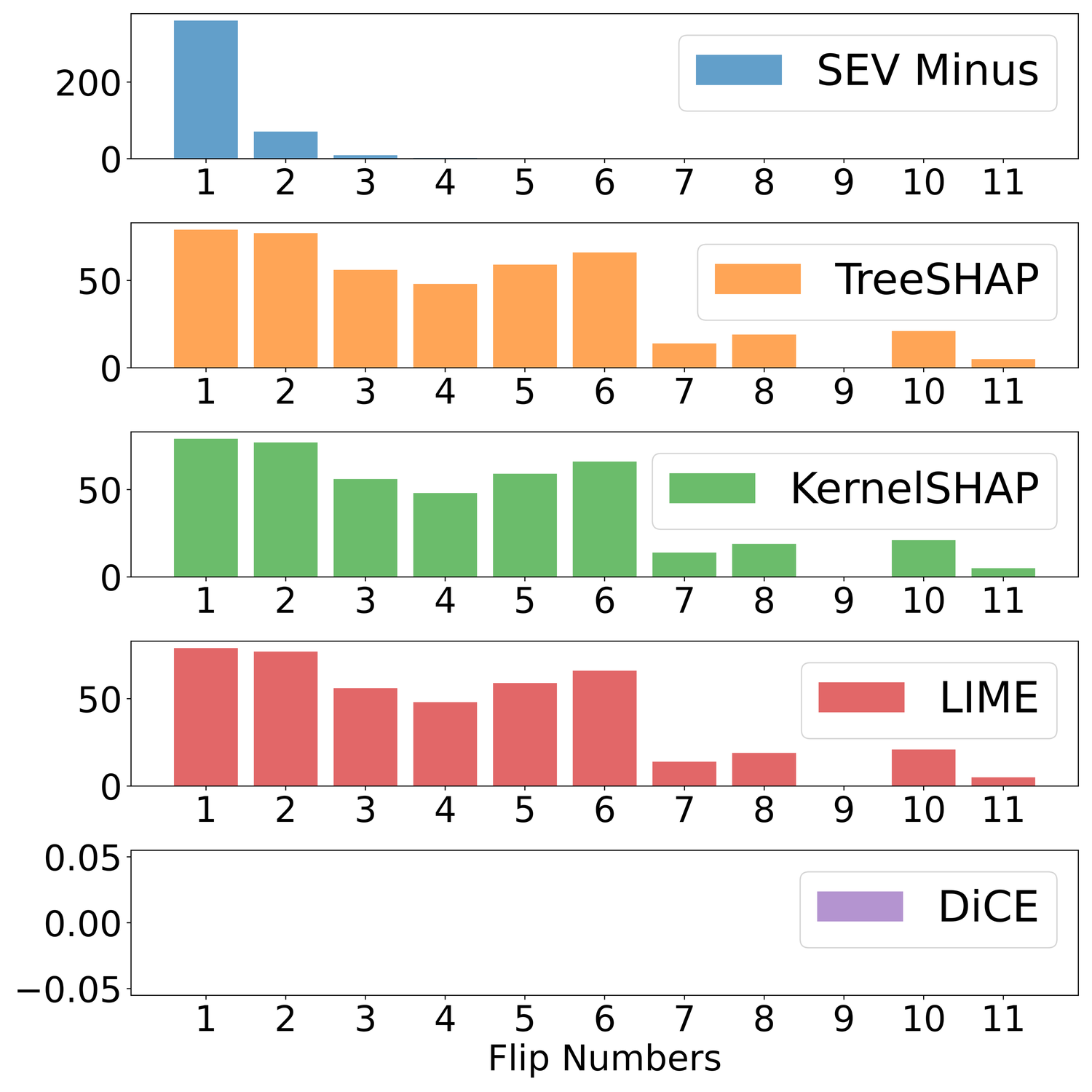}
        \caption{MIMIC Dataset*}
    \end{subfigure}
    \hfill
    \begin{subfigure}[b]{0.32\textwidth}
        \centering
        \includegraphics[width=\textwidth]{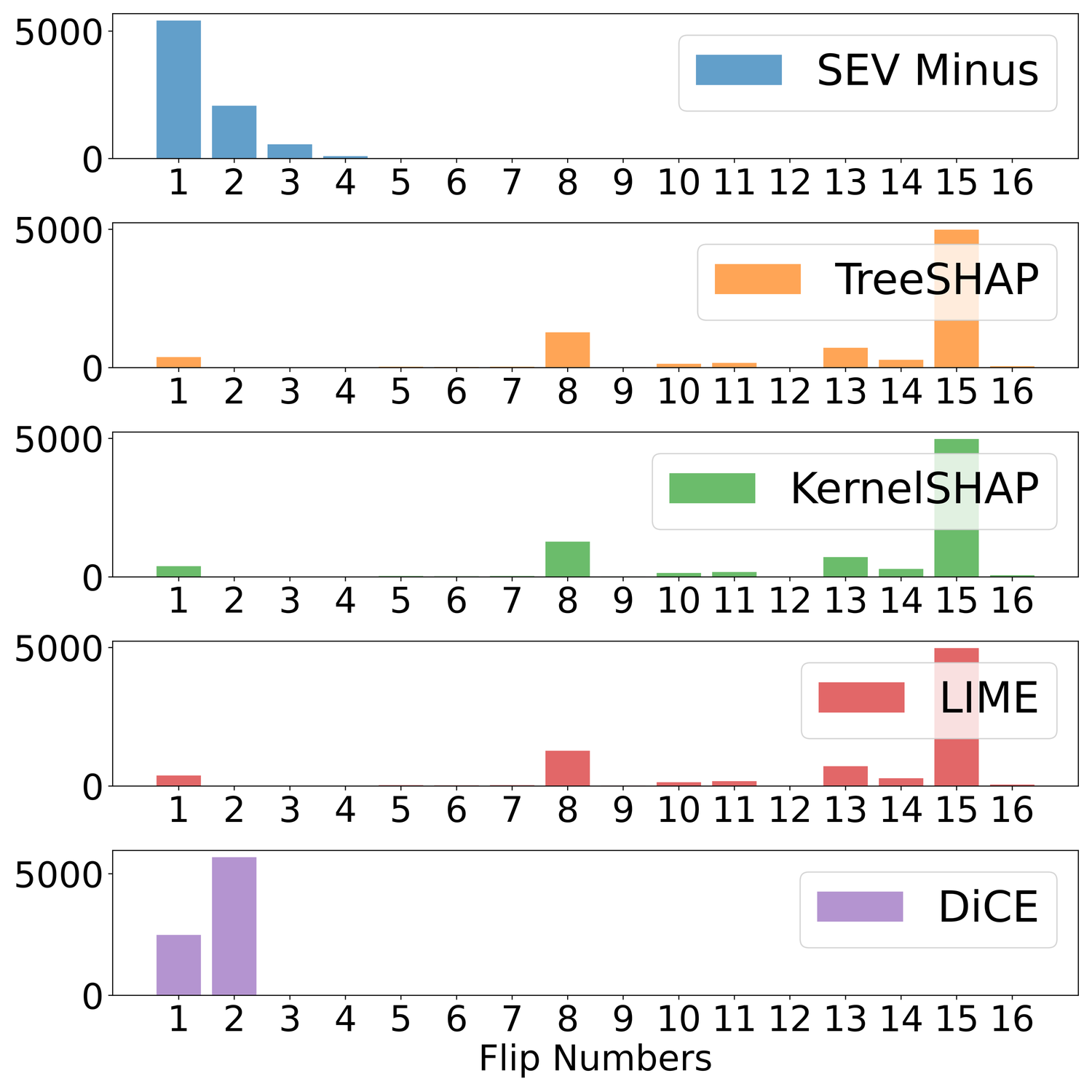}
        \caption{Diabetes Dataset}
    \end{subfigure}
    \caption{The flip counts for different local explanations in different datasets (continued)}
    \label{fig:flip_total_counts_xai_2}
\end{figure}

* Values for DiCE are not plotted for MIMIC, as after one week of running it still had not generated counterfactual explanations. Additional details are shown in Table \ref{fig:local_explanation_time}. Figure \ref{fig:flip_total_counts_xai_2} shows that the local feature importance scores created by LIME-C and SHAP-C do not lead to sparse explanations. Some of the explanations need more than 5 features to explain one prediction. \sev{-} and DiCE, on the other hand, do generate sparse explanations.

\newpage
\section{Local Explanation Time Comparsion between Kernel SHAP and \sev{-}}
\label{sec:sev_scability}

Here, we compare the time required to generate a local explanation for our method and KernelSHAP. We use \textbf{GBDT} as the baseline model and evaluate the median explanation runtime on queries in the test set. 

Figure \ref{fig:time_distribution_compare} and Table \ref{tab:time_compare_shap_sev} give additional information on the runtime distribution across queries for KernelSHAP and \sev{-}. We see that for 3 out of the 6 datasets in which KernelSHAP is faster than \sev{-} for explaining all queries, the total \sev{-} runtime is skewed by a small number of queries. In all cases, with \sev{-}, a greater proportion of queries can be explained in less than 0.1 seconds than with KernelSHAP. 
\ifshow
\begin{figure}[ht]
    \centering
    \begin{subfigure}[b]{0.47\textwidth}
        \centering
        \includegraphics[width=\textwidth]{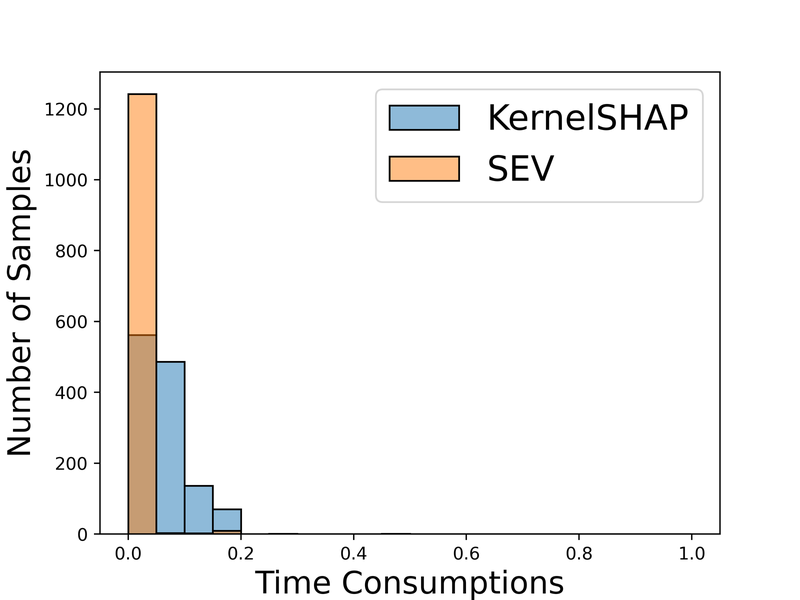}
        \caption{Adult Dataset}
    \end{subfigure}
    \hfill
    \begin{subfigure}[b]{0.47\textwidth}
        \centering
        \includegraphics[width=\textwidth]{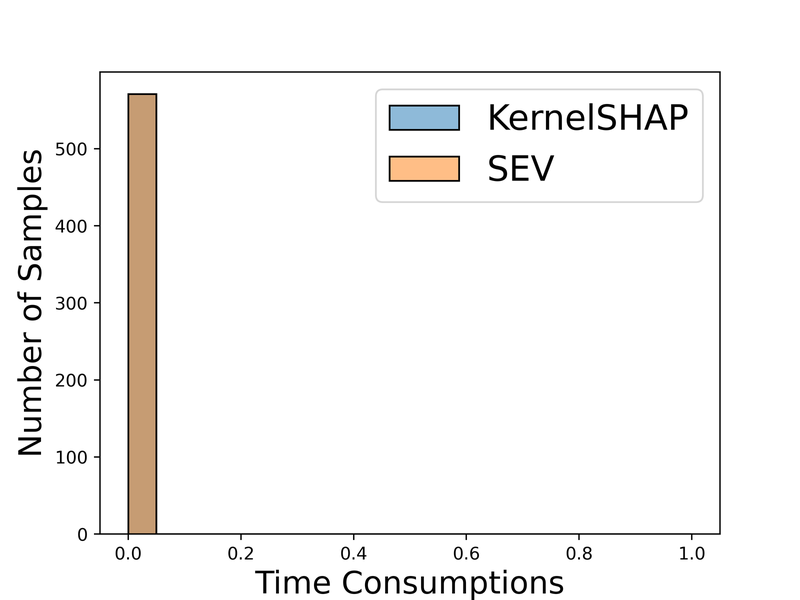}
        \caption{COMPAS Dataset}
    \end{subfigure}
    \begin{subfigure}[b]{0.47\textwidth}
        \centering
        \includegraphics[width=\textwidth]{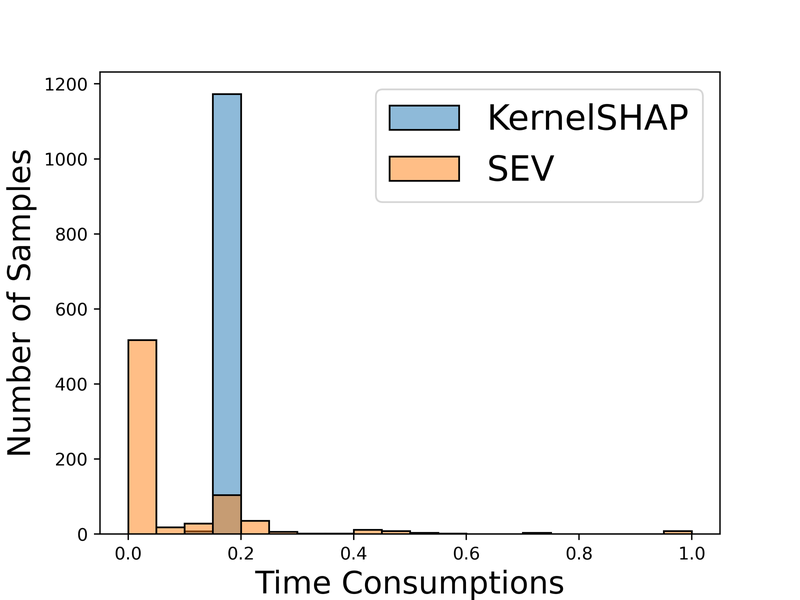}
        \caption{FICO Dataset}
    \end{subfigure}
    \hfill
    \begin{subfigure}[b]{0.47\textwidth}
        \centering
        \includegraphics[width=\textwidth]{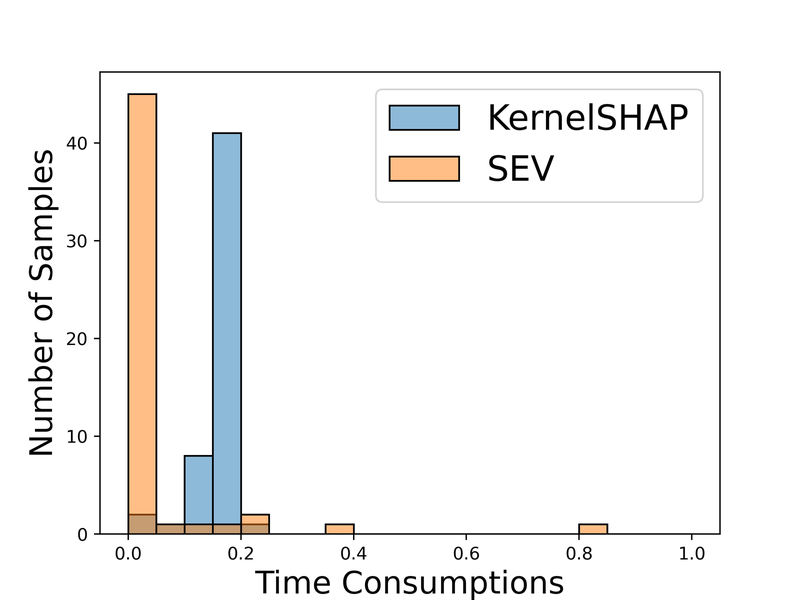}
        \caption{German Dataset}
    \end{subfigure}
    \begin{subfigure}[b]{0.47\textwidth}
        \centering
        \includegraphics[width=\textwidth]{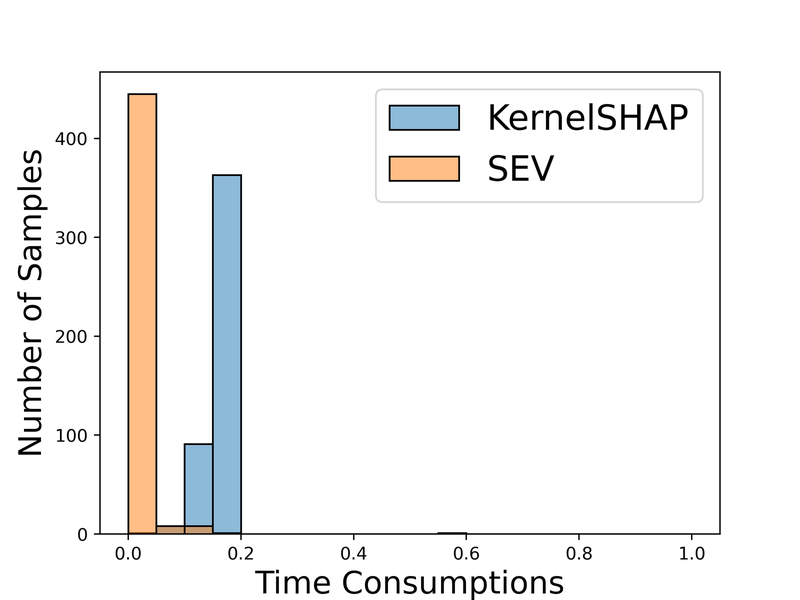}
        \caption{MIMIC Dataset}
    \end{subfigure}
    \hfill
    \begin{subfigure}[b]{0.47\textwidth}
        \centering
        \includegraphics[width=\textwidth]{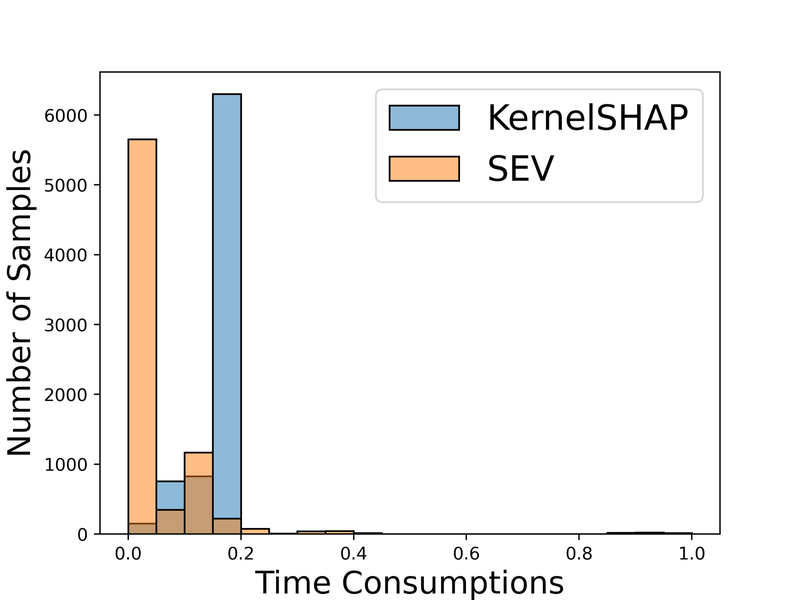}
        \caption{Diabetes Dataset}
    \end{subfigure}
    \caption{Time consumption distribution between KernelSHAP and \sev{-} less than 1 second}
    \label{fig:time_distribution_compare}
\end{figure}

\begin{table*}[ht]
\small
\centering
\scalebox{1}{%
\begin{tabular}{cccccc}
\hline
\textbf{Dataset} & \textbf{Methods} & \textbf{\begin{tabular}[c]{@{}c@{}}max single \\ query \\ explanation time\end{tabular}} & \textbf{\begin{tabular}[c]{@{}c@{}}\% of query\\ explanation time\\ less than 0.1s\end{tabular}} & \textbf{\begin{tabular}[c]{@{}c@{}}\% of query\\ explanation time\\ less than 0.5s\end{tabular}} & \textbf{\begin{tabular}[c]{@{}c@{}}\% of query\\ explanation time\\ less than 1s\end{tabular}} \\ \hline
\multirow{2}{*}{\textbf{Adult}} & \sev{-} & 0.18 & 99.12 & 100.00 & 100.00 \\
 & KernelSHAP & 0.48 & \textcolor{red}{83.44} & 100.00 & 100.00 \\
\multirow{2}{*}{\textbf{German}} & \sev{-} & 0.83 & 88.46 & 98.08 & 100.00 \\
 & KernelSHAP & 0.20 & \textcolor{red}{5.66} & 100.00 & 100.00 \\
\multirow{2}{*}{\textbf{MIMIC}} & \sev{-} & 0.58 & 97.84 & 99.78 & 100.00 \\
 & KernelSHAP & 0.18 & \textcolor{red}{1.94} & 100.00 & 100.00 \\
\multirow{2}{*}{\textbf{FICO$^{*}$}} & \sev{-} & 69.05 & 45.34 & 61.78 & 63.05 \\
 & KernelSHAP & 0.19 & \textcolor{red}{0.00} & 100.00 & 100.00 \\
\multirow{2}{*}{\textbf{COMPAS}} & \sev{-} & 0.01 & 100.00 & 100.00 & 100.00 \\
 & KernelSHAP & 0.00 & 100.00 & 100.00 & 100.00 \\
\multirow{2}{*}{\textbf{Diabetes}} & \sev{-} & 103.03 & 74.70 & 94.25 & 94.93 \\
 & KernelSHAP & 0.20 & \textcolor{red}{11.26} & 100.00 & 100.00 \\ \hline
\end{tabular}}
\caption{Time comparison between KernelSHAP and \sev{-}. The red parts emphasize that KernelSHAP has a smaller proportion of queries explained less than 0.1s. (*) For the FICO dataset, we set the max depth for \sev{-} search to be 6, and we observe that more than half (61.78\%) of the queries take less than 0.5s to generate the \sev{-} explanations.}
\label{tab:time_compare_shap_sev}
\end{table*}

\fi
\newpage
\clearpage
\section{The performance of \allopt{\circledR}}
\label{sec:sev_r_explanations}

This section will show that \allopt{\circledR} can be applied in different datasets. We use \textbf{LR} as the baseline model and compare the \sev{\circledR} distribution, and the proportion of unreachable queries between \textbf{L2 LR} and \textbf{\allopt{\circledR} LR} for the Adult, COMPAS, MIMIC, and German Credit datasets. For Adult Dataset, \texttt{age}, \texttt{marital-status}, \texttt{relationship}, \texttt{race}, \texttt{sex}, \texttt{native-country} and \texttt{occupation} are considered as restricted features. For MIMIC Dataset, \texttt{age} and \texttt{preiculos} are considered restricted features. For COMPAS Dataset, \texttt{gender=female} and \texttt{age} are restricted features. For the German Credit Dataset, \texttt{Age}, \texttt{Personal-status-sex}, and \texttt{Job} are considered restricted. The results for test Accuracy, AUC, mean \sev{\circledR}, and the proportion of unreachable cases in explanations are shown in Tables \ref{tab:alloptr_summary_adult}, \ref{tab:alloptr_summary_mimic}, \ref{tab:alloptr_summary_compas}, and \ref{tab:alloptr_summary_german}. For unexplainable queries, we consider the sparse explanations as the number of features used in the model. Figures \ref{fig:distribution_sev_r_adult}, \ref{fig:distribution_sev_r_mimic}, \ref{fig:distribution_sev_r_compas}, and \ref{fig:distribution_sev_r_german} show the SEV distribution changes before and after optimization. Figures \ref{fig:feature_r_adult}, \ref{fig:feature_r_mimic}, \ref{fig:feature_r_compas}, and \ref{fig:feature_r_german} show the proportion of features used in each query before optimization and the restricted usage of features after optimization (restricted features are marked as \textcolor{red}{red}). All of the results show that \allopt{\circledR} performs well in avoiding unexplainable queries and can still effectively lower SEV.
\ifshow
\begin{figure}[ht]
    \centering
    \begin{subfigure}[b]{0.3\textwidth}
        \centering
        \small
        \begin{tabular}{ccc}
        \hline
        \textbf{} & \textbf{\begin{tabular}[c]{@{}c@{}}Before \\ Optimization\end{tabular}} & \textbf{\begin{tabular}[c]{@{}c@{}}After\\ Optimization\end{tabular}} \\ \hline
        \textbf{Test Acc} & $0.85\pm0.00$ & $0.84\pm0.00$ \\
        \textbf{Test AUC} & $0.90\pm0.00$ & $0.90\pm0.00$ \\
        \textbf{Mean \sev{\circledR}} & $2.04\pm0.06$ & $1.02\pm0.01$ \\
        \textbf{\makecell{Unexplainable\\ queries(\%)}} & $5.40\pm0.95$ & $0$ \\ \hline
        \end{tabular}
        \caption{\allopt{\circledR} Summary Result}
        \label{tab:alloptr_summary_adult}
    \end{subfigure}
    \hfill
    \begin{subfigure}[b]{0.45\textwidth}
        \centering
        \includegraphics[width=\textwidth]{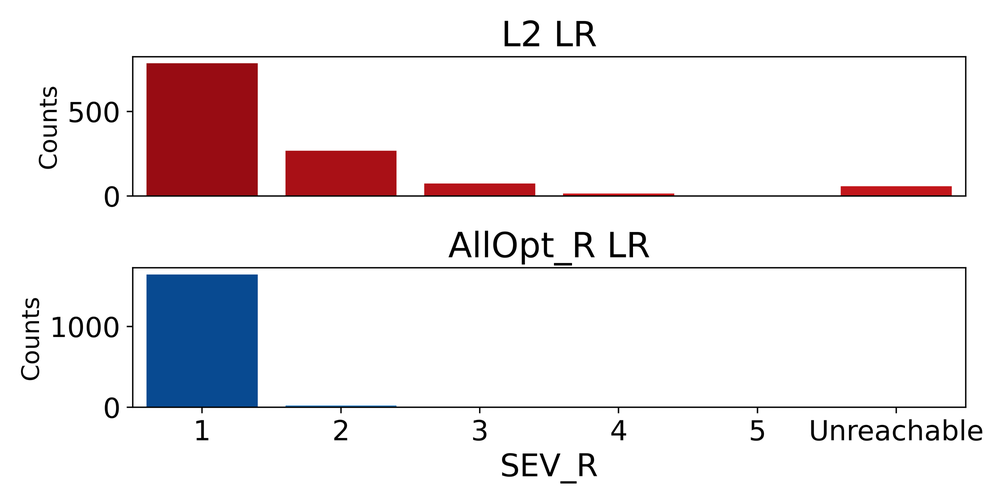}
        \caption{\sev{\circledR} Distribution with and without \allopt{\circledR}}
        \label{fig:distribution_sev_r_adult}
    \end{subfigure}
    \hfill
    \begin{subfigure}[b]{0.6\textwidth}
        \centering
        \includegraphics[width=\textwidth]{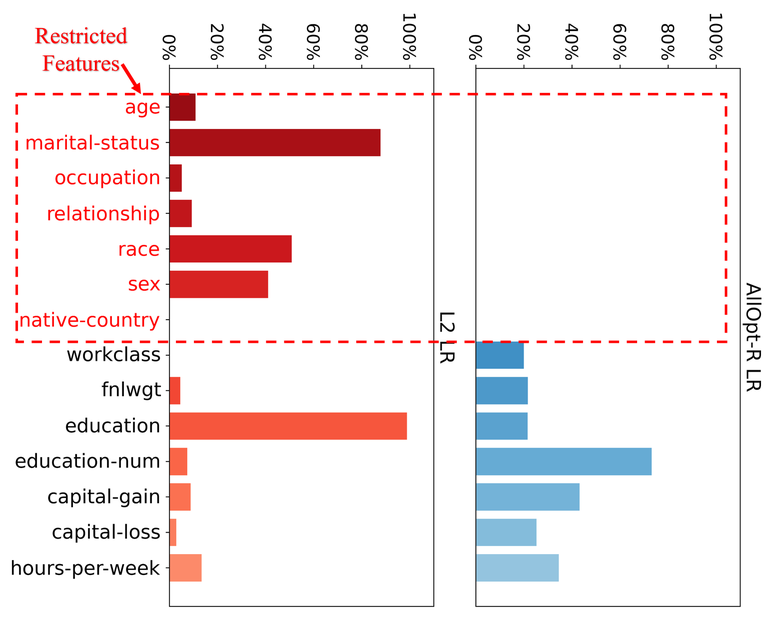}
        \caption{Feature used in explanations( \%)}
        \label{fig:feature_r_adult}
    \end{subfigure}
    \caption{\allopt{\circledR} performance on Adult Dataset}
    \label{fig:adult_restrictd_r}
\end{figure}

\begin{figure}[ht]
    \centering
    \begin{subfigure}[b]{0.45\textwidth}
        \centering
        \small
        \begin{tabular}{ccc}
        \hline
        \textbf{} & \textbf{\begin{tabular}[c]{@{}c@{}}Before \\ Optimization\end{tabular}} & \textbf{\begin{tabular}[c]{@{}c@{}}After\\ Optimization\end{tabular}} \\ \hline
        \textbf{Test Acc} & $0.89\pm0.00$ & $0.89\pm0.00$ \\
        \textbf{Test AUC} & $0.80\pm0.01$ & $0.74\pm0.04$ \\
        \textbf{Mean \sev{\circledR}} & $1.26\pm0.0$ & $1.05\pm0.04$ \\
        \textbf{\makecell{Unexplainable\\ queries(\%)}} & $0.35\pm0.32$ & $0$ \\ \hline
        \end{tabular}
        \caption{\allopt{\circledR} Summary Result}
        \label{tab:alloptr_summary_mimic}
    \end{subfigure}
    \hfill
    \begin{subfigure}[b]{0.45\textwidth}
        \centering
        \includegraphics[width=\textwidth]{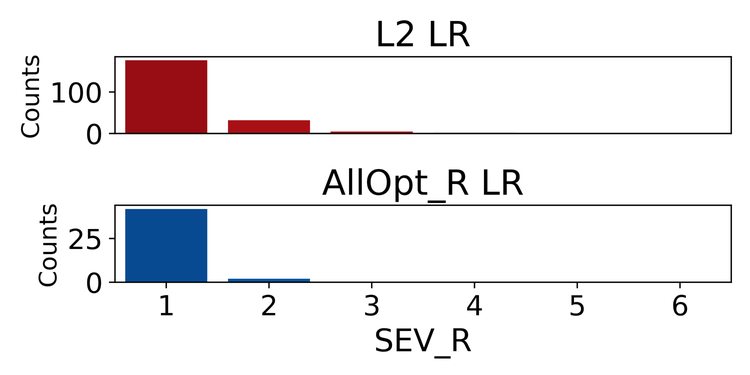}
        \caption{\sev{\circledR} Distribution with and without\allopt{\circledR}}
        \label{fig:distribution_sev_r_mimic}
    \end{subfigure}
    \hfill
    \begin{subfigure}[b]{0.4\textwidth}
        \centering
        \includegraphics[width=\textwidth]{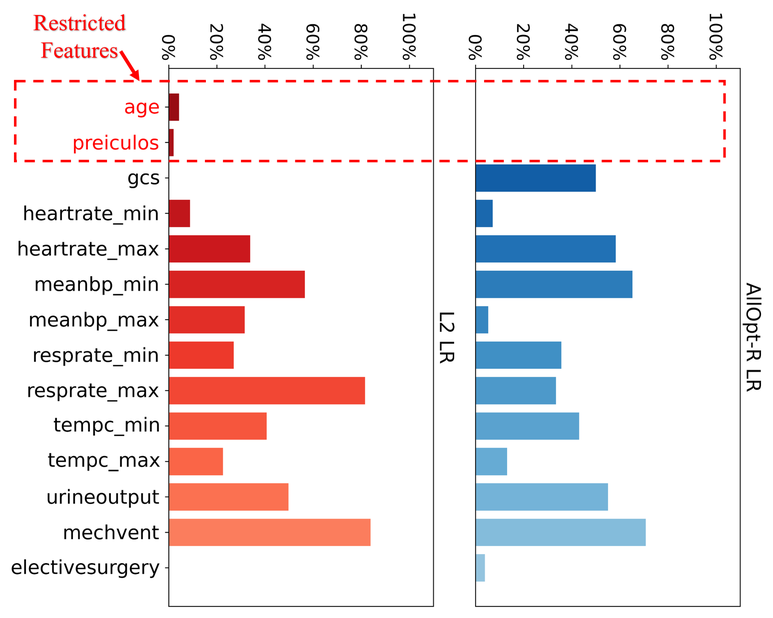}
        \caption{Feautre used in explanations(\%)}
        \label{fig:feature_r_mimic}
    \end{subfigure}
    \caption{\allopt{\circledR} performance in MIMIC Dataset}
    \label{fig:mimic_restrictd_r}
\end{figure}

\begin{figure}[!ht]
    \centering
    \begin{subfigure}[b]{0.45\textwidth}
        \centering
        \small
        \begin{tabular}{ccc}
        \hline
        \textbf{} & \textbf{\begin{tabular}[c]{@{}c@{}}Before \\ Optimization\end{tabular}} & \textbf{\begin{tabular}[c]{@{}c@{}}After\\ Optimization\end{tabular}} \\ \hline
        \textbf{Test Acc} & $0.68\pm0.01$ & $0.65\pm0.01$ \\
        \textbf{Test AUC} & $0.73\pm0.01$ & $0.72\pm0.01$ \\
        \textbf{Mean \sev{\circledR}} & $3.20\pm0.26$ & $1.28\pm0.38$ \\
        \textbf{\makecell{Unexplainable\\ queries(\%)}} & $34.4\pm4.4$ & $3.35\pm6.2$ \\ \hline
        \end{tabular}
        \caption{\allopt{\circledR} Summary Result}
        \label{tab:alloptr_summary_compas}
    \end{subfigure}
    \hfill
    \begin{subfigure}[b]{0.45\textwidth}
        \centering
        \includegraphics[width=\textwidth]{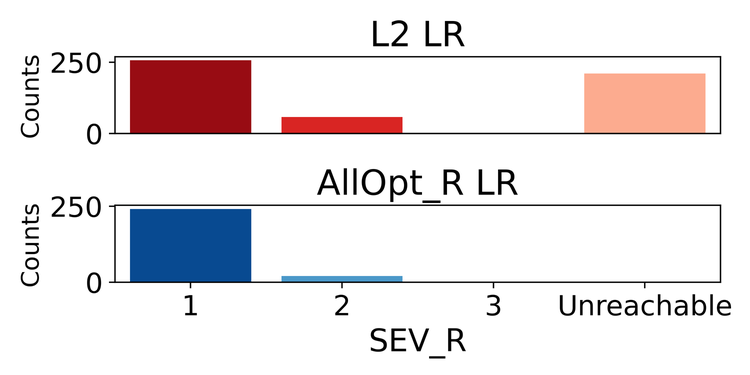}
        \caption{\sev{\circledR} Distribution with and without \allopt{\circledR}}
        \label{fig:distribution_sev_r_compas}
    \end{subfigure}
    \hfill
    \begin{subfigure}[b]{0.4\textwidth}
        \centering
        \includegraphics[width=\textwidth]{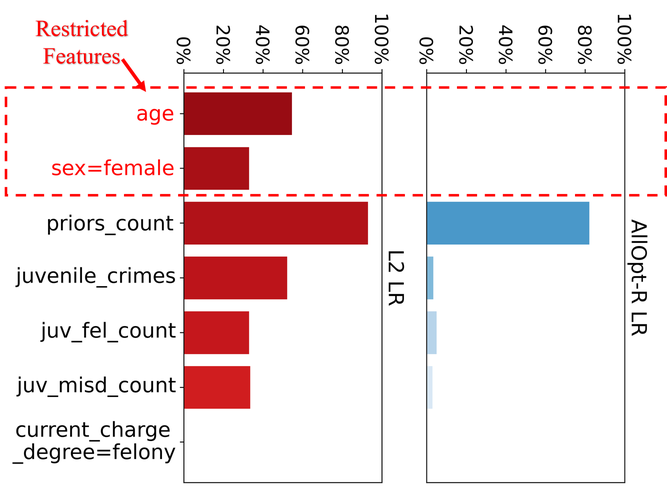}
        \caption{Feautre used in explanations(\%)}
        \label{fig:feature_r_compas}
    \end{subfigure}
    \caption{\allopt{\circledR} performance in COMPAS Dataset}
    \label{fig:compas_restrictd_r}
\end{figure}

\begin{figure}[!ht]
    \centering
    \begin{subfigure}[b]{0.3\textwidth}
        \centering
        \small
        \begin{tabular}{ccc}
        \hline
        \textbf{} & \textbf{\begin{tabular}[c]{@{}c@{}}Before \\ Optimization\end{tabular}} & \textbf{\begin{tabular}[c]{@{}c@{}}After\\ Optimization\end{tabular}} \\ \hline
        \textbf{Test Acc} & $0.74\pm0.03$ & $0.76\pm0.03$ \\
        \textbf{Test AUC} & $0.79\pm0.05$ & $0.78\pm0.04$ \\
        \textbf{Mean \sev{\circledR}} & $1.27\pm0.05$ & $1.05\pm0.06$ \\
        \textbf{\makecell{Unexplainable\\ Queries(\%)}} & $0$ & $0$ \\ \hline
        \end{tabular}
        \caption{\allopt{\circledR} Summary Result}
        \label{tab:alloptr_summary_german}
    \end{subfigure}
    \hfill
    \begin{subfigure}[b]{0.45\textwidth}
        \centering
        \includegraphics[width=\textwidth]{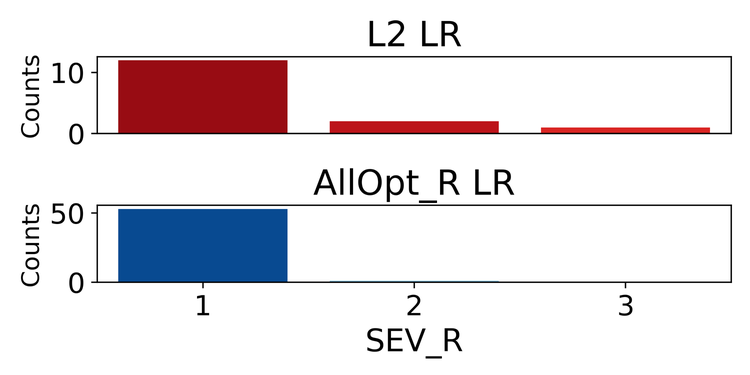}
        \caption{\sev{\circledR} Distribution under \allopt{\circledR}}
        \label{fig:distribution_sev_r_german}
    \end{subfigure}
    \hfill
    \begin{subfigure}[b]{0.4\textwidth}
        \centering
        \includegraphics[width=\textwidth]{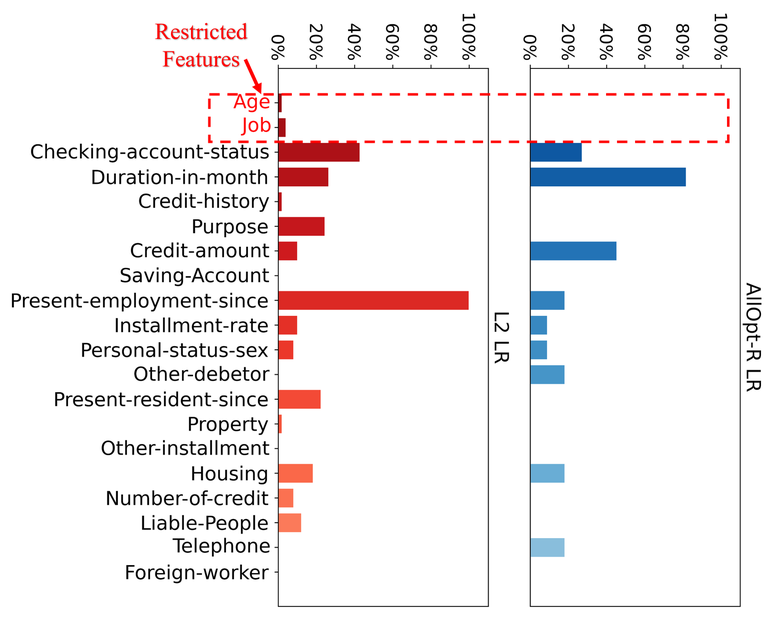}
        \caption{Feautre used in explanations(\%)}
        \label{fig:feature_r_german}
    \end{subfigure}
    \caption{\allopt{\circledR} performance in German Credit Dataset}
    \label{fig:german_restrictd_r}
\end{figure}
\fi
\clearpage
\section{Proof of Theorem \ref{thm:volume}}
\label{sec:theorem41_proof}

\textbf{Theorem} \ref{thm:volume} 


Consider a linear classifier, $
    f(\bx):= \mathbf{1}[(\beta_0 + \sum_{j=1}^p \beta_j x_j)>0]
$, where $\forall j$, $\beta_j\neq 0$, and for reference $\tilde{\bx}$, we have $f(\tilde \bx) = 0$ (i.e., reference predicted as negative). Let $g^{\textrm{\rm ref}}(\boldsymbol{\beta}) = \beta_0+\sum_{j=1}^{p} \beta_{j}\tilde{x}_j$ be the raw score of the classifier $f$ at reference $\tilde \bx$. For all $k\in\{2,3,\cdots,p\}$, the volume of region $V_k$ in the input space $\mathbb{R}^p$ with SEV${}^{+}= k$ is: 
\begin{equation}
V_k=c_k \cdot \prod_{j=1}^{p} \left\lvert \frac{g^{\textrm{\rm ref}}(\boldsymbol{\beta})}{\beta_j}\right\rvert,
\end{equation}
where $c_k$ is a finite constant unrelated to the $\beta$'s.

\noindent \textit{Proof:}\newline

\noindent Consider a simple linear classifier $h(\bx):= \mathbf{1}[(-1 + \sum_{j=1}^p x_j)>0]$. Let $\tilde \bx$ be a reference where the reference feature values are all 0, i.e., $\tilde \bx=\mathbf{0}$. Consider a point $\bx$ predicted as positive by the classifier $h$. The \sev{+} of $\bx$ is greater or equal to 2 if and only if $x_j\leq 1,\forall j\in\{1,\cdots,p\}.$ Let us show both directions.

\paragraph{If:}

If $x_j\leq 1,\forall j\in\{1,\cdots,p\},$ for any vertex on the \sev{} hypercube one step away from the origin, which can be represented as $\bx_v^{(j)} = \be_j\odot \bx^{\q} + (\mathbf{1} - \be_j) \odot \tilde{\bx}$, we have $h(\bx_v^{(j)}) = \mathbf{1}[x_j - 1 > 0] = 0.$ This suggests that any vertex one step away from the origin fails to flip the prediction, which means \sev{+}$\geq 2.$

\paragraph{Only if:}

Suppose there exists $k$ such that $x_k>1$, then for the vertex $\bx_v^{(k)} = \be_k\odot \bx^{\q} + (\mathbf{1} - \be_k) \odot \tilde{\bx}$ in the \sev{} hypercube, $h(\bx_v^{(k)}) = \mathbf{1}[-1 + x_k > 0]=1,$ which flips the sign of prediction but with \sev{+}$=1.$\newline

\noindent Therefore, the region with \sev{+}$\geq 2$ is the polytope comprised by the hyperplanes $(-1 + \sum_{j=1}^p x_j)>0$, and $x_j\leq 1,\forall j\in\{1,\cdots,p\}.$ The volume of this polytope is $1-\frac{1}{p!}=\frac{p!-1}{p!}$ (a unit hypercube whose volume is 1 with one corner cut off, and the volume of the corner is $\frac{1}{p!}$). \newline

\noindent As the corner represents a unit simplex, its volume is well known to be $1/p!$. For completeness, however, we supply a brief proof. Define the corner as a set $S:= \{(x_1,\cdots,x_p)^T| 0\leq x_i\leq 1,\ \sum_{m=1}^{p}x_m\leq 1\}.$ Consider another set  $T:= \{(y_1,\cdots,y_p)^T| 0\leq y_1 \leq y_2 \leq \cdots \leq y_p\leq 1\}.$ There exists an invertible linear transformation, represented by a matrix $A:= (a_{mj})$, where $a_{mm}=1$, $a_{mj} = -1$ for $m = j + 1$, and $a_{mj} = 0$ otherwise, that defines a one-to-one mapping between points in $S$ and points in $T$. To see this, consider a point $\mathbf{y} = (y_1, \cdots, y_p)^T \in T$ and its mapping $A\mathbf{y} = (y_1, y_2 - y_1, \cdots, y_p - y_{p - 1})^T$. The monotonicity of the $y_j$ ensures that each of these entries lies in $[0, 1]$. And the sum of the entries in $A\mathbf{y}$ is simply $\sum_{j=1}^p y_j = y_p \leq 1$. Therefore $A\mathbf{y} \in S$ and $\mathrm{Vol}(S) = \mathrm{Det}(A)\mathrm{Vol}(T)$. $\mathrm{Vol}(T)=\frac{1}{p!}$ because the volume is the same as the probability that a random permutation is in sorted order, and $\mathrm{Det}(A)=1$ since $A$ is lower triangular with all diagonal elements equal to 1. Thus, $\mathrm{Vol}(S)=\frac{1}{p!}.$ \newline

\noindent Within this polytope (unit cube with one corner cut off), suppose the proportion of the region with \sev{+}$=k$ is $q_k$, then the size of this region is $\frac{q_k(p!-1)}{p!}.$ \newline

\noindent Define $c_k:=\frac{q_k(p!-1)}{p!}.$ For a linear classifier, $f(\bx):= \mathbf{1}[(\beta_0 + \sum_{j=1}^p \beta_j x_j)>0]
$, where for all $j$, and $f(\tilde \bx)=0$, let $g^{\textrm{\rm ref}}(\boldsymbol{\beta})$ as the raw output of the linear classifier $f$, we can apply change of variables: $x'_j = -\frac{\beta_j(x_j-\tilde x_j)}{g^{\textrm{\rm ref}}(\boldsymbol{\beta})}$ for all $j\in \{1,\cdots,p\}$. Then, the classifier becomes  $\mathbf{1}[(-1 + \sum_{j=1}^p x'_j)>0]$ (the greater than sign remains because we assume $f(\tilde \bx)=0$, i.e., $g^{\textrm{\rm ref}}(\boldsymbol{\beta}) =\beta_0+\sum_{j=1}^{p}\beta_j \tilde x_j<0$),  and the reference value $\tilde x'_j$ becomes 0. This is the same as the setting of $g$ mentioned above. Hence, the volume of the region with \sev{+}$=k$ after the change of variables is $c_k.$ Since the change of variable is a linear transformation, the volume of the same region before transformation can be obtained by dividing the new volume with the absolute value of the determinant of the transformation matrix, i.e.,

\begin{equation*}
\begin{aligned}
    V_k &= c_k / \left|\prod_{j=1}^p -\frac{\beta_j}{g^{\textrm{\rm ref}}(\boldsymbol{\beta})} \right|\\ &= c_k \cdot \prod_{j=1}^{p} \left\lvert \frac{g^{\textrm{\rm ref}}(\boldsymbol{\beta})}{\beta_j}\right\rvert,
\end{aligned}
\end{equation*}

\noindent where, as mentioned above, $c_k=\frac{q_k(p!-1)}{p!}$ is a finite constant unrelated to the $\beta$'s.

\rightline{$\square$}

\clearpage
\section{The difference between explanations of \citet{chen2018learning} and sparse explanation value}
\label{subsec:chen2018-sev-compare}

The work of \citet{chen2018learning} is fundamentally different from ours in several ways. 
\begin{itemize}
\item First, they impose that each query has an explanation of size $k$ (manually determined), whereas SEV varies across queries and is often 1. 
\item Second, their explanations are designed (i.e., optimized) to set features \textit{outside} the explanation to 0. Note that setting many features to 0 is likely to be out of distribution. Thus, predictions made at these points are not meaningful. (For instance, they would set all dummy categorical variables of a data point to 0, which means the point is no longer in any category, which is nonsensical.) Our \sev{-} instead considers what would happen if we set positive features \textit{inside} the explanation to a reference value, leaving all other features the same; these points are more likely to be in distribution. \sev{+} starts from a reference (which, when chosen properly, is in distribution), and some features are set to the feature value of the data point.
\item Third, SEV can be used for many different kinds of models, whereas their approach must be applied only to differentiable models, such as neural networks. (It cannot be used for decision trees, for example.) Because they use a complex neural network as their model, the computation within the explanation is a black box -- we do not know how the features are combined to form a prediction.
\item Fourth, their approach requires optimization of their sparsity metric, which hurts accuracy (see Figure \ref{fig:L2X_Adult_Accuracy} to \ref{fig:L2X_FICO_AUC}). As we show in Appendix \ref{subsec:results_overview}, SEV is already small for black box models as shown in Section \ref{subsec:low_sev}, so optimizing it might not even be necessary.
\end{itemize}
\ifshow
\begin{figure}[ht]
    \centering
    \begin{subfigure}[b]{0.4\textwidth}
    \includegraphics[width=\textwidth]{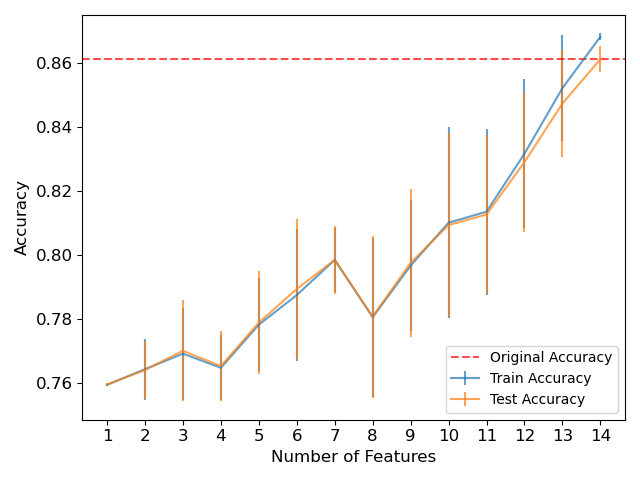}
    \caption{The relationship between accuracy and $k$ for L2X Model in Adult}
    \label{fig:L2X_Adult_Accuracy}
    \end{subfigure}
    \hfill
    \begin{subfigure}[b]{0.4\textwidth}
    \centering
    \includegraphics[width=\textwidth]{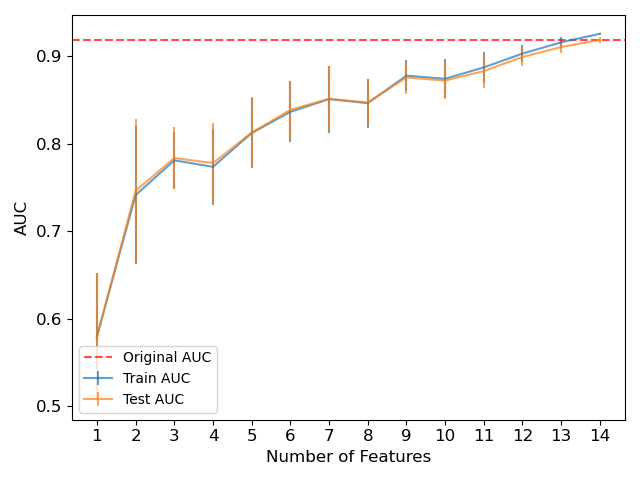}
    \caption{The relationship between AUC and $k$ for L2X Model in Adult}
    \label{fig:L2X_Adult_AUC}
    \end{subfigure}
    \begin{subfigure}[b]{0.4\textwidth}
    \centering
    \includegraphics[width=\textwidth]{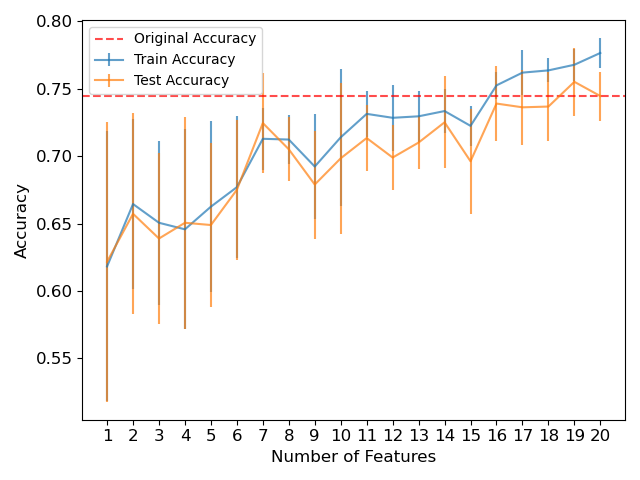}
    \caption{The relationship between accuracy and $k$ for L2X Model in German Credit}
    \label{fig:L2X_German_Accuracy}
    \end{subfigure}
    \hfill
    \begin{subfigure}[b]{0.4\textwidth}
    \centering
    \includegraphics[width=\textwidth]{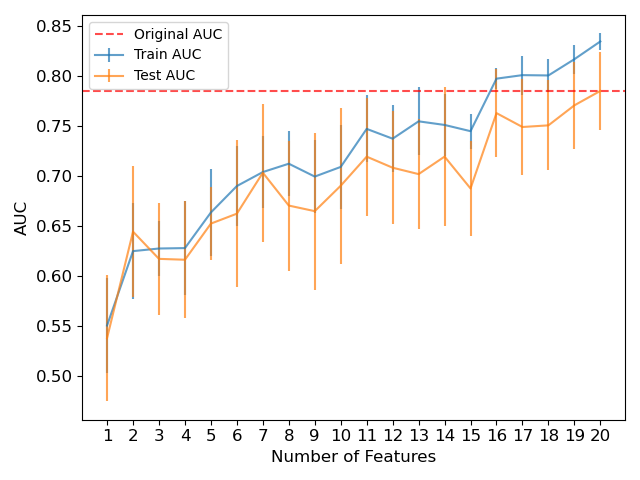}
    \caption{The relationship between AUC and $k$ for L2X Model in German Credit}
    \label{fig:L2X_German_AUC}
    \end{subfigure}
    \caption{The relationship between model performance and $k$ in L2X Model}
\end{figure}
\begin{figure}[ht]
\begin{subfigure}[b]{0.4\textwidth}
    \centering
    \includegraphics[width=\textwidth]{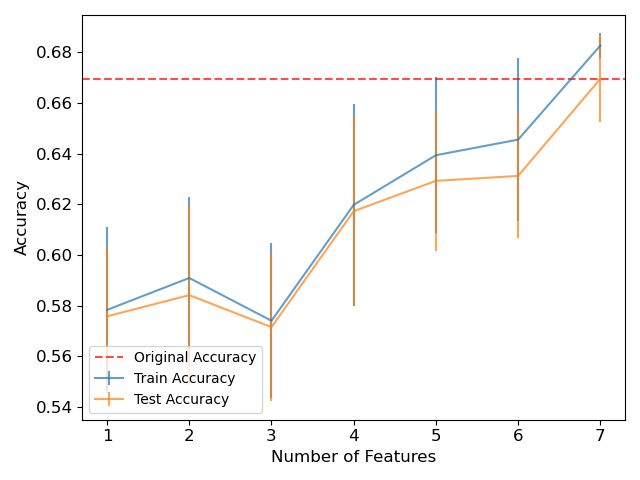}
    \caption{The relationship between accuracy and $k$ for L2X Model in COMPAS}
    \label{fig:L2X_COMPAS_Accuracy}
\end{subfigure}
\hfill
\begin{subfigure}[b]{0.4\textwidth}
    \centering
    \includegraphics[width=\textwidth]{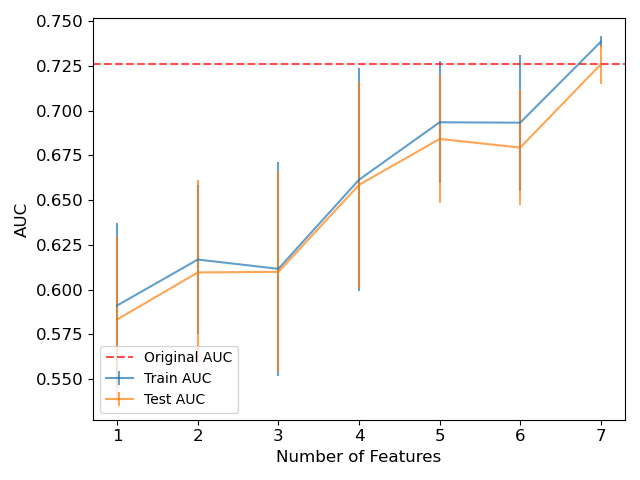}
    \caption{The relationship between AUC and $k$ for L2X Model in COMPAS}
    \label{fig:L2X_COMPAS_AUC}
\end{subfigure}
\begin{subfigure}[b]{0.4\textwidth}
    \centering
    \includegraphics[width=\textwidth]{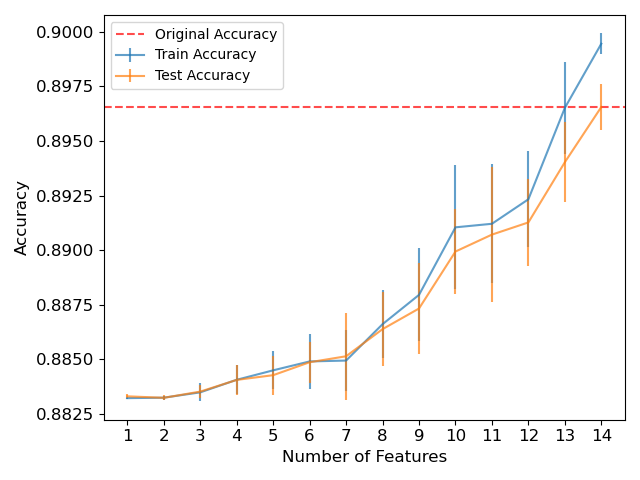}
    \caption{The relationship between accuracy and $k$ for L2X Model in MIMIC}
    \label{fig:L2X_MIMIC_Accuracy}
\end{subfigure}
\hfill
\begin{subfigure}[b]{0.4\textwidth}
    \centering
    \includegraphics[width=\textwidth]{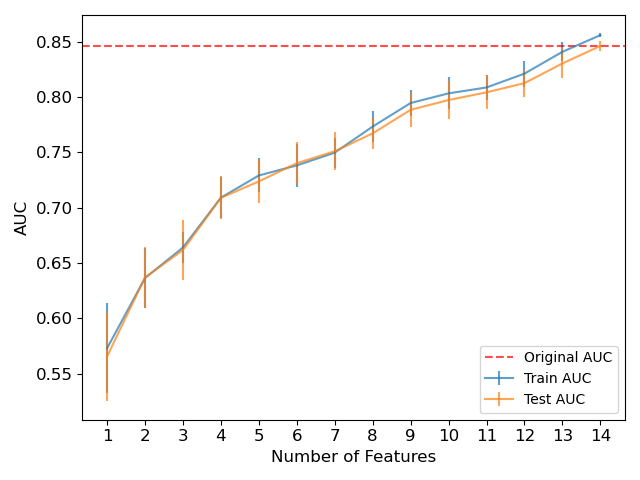}
    \caption{The relationship between AUC and $k$ for L2X Model in MIMIC}
    \label{fig:L2X_MIMIC_AUC}
\end{subfigure}
\begin{subfigure}[b]{0.4\textwidth}
    \centering
    \includegraphics[width=\textwidth]{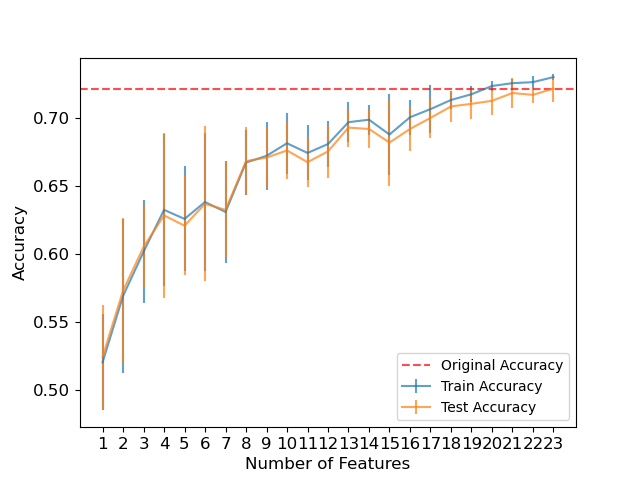}
    \caption{The relationship between accuracy and $k$ for L2X Model in FICO}
    \label{fig:L2X_FICO_Accuracy}
\end{subfigure}
\hfill
\begin{subfigure}[b]{0.4\textwidth}
    \centering
    \includegraphics[width=\textwidth]{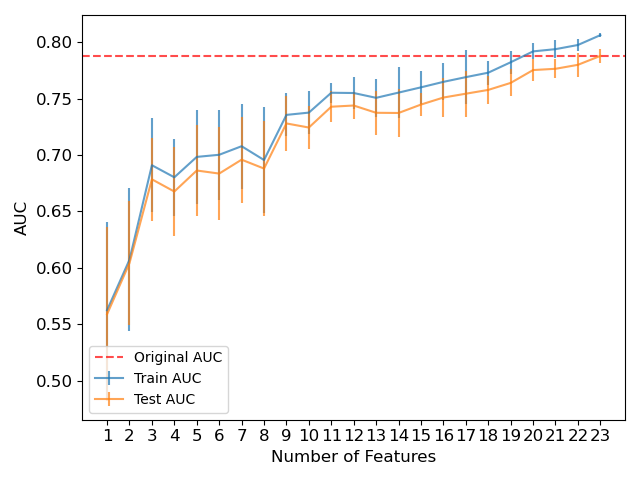}
    \caption{The relationship between AUC and $k$ for L2X Model in FICO}
    \label{fig:L2X_FICO_AUC}
\end{subfigure}
\caption{The relationship between model performance and $k$ in L2X Model (cont.)}
\end{figure}

\fi

}\fi

\clearpage
\section{Examples for \sev{+} and \sev{-} explanations}

In this section, we show some examples of \sev{+} and \sev{-} explanations. Tables \ref{tab:adult_explanation} to \ref{tab:german_explanation} show a few \sev{+} and \sev{-} calculations for various queries. 
For \sev{+} explanations, we show the query, the reference, and values that are aligned to the query (from the reference). For \sev{-}, we show the values aligned to the reference (from the query). 
For instance, taking the first query of Table \ref{tab:adult_explanation}, the first row represents the query, and the second row is the reference. The following two rows show different \sev{+} explanations: if we start from the reference and change its 9.6 year education to the query's 13 years, and change the occupation from administrative clerical to executive manager, the income will be predicted as over 50K. Or, if, starting from the reference,  education year is changed to 13, and the marital status is changed to civilian spouse, then income will be predicted as over 50K. The next three lines are the \sev{-} explanations: starting from the query, if we change either the year of education, the hours worked per week, or the marital status from the query values to the reference values, the income will be predicted as less than 50K.
\begin{table}[ht]
\caption{Examples of \sev{+}/\sev{-} Explanations for Adult Income Dataset}
\label{tab:adult_explanation}
\centering
\scalebox{0.72}{
\begin{tabular}{ccccccccccc}
\hline
\textbf{Type} & \textbf{age} & \textbf{\begin{tabular}[c]{@{}c@{}}education\\ year\end{tabular}} & \textbf{capital gain} & \textbf{capital loss} & \textbf{hour per week} & \textbf{workclass} & \textbf{marital status} & \textbf{occupation} & \textbf{relationship} & \textbf{race} \\ \hline\hline
Query & 29 & 13 & 0 & 0 & 55 & Private & Married-civ-spouse & Exec-managerial & Husband & Black \\
Reference & 36.78 & 9.6 & 148.75 & 53.14 & 38.84 & Private & Never-married & Adm-clerical & Not-in-family & White \\ \hline
\begin{tabular}[c]{@{}c@{}}\sev{+}\\ Explanations\end{tabular} & --- & 13 & --- & --- & --- & --- & --- & Exec-managerial & --- & --- \\
 & --- & 13 & --- & --- & --- & --- & Married-civ-spouse & --- & --- & --- \\\hline
\begin{tabular}[c]{@{}c@{}}\sev{-}\\ Explanations\end{tabular} & --- & 9.6 & --- & --- & --- & --- & --- & --- & --- & --- \\
 & --- & --- & --- & --- & --- & --- & Never-married & --- & --- & --- \\
 & --- & --- & --- & --- & 38.84 & --- & --- & --- & --- & --- \\ \hline\hline
Query & 50 & 9 & 3103 & 0 & 40 & Private & Married-civ-spouse & Craft-repair & Husband & White \\
Reference & 36.78 & 9.6 & 148.75 & 53.14 & 38.84 & Private & Never-married & Adm-clerical & Not-in-family & White \\ \hline
\begin{tabular}[c]{@{}c@{}}\sev{+}\\ Explanation\end{tabular} & --- & --- & 3103 & --- & --- & --- & Married-civ-spouse & --- & --- & --- \\\hline
\begin{tabular}[c]{@{}c@{}}\sev{-}\\ Explanation\end{tabular} & 36.78 & --- & --- & --- & --- & --- & --- & --- & --- & --- \\
 & --- & --- & --- & --- & --- & --- & Never-married & --- & --- & --- \\
 & --- & --- & --- & --- & --- & --- & --- & --- & --- & --- \\
 & --- & --- & 148.75 & --- & --- & --- & --- & --- & --- & --- \\ \hline\hline
Query & 30 & 13 & 0 & 1902 & 40 & Private & Married-civ-spouse & Adm-clerical & Husband & White \\
Reference & 36.78 & 9.6 & 148.75 & 53.14 & 38.84 & Private & Never-married & Adm-clerical & Not-in-family & White \\ \hline
\begin{tabular}[c]{@{}c@{}}\sev{+}\\ Explanation\end{tabular} & --- & --- & --- & 1902 & --- & --- & Married-civ-spouse & --- & --- & --- \\ \hline
\begin{tabular}[c]{@{}c@{}}\sev{-}\\ Explanation\end{tabular} & --- & --- & --- & --- & --- & --- & Never-married & --- & --- & --- \\
 & --- & --- & --- & 53.14 & --- & --- & --- & --- & --- & --- \\ \hline\hline
Query & 45 & 10 & 0 & 0 & 65 & Federal-gov & Married-civ-spouse & Transport-moving & Husband & White \\
Reference & 36.78 & 9.6 & 148.75 & 53.14 & 38.84 & Private & Never-married & Adm-clerical & Not-in-family & White \\ \hline
\begin{tabular}[c]{@{}c@{}}\sev{+}\\ Explanation\end{tabular} & --- & --- & --- & --- & 65 & Federal-gov & Married-civ-spouse & --- & --- & --- \\
 & --- & --- & 0 & --- & 65 & --- & Married-civ-spouse & --- & --- & --- \\ \hline
\begin{tabular}[c]{@{}c@{}}\sev{-}\\ Explanation\end{tabular} & --- & --- & --- & --- & --- & --- & Never-married & --- & --- & --- \\ \hline\hline
Query & 41 & 13 & 0 & 0 & 12 & Private & Married-civ-spouse & Sales & Wife & White \\
Reference & 36.78 & 9.6 & 148.75 & 53.14 & 38.84 & Private & Never-married & Adm-clerical & Not-in-family & White \\ \hline
\begin{tabular}[c]{@{}c@{}}\sev{+}\\ Explanation\end{tabular} & --- & 13 & --- & --- & --- & --- & Married-civ-spouse & --- & Wife & --- \\ \hline
\begin{tabular}[c]{@{}c@{}}\sev{-}\\ Explanation\end{tabular} & --- & 9.6 & --- & --- & --- & --- & --- & --- & --- & --- \\
 & --- & --- & --- & --- & --- & --- & Never-married & --- & --- & --- \\
 & --- & --- & --- & --- & --- & --- & --- & --- & Not-in-family & --- \\
 & --- & --- & 148.75 & --- & --- & --- & --- & --- & --- & --- \\ \hline
\end{tabular}}
\end{table}

\begin{table}[ht]
\caption{Examples of \sev{+}/\sev{-} Explanations for MIMIC Dataset}
\label{tab:mimic_explanation}
\centering
\scalebox{0.88}{
\begin{tabular}{cccccccccccc}
\hline
\textbf{Type} & \textbf{age} & \textbf{preiculos} & \textbf{gcs} & \textbf{\begin{tabular}[c]{@{}c@{}}heartrate\\ min\end{tabular}} & \textbf{\begin{tabular}[c]{@{}c@{}}meanbp\\ min\end{tabular}} & \textbf{\begin{tabular}[c]{@{}c@{}}resprate\\ min\end{tabular}} & \textbf{\begin{tabular}[c]{@{}c@{}}resprate\\ max\end{tabular}} & \textbf{\begin{tabular}[c]{@{}c@{}}tempc\\ min\end{tabular}} & \textbf{\begin{tabular}[c]{@{}c@{}}tempc\\ max\end{tabular}} & \textbf{\begin{tabular}[c]{@{}c@{}}urine\\ output\end{tabular}} & \textbf{mechvent} \\ \hline\hline
Query & 59.2 & 0.4 & 3.0 & 49.0 & 54.0 & 9.0 & 16.0 & 34.3 & 36.3 & 240.0 & 1.0 \\
Reference & 72.8 & 2215.9 & 13.9 & 70.9 & 59.4 & 12.2 & 26.9 & 36.1 & 37.5 & 2024.1 & 0.0 \\ \hline
\begin{tabular}[c]{@{}c@{}}\sev{+}\\ Explanation\end{tabular} & --- & --- & 3.0 & --- & --- & --- & --- & 34.3 & 36.3 & 240.0 & --- \\\hline
\begin{tabular}[c]{@{}c@{}}\sev{-}\\ Explanation\end{tabular} & --- & --- & 13.9 & --- & --- & --- & --- & --- & --- & --- & --- \\
 & --- & --- & --- & --- & --- & --- & --- & 36.1 & --- & --- & --- \\
 & --- & --- & --- & --- & --- & --- & --- & --- & 37.5 & --- & --- \\
 & --- & --- & --- & --- & --- & --- & --- & --- & --- & 2024.1 & --- \\
 & --- & --- & --- & --- & --- & --- & --- & --- & --- & --- & 0.0 \\ \hline\hline
Query & 85.8 & 1.6 & 15.0 & 87.0 & 61.0 & 18.0 & 26.0 & 35.9 & 36.4 & 183.0 & 1.0 \\
Reference & 72.8 & 2215.9 & 13.9 & 70.9 & 59.4 & 12.2 & 26.9 & 36.1 & 37.5 & 2024.1 & 0.0 \\ \hline
\begin{tabular}[c]{@{}c@{}}\sev{+}\\ Explanation\end{tabular} & 85.8 & 1.6 & --- & --- & --- & 18.0 & --- & --- & 36.4 & 183.0 & 1.0 \\\hline
\begin{tabular}[c]{@{}c@{}}\sev{-}\\ Explanation\end{tabular} & 72.8 & --- & --- & --- & --- & --- & --- & --- & --- & --- & --- \\
 & --- & --- & --- & --- & --- & 12.2 & --- & --- & --- & --- & --- \\
 & --- & --- & --- & --- & --- & --- & --- & --- & --- & 2024.1 & --- \\
 & --- & --- & --- & --- & --- & --- & --- & --- & --- & --- & 0.0 \\ \hline\hline
Query & 87.4 & 6953.8 & 3.0 & 38.0 & 47.0 & 11.0 & 45.0 & 35.9 & 37.4 & 852.0 & 1.0 \\
Reference & 72.8 & 2215.9 & 13.9 & 70.9 & 59.4 & 12.2 & 26.9 & 36.1 & 37.5 & 2024.1 & 0.0 \\ \hline
\begin{tabular}[c]{@{}c@{}}\sev{+}\\ Explanation\end{tabular} & 87.4 & 6953.8 & 3.0 & 38.0 & --- & --- & 45.0 & --- & --- & 852.0 & --- \\
 & 87.4 & --- & 3.0 & 38.0 & --- & --- & 45.0 & --- & --- & 852.0 & 1.0 \\ \hline
\begin{tabular}[c]{@{}c@{}}\sev{-}\\ Explanation\end{tabular} & 72.8 & --- & --- & --- & --- & --- & --- & --- & --- & --- & --- \\
 & --- & --- & 13.9 & --- & --- & --- & --- & --- & --- & --- & --- \\
 & --- & --- & --- & 70.9 & --- & --- & --- & --- & --- & --- & --- \\
 & --- & --- & --- & --- & --- & --- & 26.9 & --- & --- & --- & --- \\
 & --- & --- & --- & --- & --- & --- & --- & --- & --- & 2024.1 & --- \\
 & --- & --- & --- & --- & --- & --- & --- & --- & --- & --- & 0.0 \\ \hline\hline
Query & 78.7 & 0.5 & 15.0 & 17.0 & 21.0 & 20.0 & 35.0 & 32.3 & 35.2 & 8.0 & 1.0 \\
Reference & 72.8 & 2215.9 & 13.9 & 70.9 & 59.4 & 12.2 & 26.9 & 36.1 & 37.5 & 2024.1 & 0.0 \\ \hline
\begin{tabular}[c]{@{}c@{}}\sev{+}\\ Explanation\end{tabular} & --- & --- & --- & 17.0 & 21.0 & --- & --- & --- & --- & 8.0 & --- \\
 & --- & --- & --- & 17.0 & --- & --- & --- & 32.3 & 35.2 & --- & --- \\
 & --- & --- & --- & 17.0 & --- & --- & --- & 32.3 & --- & 8.0 & --- \\
 & --- & --- & --- & 17.0 & --- & --- & --- & --- & 35.2 & 8.0 & --- \\
 & --- & --- & --- & --- & 21.0 & --- & --- & 32.3 & --- & 8.0 & --- \\
 & --- & --- & --- & --- & 21.0 & --- & --- & --- & 35.2 & 8.0 & --- \\
 & --- & --- & --- & --- & --- & 20.0 & --- & 32.3 & 35.2 & --- & --- \\
 & --- & --- & --- & --- & --- & --- & --- & 32.3 & 35.2 & 8.0 & --- \\\hline
\begin{tabular}[c]{@{}c@{}}\sev{-}\\ Explanation\end{tabular} & --- & --- & --- & --- & --- & --- & --- & 36.1 & --- & 2024.1 & 0.0 \\ \hline
\end{tabular}}
\end{table}

\begin{table}[ht]
\caption{Examples \sev{+}/\sev{-} Explanations for FICO Dataset}
\label{tab:fico_explanation}
\centering
\scalebox{0.60}{
\begin{tabular}{ccccccccccc}
\hline
Type & \begin{tabular}[c]{@{}c@{}}External\\ RiskEstimate\end{tabular} & \begin{tabular}[c]{@{}c@{}}MSinceMost\\ RecentTradeOpen\end{tabular} & \begin{tabular}[c]{@{}c@{}}NumSatis-\\ factoryTrades\end{tabular} & \begin{tabular}[c]{@{}c@{}}PercentTrades\\ NeverDelq\end{tabular} & \begin{tabular}[c]{@{}c@{}}MaxDelq2Public\\ RecLast12M\end{tabular} & \begin{tabular}[c]{@{}c@{}}PercentIn-\\ stallTrades\end{tabular} & \begin{tabular}[c]{@{}c@{}}NumInq\\ Last6M\end{tabular} & \begin{tabular}[c]{@{}c@{}}NumInqLast6\\ Mexcl7days\end{tabular} & \begin{tabular}[c]{@{}c@{}}NetFraction\\ RevolvingBurden\end{tabular} & \begin{tabular}[c]{@{}c@{}}NumRevolving\\ TradesWBalance\end{tabular} \\ \hline
Query & 79 & 138 & 1 & 100 & 7 & 100 & 0 & 0 & Missing & Missing \\
Reference & 72.21 & 9.2 & 21.1 & 89.98 & 5.36 & 29.82 & 0.6 & 0.56 & 22.26 & 2.95 \\ \hline
\begin{tabular}[c]{@{}c@{}}\sev{+}\\ Explanation\end{tabular} & --- & 138 & --- & --- & --- & --- & --- & --- & --- & --- \\
 & --- & --- & --- & --- & --- & 100 & --- & --- & --- & --- \\ \hline
\sev{-} Explanation & --- & 9.2 & --- & --- & --- & --- & --- & --- & --- & --- \\
 & --- & --- & --- & --- & --- & 29.82 & --- & --- & --- & --- \\ \hline\hline
Query & 60 & 8 & 55 & 95 & 4 & 34 & 1 & 1 & 54 & 6 \\
Reference & 72.21 & 9.2 & 21.1 & 89.98 & 5.36 & 29.82 & 0.6 & 0.56 & 22.26 & 2.95 \\ \hline
\begin{tabular}[c]{@{}c@{}}\sev{+}\\ Explanation\end{tabular} & --- & --- & --- & --- & --- & --- & --- & --- & 54 & --- \\ \hline
\begin{tabular}[c]{@{}c@{}}\sev{-}\\ Explanation\end{tabular} & 72.21 & --- & --- & --- & --- & --- & --- & --- & --- & --- \\
 & --- & --- & --- & --- & --- & --- & --- & --- & 22.26 & --- \\ \hline\hline
Query & 59 & 12 & 18 & 85 & 2 & 30 & 10 & 10 & 95 & 5 \\
Reference & 72.21 & 9.2 & 21.1 & 89.98 & 5.36 & 29.82 & 0.6 & 0.56 & 22.26 & 2.95 \\ \hline
\begin{tabular}[c]{@{}c@{}}\sev{+}\\ Explanation\end{tabular} & --- & --- & --- & --- & --- & --- & 10 & --- & --- & --- \\
 & --- & --- & --- & --- & --- & --- & --- & --- & 95 & --- \\ \hline
\begin{tabular}[c]{@{}c@{}}\sev{-}\\ Explanation\end{tabular} & 72.21 & --- & --- & --- & 5.36 & --- & 0.6 & 0.56 & 22.26 & --- \\ \hline\hline
Query & 57 & 4 & 9 & 43 & 0 & 14 & 6 & 6 & 43 & 3 \\
Reference & 72.21 & 9.2 & 21.1 & 89.98 & 5.36 & 29.82 & 0.6 & 0.56 & 22.26 & 2.95 \\ \hline
\begin{tabular}[c]{@{}c@{}}\sev{+}\\ Explanation\end{tabular} & --- & --- & --- & 43 & --- & --- & --- & --- & --- & --- \\
 & --- & --- & --- & --- & --- & --- & 6 & --- & --- & --- \\ \hline
\begin{tabular}[c]{@{}c@{}}\sev{-}\\ Explanation\end{tabular} & 72.21 & --- & --- & 89.98 & 5.36 & --- & 0.6 & --- & 22.26 & --- \\ \hline\hline
Query & 82 & 61 & 3 & 100 & 7 & 33 & 0 & 0 & 50 & 1 \\
Reference & 72.21 & 9.2 & 21.1 & 89.98 & 5.36 & 29.82 & 0.6 & 0.56 & 22.26 & 2.95 \\ \hline
\begin{tabular}[c]{@{}c@{}}\sev{+}\\ Explanation\end{tabular} & --- & 61 & --- & --- & --- & --- & --- & --- & --- & --- \\ \hline
\begin{tabular}[c]{@{}c@{}}\sev{-}\\ Explanation\end{tabular} & --- & 9.2 & 21.1 & --- & --- & --- & --- & --- & --- & --- \\
 & --- & 9.2 & --- & --- & --- & --- & --- & --- & 22.26 & --- \\ \hline
\end{tabular}}
\end{table}

\begin{table}[ht]
\caption{Examples \sev{+}/\sev{-} Explanations for German Credit Dataset. 0DM in the Checking account status column means that checking amount has no Deutsche marks (DM) in it, and 0-200DM means that the checking amount has 0-200 Deutsche marks (DM).}
\label{tab:german_explanation}
\centering
\scalebox{0.95}{
\begin{tabular}{cccccccc}
\hline
\textbf{Type} & \textbf{Duration} & \textbf{Credit Amount} & \textbf{Age} & \textbf{\begin{tabular}[c]{@{}c@{}}Checking\\  account status\end{tabular}} & \textbf{Purpose} & \textbf{Other debtors} & \textbf{Housing} \\ \hline\hline
Query & 18 & 4380 & 35 & 0DM & car(new) & none & own \\
Reference & 19.21 & 2985.46 & 36.22 & \begin{tabular}[c]{@{}c@{}}no checking\\ account\end{tabular} & radio/television & none & own \\ \hline
\begin{tabular}[c]{@{}c@{}}\sev{+}\\ Explanation\end{tabular} & --- & --- & --- & 0DM & --- & --- & --- \\
 & --- & --- & --- & --- & car(new) & --- & --- \\
 & --- & 2985.46 & --- & --- & --- & --- & --- \\ \hline
\begin{tabular}[c]{@{}c@{}}\sev{-}\\ Explanation\end{tabular} & --- & --- & --- & \begin{tabular}[c]{@{}c@{}}no checking\\ account\end{tabular} & --- & --- & --- \\
 & --- & --- & --- & --- & radio/television & --- & --- \\ \hline\hline
Query & 24 & 2325 & 32 & 0DM & car(new) & none & own \\
Reference & 19.21 & 2985.46 & 36.22 & \begin{tabular}[c]{@{}c@{}}no checking\\ account\end{tabular} & radio/television & none & own \\ \hline
\begin{tabular}[c]{@{}c@{}}\sev{+}\\ Explanation\end{tabular} & --- & --- & --- & 0DM & --- & --- & --- \\
 & 24 & --- & --- & --- & --- & --- & --- \\
 & --- & --- & --- & --- & car(new) & --- & --- \\ \hline
\begin{tabular}[c]{@{}c@{}}\sev{-}\\ Explanation\end{tabular} & --- & --- & --- & \begin{tabular}[c]{@{}c@{}}no checking\\ account\end{tabular} & radio/television & --- & --- \\ \hline\hline
Query & 12 & 585 & 20 & 0-200DM & radio/television & co-applicant & rent \\
Reference & 19.21 & 2985.46 & 36.22 & \begin{tabular}[c]{@{}c@{}}no checking\\ account\end{tabular} & radio/television & none & own \\ \hline
\begin{tabular}[c]{@{}c@{}}\sev{+}\\ Explanation\end{tabular} & --- & 585 & --- & --- & --- & --- & --- \\ \hline
\begin{tabular}[c]{@{}c@{}}\sev{-}\\ Explanation\end{tabular} & --- & --- & --- & --- & --- & none & --- \\
 & --- & --- & 36.22 & --- & --- & --- & --- \\
 & --- & --- & --- & --- & --- & --- & own \\ \hline\hline
Query & 36 & 2323 & 24 & 0-200DM & radio/television & none & rent \\
Reference & 19.21 & 2985.46 & 36.22 & \begin{tabular}[c]{@{}c@{}}no checking\\ account\end{tabular} & radio/television & none & own \\ \hline
\begin{tabular}[c]{@{}c@{}}\sev{+}\\ Explanation\end{tabular} & 36 & --- & --- & --- & --- & --- & --- \\ \hline
\begin{tabular}[c]{@{}c@{}}\sev{-}\\ Explanation\end{tabular} & 19.21 & --- & --- & --- & --- & --- & --- \\
 & --- & --- & 36.22 & --- & --- & --- & --- \\ \hline
\end{tabular}}
\end{table}

\end{document}